\documentclass{article}

\usepackage[main, preprint, nonatbib]{neurips_2026}

\usepackage[utf8]{inputenc}
\usepackage[T1]{fontenc}
\usepackage{hyperref}
\usepackage{url}
\usepackage{booktabs}
\usepackage{amsfonts}
\usepackage{nicefrac}
\usepackage{microtype}
\usepackage[table]{xcolor}
\usepackage{wrapfig}
\hypersetup{hidelinks}
\usepackage{graphicx}
\usepackage{subcaption}
\usepackage{bm}
\usepackage[percent]{overpic}
\usepackage[most]{tcolorbox}
\usepackage{multirow}
\usepackage{fontawesome5}

\usepackage{tikz}
\newcommand{\circnum}[1]{\tikz[baseline=(n.base)]\node[circle,fill=black,text=white,inner sep=0pt,minimum size=1.9ex,font=\footnotesize\bfseries\sffamily](n){#1};}

\definecolor{panelgreen}{RGB}{242,249,239}

\usepackage{amsmath}
\usepackage{amssymb}
\usepackage{mathtools}
\usepackage{amsthm}
\usepackage{needspace}
\usepackage{capt-of}

\usepackage[capitalize,noabbrev]{cleveref}

\theoremstyle{plain}
\newtheorem{theorem}{Theorem}[section]
\newtheorem{proposition}[theorem]{Proposition}
\newtheorem{lemma}[theorem]{Lemma}
\newtheorem{remark}[theorem]{Remark}
\newtheorem{corollary}[theorem]{Corollary}
\theoremstyle{definition}
\newtheorem{definition}[theorem]{Definition}
\newtheorem{assumption}[theorem]{Assumption}

\usepackage{algorithmic}
\usepackage{algorithm}

\def\mI{{\bm{I}}}
\definecolor{lightblue}{rgb}{0.95, 0.95, 1}

\newcommand{\IFNew}{\operatorname{IF}}

\usepackage{bm}

\newcommand{\x}{{\bf x}}

\def\mI{{\bm{I}}}

\newtcolorbox{mybox}[2][]
{
  colframe = #2!25,
  colback  = #2!25!white!25,
  left=1mm,
  top=1mm,
  #1,
  breakable
}

\newenvironment{theorembox}
   {\begin{mybox}{gray}\begin{theorem}}
   {\end{theorem}\end{mybox}}

\newenvironment{propositionbox}
   {\begin{mybox}{gray}\begin{proposition}}
   {\end{proposition}\end{mybox}}

\newenvironment{remarkbox}
{\begin{mybox}{gray}\begin{remark}}
{\end{remark}\end{mybox}}

\usepackage{amsmath,amsfonts,bm}

\def\eqref#1{equation~\ref{#1}}

\def\1{\bm{1}}

\def\mI{{\bm{I}}}

\DeclareMathAlphabet{\mathsfit}{\encodingdefault}{\sfdefault}{m}{sl}
\SetMathAlphabet{\mathsfit}{bold}{\encodingdefault}{\sfdefault}{bx}{n}

\title{A Theoretical Bridge Between Long-Tailed Recognition and Continual Learning}

\author{Mahdiyar Molahasani, Michael Greenspan, Ali Etemad \\
Department of Electrical and Computer Engineering\\
Queen’s University, Kingston, Canada \\
\texttt{\{m.molahasani,michael.greenspan, ali.etemad\}@queensu.ca} \\
}

\begin{document}

\maketitle

\begin{abstract}

We theoretically and empirically establish a previously unexplored connection between Long-Tailed Recognition (LTR) and Continual Learning (CL). Specifically, we show that training on a long-tailed dataset with imbalance factor $\mathrm{IF}$ drives model parameters into an $\mathcal{O}(1/\sqrt{\mathrm{IF}})$ neighborhood of the dominant-class solution, under both uniform and exponentially decaying cardinality distributions. Building on this result, we establish an objective-level reformulation of LTR as a sequential learning problem and prove that the CL objective upper-bounds the balanced LTR loss up to a constant determined by the residual Head loss. Motivated by this connection, we introduce Continual Learning for Long-Tailed Recognition (CLTR), a principled framework that leverages standard off-the-shelf CL methods to sequentially learn Head and Tail classes while mitigating catastrophic forgetting. Extensive experiments on CIFAR100-LT, CIFAR10-LT, ImageNet-LT, and Caltech256 validate our theoretical predictions and demonstrate strong performance across diverse LTR benchmarks. In the foundation-model regime, CLTR with DualPrompt on a CLIP backbone outperforms standard adaptation baselines and is competitive with specialized methods using external semantic supervision. Our work bridges LTR and CL both theoretically and empirically, providing a principled approach for addressing long-tailed learning using standard CL strategies.

\end{abstract}

\section{Introduction}

 Real-world data often exhibits long-tailed distributions \cite{buda2018systematic,reed2001pareto,zhang2021deep,fu2022long}, where a  small set of ``Head'' classes dominates with abundant samples, while the vast majority of ``Tail'' classes are severely underrepresented. This imbalance poses a fundamental challenge for deep learning: the Head classes disproportionately influence gradients during training, causing models to converge toward solutions that favor the Head while failing to generalize to the Tail \cite{alshammari2022long, haghanifar2022covid, rs16081398}.

\begin{figure*}[t]
    \centering
        \includegraphics[width=\textwidth]{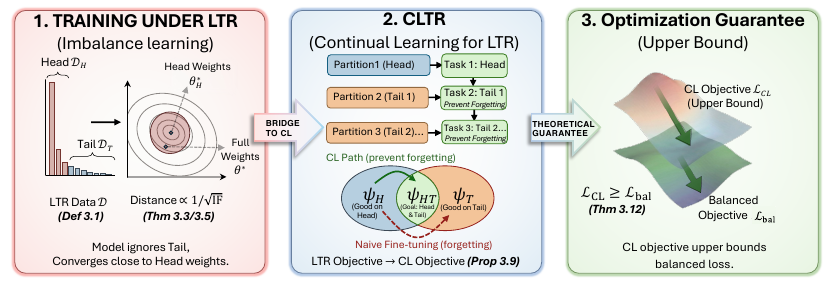}
\caption{Overview of our framework: (left) theoretical analysis of convergence under the LTR setup, showing how imbalance drives convergence toward head-class solutions; (middle) our proposed CLTR reformulating LTR as a CL problem;
and (right) the conditional optimization guarantee establishing that the CL objective upper-bounds the balanced LTR loss up to an additive constant.}
    \label{fig:overview}
\vspace{-0.5cm}
\end{figure*}

To address this issue, various methods have been proposed, including sample-wise balancing through over-sampling the Tail classes \cite{chawla2002smote,estabrooks2004multiple,feng2021exploring}, transfer learning approaches to adapt knowledge from Head to Tail \cite{liu2019large,wang2017learning,zhong2019unequal,jamal2020rethinking}, loss- or gradient-wise balancing to mitigate class size effects \cite{cao2019learning,cui2019class,tang2020long}, and weight regularization for ensuring balanced weight magnitudes during training \cite{alshammari2022long}. More recently, \textit{multi-stage training} has emerged as a promising solution for reducing sensitivity to Tail variations and model variance \cite{zhou2020bbn, zhang2022self}.  Concurrently, Continual Learning (CL) has evolved as a paradigm to learn sequences of tasks without forgetting previously acquired knowledge \cite{li2017learning}.  The structural similarity between multi-stage Long-Tailed Recognition (LTR) and sequential learning prompts a fundamental question:
\textit{Can solutions for continual learning be harnessed to solve long-tailed recognition?}

To answer this, we first analyze the optimization behavior induced by long-tailed training. We prove that training on a long-tailed dataset drives the learned parameters into an $\mathcal{O}(1/\sqrt{\IFNew})$ neighborhood of the solution obtained by training exclusively on the Head classes (Fig.~\ref{fig:overview} left). This reveals a structural bias in long-tailed optimization toward Head-dominated solutions and naturally motivates a sequential learning strategy: first learn the Head, then incorporate the Tail while preserving previously acquired Head knowledge. Building on this insight, we establish a principled connection between LTR and CL, showing that LTR can be reformulated as a sequential learning problem over cardinality-sorted partitions. Motivated by this perspective, we introduce Continual Learning for Long-Tailed Recognition (CLTR), a simple framework that sequentially learns Head and Tail partitions using standard off-the-shelf CL methods to mitigate catastrophic forgetting and maintain balanced performance (Fig.~\ref{fig:overview} middle). Finally, we provide a theoretical guarantee showing that the CL objective upper-bounds the balanced LTR objective, establishing CL as a principled optimization framework for LTR rather than merely a heuristic sequential training strategy (Fig.~\ref{fig:overview} right).

We validate CLTR through extensive experiments on five datasets: MNIST-LT, CIFAR100-LT, CIFAR10-LT, ImageNet-LT, and Caltech256. We first use MNIST-LT and CIFAR100-LT to empirically verify our theoretical predictions regarding the parameter dynamics of Head-only and full-dataset training. We then evaluate CLTR across standard LTR benchmarks under varying imbalance factors, consistently obtaining strong performances. Beyond standard ResNet settings, CLTR naturally integrates off-the-shelf CL methods, achieves strong results in the foundation-model adaptation regime, and extends effectively to the more challenging Long-Tailed Class-Incremental Learning (LT-CIL) setting.

Our contributions are summarized as follows:
\circnum{1}~
\textbf{Theoretical Analysis of Convergence:} We prove upper bounds on the distance between parameters learned on imbalanced partitions, showing this distance scales inversely with the square root of the imbalance factor, $\mathcal{O}(1/\sqrt{\IFNew})$ (Theorems \ref{theorem1} and \ref{theorem_non}).
\circnum{2}~\textbf{Reformulating LTR as CL:} We uncover a previously unexplored connection between long-tailed recognition and continual learning, establishing a formal objective-level reformulation of LTR as a CL problem (Proposition~\ref{proposition}), directly motivated by the convergence analysis in Theorems~\ref{theorem1} and~\ref{theorem_non}.
\circnum{3}~\textbf{The CLTR Framework:} Building on this reformulation, we introduce CLTR, a streamlined framework that learns Head and Tail classes sequentially using standard CL techniques to mitigate catastrophic forgetting (Algorithm \ref{alg:CLTR}).
\circnum{4}~\textbf{Optimization Guarantee:}
We establish sufficient conditions under which the CL objective upper-bounds the balanced LTR loss up to an additive constant (Theorem~\ref{theorem_general}), providing a principled surrogate objective for long-tailed recognition.
\circnum{5}~\textbf{Empirical Validation:} We demonstrate that CLTR achieves strong performance across standard LTR benchmarks, enables effective foundation model adaptation under severe class imbalance, and extends to complex LT-CIL scenarios.

\section{Related Work}
LTR methods have largely focused on re-balancing strategies, including re-sampling, loss re-weighting, decoupled training, and multi-stage optimization~\cite{cui2019class,cao2019learning,38kang2019decoupling,zhou2020bbn,b1wang2020long}. In parallel, CL methods address catastrophic forgetting in sequential learning through regularization, replay, expansion-based, or parameter-isolation mechanisms~\cite{kirkpatrick2017overcoming,li2017learning,chaudhry2018efficient,wang2022foster,saha2021gradient}. Despite structural similarities between multi-stage LTR and sequential CL, prior work has treated the two paradigms independently; in contrast, we establish a formal theoretical connection showing that LTR admits an objective-level reformulation as a CL problem. Please see an overview of the related works in Appendix \ref{app:related_works}.

\section{Proposed Approach}

\subsection{Training on Long-Tailed Distributions}
Here, we first define the LTR problem and then analyze the behavior of a model when trained under LTR setup to provide a theoretical basis for our proposed CLTR framework. For the summary of the notations see Appendix \ref{app:notation}.

Let's consider the input space to be $ \mathbb{R}^d $, where each input is represented by $ \x_i $, and the label space is $\{1, \ldots, k\}$, where each label is denoted by $ y_i $. Let $\mathcal{D}$ denote the training set containing samples $(\x_i,y_i)$.
$\mathcal{D}_c$ is a set of all samples belonging to class $c$ denoted as $\mathcal{D}_c = \{(\x_i, y_i) \in \mathcal{D} \mid y_i = c\}$ and $|\mathcal{D}_c|$ represents its cardinality. Without loss of generality, let the classes be ordered by their cardinalities such that  $|\mathcal{D}_{i}| \geq |\mathcal{D}_{j}|$ for all  $i < j$.  Following \cite{hong2024proaug}, let $\mathcal{D}_H$ and $\mathcal{D}_T$ represent the subsets of $\mathcal{D}$ corresponding to the Head set and Tail set, respectively as $\mathcal{D}_H=\{(\x_i,y_i)\in \mathcal{D}:y_i\le c_k\}$ and $\mathcal{D}_T=\{(\x_i,y_i)\in \mathcal{D}:y_i>c_k\}$, where $c_k$ denotes the number of classes in the Head set. As a result, every class in the Head has more samples than any class in the Tail. The loss function over $\mathcal{D}_c$ is defined as $\mathcal{L}(\mathcal{D}_c, \theta) = \frac{1}{|\mathcal{D}_c|}\sum_{i=1}^{|\mathcal{D}_c|} \ell((\x_i, y_i), \theta)$, where $(\x_i, y_i) \in \mathcal{D}_c$. Note that $\ell((\x_i, y_i), \theta)$ is the loss of each individual sample. For brevity, we will henceforth use notations $\mathcal{L}_{\mathcal{D}_c} = \mathcal{L}_{\mathcal{D}_c}(\theta) = \mathcal{L}(\mathcal{D}_c, \theta) $ and $\ell((\x_i,y_i)) = \ell_{(\x_i,y_i)}(\theta) = \ell((\x_i, y_i), \theta)$. We denote the average Head and Tail losses by $\mathcal{L}_H(\theta)=\mathcal{L}(\mathcal{D}_H,\theta)$ and $\mathcal{L}_T(\theta)=\mathcal{L}(\mathcal{D}_T,\theta)$, respectively.

LTR aims to address the challenge of learning from highly imbalanced data. This occurs when the training data $\mathcal{D}$ contains more samples in Head set ($\mathcal{D}_H$) and fewer in the Tail set ($\mathcal{D}_T$).
The imbalance factor $\IFNew$ quantifies the severity of this issue in a dataset, defined as:
\begin{equation}\label{IF_def}
\IFNew=\frac{|\mathcal{D}_{c^{\max}}|}{|\mathcal{D}_{c^{\min}}|},
\end{equation}
where $c^{\max} =  \arg \max_c~|\mathcal{D}_c|$, and $c^{\min} =  \arg \min_c~|\mathcal{D}_c|$, such that $\mathcal{D}_{c^{\max}} \in \mathcal{D}_H$ and $\mathcal{D}_{c^{\min}} \in \mathcal{D}_T$. Now we formally define a long-tailed dataset and the LTR problem:

\begin{definition}\label{def}
A dataset is deemed \emph{long-tailed} when $|\mathcal{D}_{c^{\max}}| \gg |\mathcal{D}_{c^{\min}}|$ or, in other words, $\IFNew~\gg~1$. When a model is trained on such a dataset and its performance is assessed on a test set where each class $c$ has the same number of samples (i.e. $|\mathcal{D}_c| = \kappa$ for each class $c$ within the test set where $\kappa$ is a constant number), the problem is referred to as \emph{Long-Tailed Recognition}.
\end{definition}

\subsection{Convex Case: Baseline Analysis}
\begin{figure}[t]
\centering
        \begin{minipage}[t]{.32\linewidth}
        \centering
        \includegraphics[width = 0.95\linewidth]{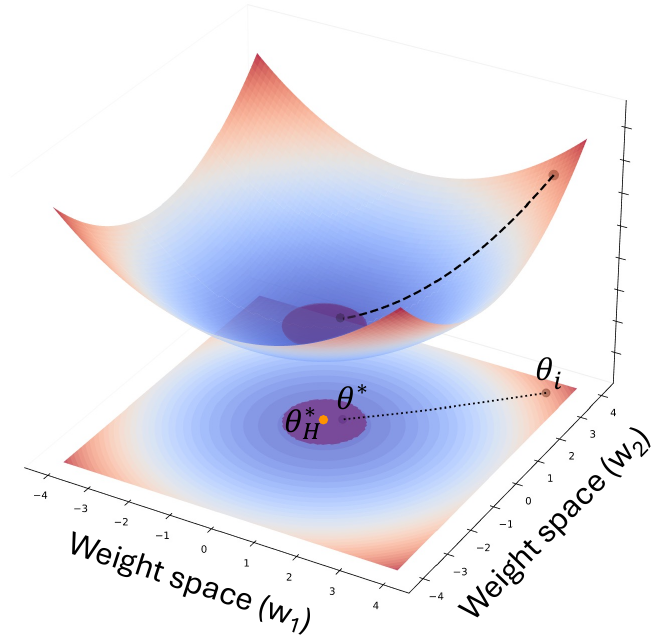}
    \end{minipage}
    \centering
    \begin{minipage}[t]{.32\linewidth}
        \centering
        \includegraphics[width = 0.95\linewidth]{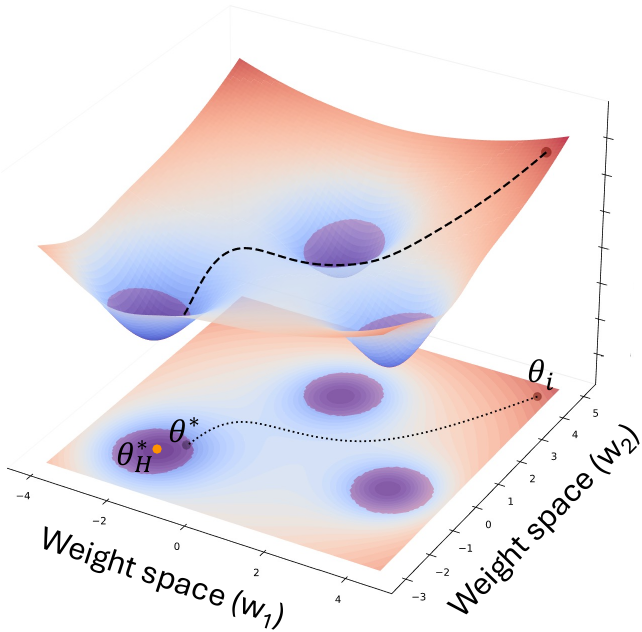}
\end{minipage}
    \begin{minipage}[t]{.31\linewidth}
        \centering
        \includegraphics[width = 1.1\linewidth]{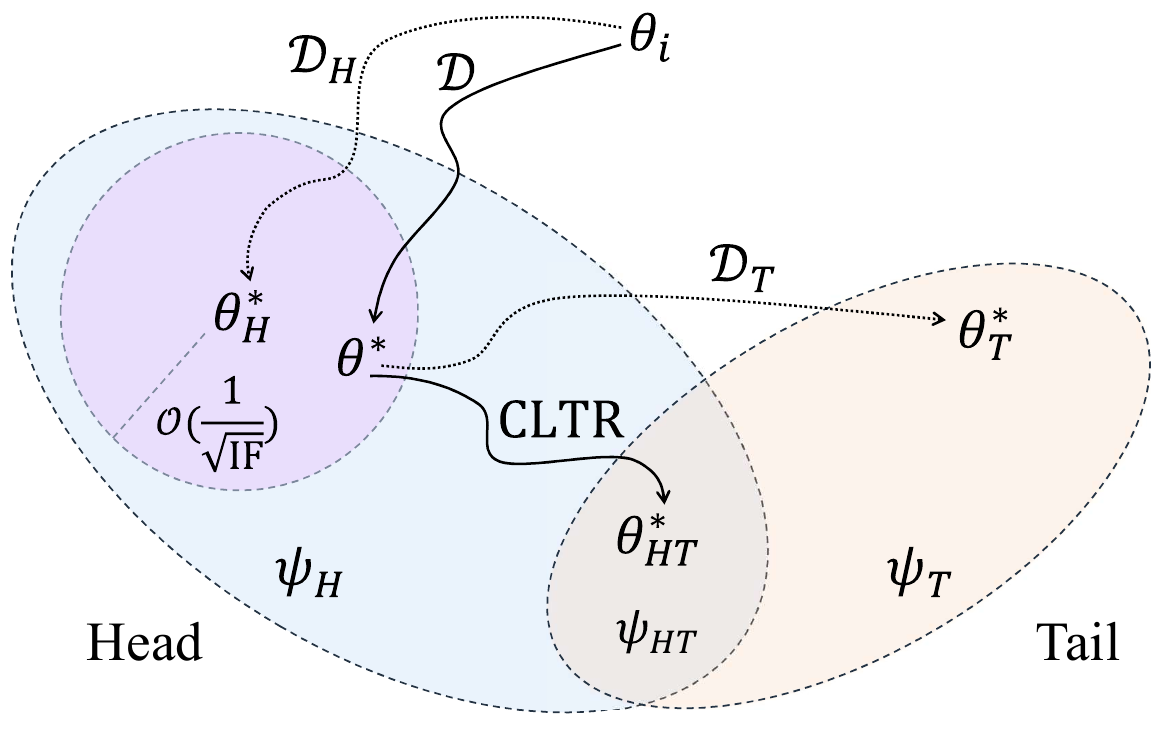}
\end{minipage}
\vspace{0.5cm}
\caption{ Left: Illustration of the Head loss landscape ($\mathcal{L}_H$) for convex setting (Assumption \ref{assumption_1}); the unique minimizer ($\theta_H^\star$) is marked in orange. Middle: Illustration of the Head loss landscape for feedforward deep network, KL((1/2)) setting (Assumption \ref{assumption_non}); multiple local minimizers (${\theta_{H,k}^\star}$) are shown in orange. In both panels, the red region denotes the provable neighborhood that contains the parameter ($\theta^\star$) obtained when training on the full long-tailed dataset; its radius scales as ($\mathcal{O}(1/\sqrt{\IFNew})$) by Theorems \ref{theorem1} and \ref{theorem_non}. (Top: surface; bottom: contour projection.). Right: Overview of learning under the LTR scenario and our proposed CLTR approach (symbols described in the text).}\label{fig:losslandscape}
\end{figure}

Here we address the case that the loss function is convex (as illustrated in Fig. \ref{fig:losslandscape} left), as detailed in the following:

\begin{assumption}\label{assumption_1}
Let the size of all Head classes be equal,
and the size of all Tail classes be equal,
i.e. $|\mathcal{D}_i| = |\mathcal{D}_j| $ for all $\mathcal{D}_i,\mathcal{D}_j \in \mathcal{D}_H$
and
$|\mathcal{D}_k| = |\mathcal{D}_l| $ for all $\mathcal{D}_k,\mathcal{D}_l \in \mathcal{D}_T$
(This assumption is a didactic simplification introduced to derive the $\mathcal{O}(1/\sqrt{\mathrm{IF}})$ scaling in closed form. It is systematically relaxed in Assumptions \ref{assumption_non} and \ref{assumption_3}.).
Also, let the model be a classifier with parameters $\theta$ trained with a convex loss function $\mathcal{L}$,
with an additional $L^2$ regularization term
utilized in the training to prevent the weights from growing excessively.
\end{assumption}

We now introduce Theorem \ref{theorem1} demonstrating the relationship between the weights of the model when it is trained solely on the Head as well as on the entire dataset.

\begin{theorembox}\label{theorem1}
Given Assumption \ref{assumption_1},
if a model is trained in an LTR setting (Definition \ref{def}), then the weights of the model after training ($\theta^*$) will lie within the bounded neighborhood of the model's weight if solely trained on Head ($\theta^*_H$), with the scale of:
\begin{equation}\label{eq_theorem}
{\|\theta^* - \theta^*_{H} \|} = \mathcal{O}(\frac{1}{\sqrt{\IFNew}}).
\end{equation}
For the full proof see Appendix \ref{theorem1_proof}.
\vspace{-3mm}
\end{theorembox}

\subsection{Extension to Feedforward Deep Networks}
\label{sec:relax-convexity}

Assumption~\ref{assumption_1} posits a convex loss with $L^2$ regularization, which is convenient for analysis but not representative of modern deep networks, whose training losses are typically nonconvex. We now relax convexity and work under the widely used \emph{Kurdyka--\L{}ojasiewicz} (KL) framework, together with a local geometric condition that is standard for deep nets near minimizers \cite{davis2020stochastic} (Fig. \ref{fig:losslandscape} middle).

\begin{assumption}
\label{assumption_non}
The model is a feedforward deep neural network with piecewise-analytic/definable activations (e.g., ReLU/LeakyReLU/softplus) and a standard empirical loss (e.g., cross-entropy or squared loss). Hence, the loss $\mathcal{L}$ enjoys the Kurdyka--\L{}ojasiewicz (KL) property. We assume that the Head loss $\mathcal{L}_H$ has KL exponent $1/2$ and satisfies a local H\"older-$1/2$ error bound in a neighborhood of $S_H=\arg\min_{\theta}\mathcal{L}_H(\theta)$, the set of Head-loss minimizers.
\end{assumption}

\noindent
For a set $S$, we define $\operatorname{dist}(\theta,S)=\inf_{\bar{\theta}\in S}\|\theta-\bar{\theta}\|_2$. Under this relaxed assumption we obtain the same scaling as in the convex case:

\begin{theorembox}\label{theorem_non}
Given Assumption \ref{assumption_non}, there exists a bound for the distance between the weights of the model after training ($\theta^*$) and one of the minima of the loss when training on the head solely, scaling inversely with the square root of the imbalance factor:
\begin{equation}\label{eq_theorem_non}
\textnormal{{\text{dist}}}(\theta^*,S_H)= \mathcal{O}(\frac{1}{\sqrt{\IFNew}}).
\end{equation}
For the full proof see Appendix \ref{theorem_non_proof}.
\vspace{-2.5mm}
\end{theorembox}

\subsection{Extension to Multiple Head–Tail Partitions}

Theorems \ref{theorem1} and \ref{theorem_non} assume that there is only one Head and one Tail in the dataset, which is not the case in many real-world datasets. We relax our assumption regarding the data as follows:
\begin{assumption}\label{assumption_3}
We relax Assumptions \ref{assumption_1} and \ref{assumption_non} by allowing individual classes within the respective Head or Tail sets to vary in size.
\end{assumption}

Under the relaxed assumption where class cardinalities within the Head and Tail sets may differ, we partition the classes into $n$ consecutive, class-disjoint subsets ordered by decreasing cardinality. These partitions are constructed to reduce within-partition imbalance while preserving the Head-to-Tail ordering. Theorem~\ref{theorem 2} extends the analysis of Theorems~\ref{theorem1} and~\ref{theorem_non} to multiple partitions, including the nonconvex setting where the Head objective may have multiple minimizers.

\begin{theorembox}\label{theorem 2}
Following Assumption~\ref{assumption_3}, let $\mathcal{D}=\bigcup_{i=1}^{n}\mathcal{D}^i$ be divided into $n\geq 2$ disjoint partitions, ordered such that $|\mathcal{D}^1|\geq\cdots\geq|\mathcal{D}^n|$. Define the aggregated Tail as $\mathcal{D}_T=\bigcup_{i=2}^{n}\mathcal{D}^i$ and the Head minimizer set as
$S_{\mathcal{D}^1} = \arg\min_{\theta} F_{\mathcal{D}^1}(\theta),$
where $F_A(\theta)=\mathcal{L}(A,\theta)+\mathcal{R}(\theta)$ denotes the training objective with a shared regularizer. Let $\theta^*$ minimize $F_{\mathcal{D}}$. Under the local error-bound and bounded loss-contrast conditions of Theorem~\ref{theorem_non}, applied to the Head and aggregated Tail, and assuming $\theta^*$ lies within the corresponding neighborhood, the following bound holds for fixed $n$:
\begin{align}\label{eq_theorem_2} \operatorname{dist} \left(\theta^*,S_{\mathcal{D}^1}\right)
&\leq C\sqrt{
2M\frac{|\mathcal{D}_T|}{|\mathcal{D}|}
}\leq C\sqrt{
2M(n-1)
\frac{|\mathcal{D}^2|}{|\mathcal{D}^1|}
}=
\mathcal{O}_n\left(
\frac{1}{
\sqrt{\IFNew_{\mathcal{D}^1,\mathcal{D}^2}}
}
\right),
\end{align}
where $\IFNew_{\mathcal{D}^1,\mathcal{D}^2} =|\mathcal{D}^1|/|\mathcal{D}^2|$, and $C,M>0$ are constants defined by the local error bound and loss-contrast condition, respectively.

When the Head objective has a unique minimizer $\theta^*_{\mathcal{D}^1}$, the set-valued distance reduces to $\|\theta^*-\theta^*_{\mathcal{D}^1}\|$. The formal proof is presented in Appendix~\ref{theorem_2_proof}.
\vspace{-2.5mm}
\end{theorembox}

Remark~\ref{rem:exp_decay} addresses the exponential class-frequency decay commonly used in LTR benchmarks.

\begin{remarkbox}[Exponential long-tailed datasets]
\label{rem:exp_decay}
The same inverse-square-root geometry in Theorem~\ref{theorem 2} holds for exponentially decaying class frequencies, which are commonly used to construct LTR benchmarks. The full statement and proof are provided in Appendix~\ref{app:exp_decay}.
\vspace{-2.5mm}
\end{remarkbox}

\subsection{Continual Learning for Long-Tailed Recognition}

\begin{wrapfigure}{r}{0.5\textwidth}
\vspace{-6pt}
\footnotesize
\refstepcounter{algorithm}
\label{alg:CLTR}
\hrule
\vspace{4pt}
\noindent\textbf{Algorithm \thealgorithm\quad
CLTR: Head-to-Tail Learning}

\vspace{2pt}
\hrule
\vspace{4pt}

\begin{algorithmic}[1]

    \STATE \textbf{Input:} Long-tailed dataset $\mathcal{D}$,
    initial parameters $\theta_0$, number of partitions $N$,
    CL algorithm $\mathcal{A}$

    \STATE \textbf{Output:} Final parameters $\theta^*_{HT}$

    \STATE Sort classes by cardinality:
    $|\mathcal{D}_{c_1}| \geq \cdots \geq |\mathcal{D}_{c_k}|$

    \STATE Partition the sorted classes into $N$
    consecutive, class-disjoint subsets
    $\{\mathcal{D}^{t}\}_{t=1}^{N}$,
    with
    \newline
    $R_t =
    \frac{
        \max_{\mathcal{D}_c\subseteq\mathcal{D}^{t}}
        |\mathcal{D}_c|
    }{
        \min_{\mathcal{D}_c\subseteq\mathcal{D}^{t}}
        |\mathcal{D}_c|
    }$
    (Proposition~\ref{prop:partitioning})

    \STATE Initialize CL state $S_0 \gets \emptyset$

    \STATE \textbf{// Stage 1: Head-dominated initialization}

    \STATE $\mathcal{D}_{\mathrm{init}} \gets \mathcal{D}^{1}$
    \textit{(or $\mathcal{D}$; see Appendix~\ref{ID})}

    \STATE $\theta_1 \gets
    \operatorname{Train}(\theta_0,\mathcal{D}_{\mathrm{init}})$

    \STATE $S_1 \gets
    \operatorname{InitState}_{\mathcal{A}}
    (\theta_1,\mathcal{D}^{1})$

    \STATE \textbf{// Stages 2--$N$: Learn Tail partitions}

    \FOR{$t=2$ \textbf{to} $N$}
        \STATE $(\theta_t,S_t) \gets
        \operatorname{Update}_{\mathcal{A}}
        (\theta_{t-1},\mathcal{D}^{t},S_{t-1})$
    \ENDFOR

    \STATE $\theta^*_{HT} \gets \theta_N$

    \STATE \textbf{return} $\theta^*_{HT}$

\end{algorithmic}

\vspace{3pt}
\hrule
\vspace{-6pt}
\end{wrapfigure}
Let us assume an LTR problem and a learner with a set of parameters denoted as $\theta$ (recall definition \ref{def}). Initially, the learner is trained on a highly imbalanced dataset $\cal{D}$, as shown in Fig. \ref{fig:losslandscape} right, where $\theta_i$ is the initialized model in the weight space. Owing to the larger number of Head samples in each iteration, they dominate the evolution of the gradients, resulting in a learner that performs significantly better on the Head set than on the Tail set at the end of training. This process leads the parameters to converge to $\theta^*$.

Under the respective assumptions, Theorems~\ref{theorem1} and~\ref{theorem 2} establish that the full-dataset minimizer $\theta^*$ lies within a bounded neighborhood of the Head-only solution in the convex case, or the Head minimizer set in the nonconvex case, with an inverse-square-root dependence on the corresponding imbalance factor.

This neighborhood falls in $\psi_H$ which represents an area within the weight space where the network performs well on the Head set. At this stage, the model should learn the Tail; however, if it is simply fine-tuned on the Tail (${\cal{D}}_T$),
then the parameters may move toward the Tail-only solution $\theta_T^\ast$ within the Tail-performance region $\psi_T$, potentially leaving $\psi_H$.
This phenomenon occurs in sequential learning and it is known as catastrophic forgetting. To mitigate this problem and guide the model to the intersection of $\psi_H$ and $\psi_T$ denoted as $\psi_{HT}$, where the model performs well on both Head and Tail, the Tail should be learned without forgetting the Head. To this end, we propose using the standard CL methods for sequentially learning the Tail after the Head
while mitigating catastrophic forgetting and seeking a solution $\theta_{HT}^\ast$ that performs well on both Head and Tail classes.

Following \cite{prabhu2020gdumb}, a general CL problem can be formulated as a model exposed to a stream of $N$ incoming training datasets $\mathcal{D}_{\mathcal{Y}_t}=\{(x_i,y_i)|y_i\in\mathcal{Y}_t\}$ for $1\leq t\leq N$, where $\mathcal{Y}_t$ is the corresponding set of labels. Up to the current timestep $t$, the set of labels $\bigcup_{i=1}^t \mathcal{Y}_i$ in dataset $\bigcup_{i=1}^t \mathcal{D}_{\mathcal{Y}_i}$ has been previously used in training of the network. The objective at the next timestep $t+1$ is to find a mapping $f_{\theta}~:~x\rightarrow y$ that accurately maps sample $x$ to $\bigcup_{i=1}^t \mathcal{Y}_i \cup \mathcal{Y}_{t+1}$, where $\mathcal{Y}_{t+1}$ is the set of new unseen labels in the incoming new dataset $\mathcal{D}_{\mathcal{Y}_{t+1}}=\{(x_i,y_i)|y_i\in \mathcal{Y}_{t+1}\}$. Therefore the ultimate objective of CL is to find an accurate mapping $ f_{\theta}:x\rightarrow y \quad \text{for all} \quad (x,y)\in \bigcup_{i=1}^N \mathcal{D}_{\mathcal{Y}_i}.$ Consider dataset $\mathcal{D}$ under the LTR setup (Definition \ref{def}) divided into $N$ partitions with substantially different sizes sorted based on cardinality such that $\mathcal{D} = \bigcup_{i=1}^{N} \mathcal{D}^{i}$ and $|\mathcal{D}^{i}| \gg |\mathcal{D}^{i+1}|$.

Under the conditions of Theorem~\ref{theorem 2}, the full-dataset minimizer $\theta^*$ lies within a bounded neighborhood of the minimizer set $S_{\mathcal{D}^1}$ associated with the largest partition $\mathcal{D}^1$. This proximity characterizes the Head-dominated geometry of long-tailed optimization and motivates sequentially learning the remaining partitions while preserving knowledge of the Head.
On the other hand, following Definition \ref{def}, the objective of LTR is to learn $f_{\theta}:x\rightarrow y$ for all $(x,y)\in \mathcal{D} = \bigcup_{i=1}^{N} \mathcal{D}^i$.
Hence, additional training steps are required for the model to further learn the rest of the partitions of the dataset ($\bigcup_{i=2}^{N} \mathcal{D}^i$).
Thus, if we consider each of the partitions of the LTR dataset ($\mathcal{D}^i$ for $1\leq i \leq N$) as an incoming CL dataset ($\mathcal{D}_{\mathcal{Y}_t}$ for $1\leq t \leq N$),
the learning objective of LTR can be cast as the objective of a CL problem (Proposition~\ref{proposition}). This reformulation operates at the objective level, showing that the same hypothesis class and dataset can be learned through a CL stream sorted by partition cardinality. It does \emph{not} claim equivalence of optimization trajectories, gradient dynamics, or solution geometry in general non-convex settings.
\begin{propositionbox}\label{proposition}
Let dataset $\mathcal{D}$ under the LTR setup (Definition \ref{def}) be divided into $N$ partitions sorted by cardinality such that $\mathcal{D} = \bigcup_{i=1}^{N} \mathcal{D}^{i}$, where $|\mathcal{D}^{i}| \gg |\mathcal{D}^{i+1}|$. If each partition $\mathcal{D}^i$ is treated as an incoming dataset in a continual learning stream, then
the objective of the LTR problem admits an objective-level reformulation as the objective of a CL problem
, with the goal of learning:
$f_{\theta}:x\rightarrow y \quad\ s.t. \quad\ (x,y)\in \bigcup_{t=1}^{N} \mathcal{D}_{\mathcal{Y}_t}$ and $   \mathcal{D}_{\mathcal{Y}_1}=\mathcal{D}^1,\ \mathcal{D}_{\mathcal{Y}_2}=\mathcal{D}^2,\ \dots ,\ \mathcal{D}_{\mathcal{Y}_N}=\mathcal{D}^N$.
\vspace{-2mm}
\end{propositionbox}

A corollary of Proposition~\ref{proposition} is that CL solutions can be employed to address LTR. To operationalize this insight, we propose CLTR, a simple yet effective framework that leverages off-the-shelf CL methods to solve LTR.
 The complete procedure is summarized in Algorithm~\ref{alg:CLTR}.

First, the long-tailed dataset is partitioned into $N$ consecutive, class-disjoint subsets based on class cardinality, with boundaries selected to approximately equalize the local imbalance factors following Proposition~\ref{prop:partitioning}. Starting with the largest partition (Head), each subsequent partition is treated as a new task and learned using a selected CL method $\mathcal{A}$ while preserving knowledge from previous partitions. The method-specific state $S_t$ maintains the information required by $\mathcal{A}$, such as replay samples, regularization statistics, or teacher models. This Head-to-Tail ordering ensures that the solution obtained on the first partition serves as the Head solution $\theta_H^\ast$ underlying our analysis in Proposition~\ref{proposition} and Theorem~\ref{theorem 2}, and is consistent with established LTR practice~\cite{zhou2020bbn, zhang2022self}. For $N=2$, the procedure reduces to the two-stage Head--Tail setting analyzed in Theorem~\ref{theorem_general}. We empirically demonstrate the importance of this ordering through a task-order ablation in Appendix~\ref{app:order}.

\begin{propositionbox}
\label{prop:partitioning}
For a CL algorithm $\mathcal A$, let $\eta_{\mathcal A}>0$ denote its effective forgetting coefficient.
Smaller values indicate stronger preservation of previously learned partitions.
 Let $B>0$ denote a uniform upper bound on per-class losses over the considered parameter region.
Under the exponential long-tailed class-count profile used in standard LTR benchmarks, the optimal number of partitions is given by
\begin{equation}\label{tradeoff_eq}
N^\star_{\mathcal A}
\in
\arg\min_N
\left[
2B
\frac{\IFNew^{1/(2N)}-1}{\IFNew^{1/(2N)}+1}
+
(N-1)\eta_{\mathcal A}
\right].
\end{equation}
Moreover, for any two CL algorithms $\mathcal A_i$ and $\mathcal A_j$:
$
\eta_{\mathcal A_i}<\eta_{\mathcal A_j}
\quad\Longrightarrow\quad
N^\star_{\mathcal A_i}\ge N^\star_{\mathcal A_j}.
$
Hence, stronger CL methods support a larger optimal number of partitions, while weaker CL methods should use fewer partitions.
For any fixed $N$, the optimal partition boundaries equalize the local imbalance factors:
$R_1=R_2=\cdots=R_N=\IFNew^{1/N}$,
$R_i=\frac{\max_{\mathcal{D}_c\subseteq\mathcal{D}^{i}}|\mathcal{D}_c|}{\min_{\mathcal{D}_c\subseteq\mathcal{D}^{i}}|\mathcal{D}_c|}$.
The full bias-forgetting bound and proof are given in Appendix~\ref{app:partitioning}.
\vspace{-2.5mm}
\end{propositionbox}

Proposition~\ref{prop:partitioning} provides a principled criterion for selecting the number of partitions by balancing within-partition imbalance against accumulated forgetting. In practice, the effective forgetting coefficient $\eta_{\mathcal{A}}$ can be estimated from the increase in retained-partition loss after a short pilot transition, while the observed per-class loss range on held-out data provides a proxy for $B$. The criterion can then be evaluated over a small set of candidate values of $N$. Following this principle, we use $N=2$ for more forgetting-sensitive CL methods and $N=4$ for stronger methods such as FOSTER and TPL. For fixed $N$, partition boundaries approximately equalize the local imbalance factors $R_i\approx IF^{1/N}$. We empirically evaluate sensitivity to both the number of partitions and boundary placement in Appendix~\ref{app:step}.

\subsection{General Guarantees for Effectiveness of CLTR}
\label{sec:general-guarantee}

We define the balanced LTR objective as $\mathcal{L}_{\mathrm{bal}}(\theta)=\frac{1}{2}\mathcal{L}_H(\theta)+\frac{1}{2}\mathcal{L}_T(\theta)$. Here we provide a theoretical \emph{guarantee} for the effectiveness of the proposed CLTR, demonstrating that employing any off-the-shelf CL method can address LTR. To this end, we first employ the unified and general formulation of CL solutions proposed by \cite{wang2024unified}, which covers a wide variety of off-the-shelf CL methods:
\begin{equation}\label{eq:wang-app}
\mathcal{L}_{\mathrm{CL}}(\theta)
:= \mathcal{L}_{\mathrm{task}}(\theta)
 + \alpha\, D_{\Phi}\!\big(h_{\theta}(\cdot), z\big)
 + \beta\, D_{\Psi}\!\big(\theta,\theta_H^\star\big), \:\:\:\:\:\:\:\:\:\: \alpha,\beta \ge 0
\end{equation}
where $h_\theta$ is the model, $D_{\Phi}$ and $D_{\Psi}$ are Bregman divergences generated by convex $C^1$ functions $\Phi$ and $\Psi$, and $z$ is a fixed Head reference (e.g., stored logits/predictions at $\theta_H^\star$). In CLTR stage-2, $\mathcal{L}_{\mathrm{task}}(\theta)$ is the Tail loss ($\mathcal{L}_{\mathrm{task}}\equiv \mathcal{L}_T$). The assumptions that we have regarding the CL solution on a neighborhood $\mathcal{N}\subset\mathbb{R}^d$ of $\theta_H^\star$ are as follows:

\begin{assumption}\label{ass:cltr}
\textbf{(Head smoothness)} $\mathcal{L}_H$ is $L_H^{\mathrm{sm}}$-smooth at $\theta_H^\star$:
for all $\theta\in\mathcal{N}$,
\begin{equation}\label{eq:head-smooth}
\mathcal{L}_H(\theta)\ \le\ \mathcal{L}_H(\theta_H^\star)\ +\ \tfrac{L_H^{\mathrm{sm}}}{2}\,\|\theta-\theta_H^\star\|^2,
\;\;\forall \theta\in\mathcal{N}.
\end{equation}

\textbf{(Parameter divergence curvature)} $D_{\Psi}(\cdot,\theta_H^\star)$ is $m_{\Psi}$-strongly convex:
\begin{equation}\label{eq:psi-strong}
D_{\Psi}(\theta,\theta_H^\star)\ \ge\ \tfrac{m_{\Psi}}{2}\,\|\theta-\theta_H^\star\|^2,
\;\;\forall \theta\in\mathcal{N}.
\end{equation}

\textbf{(Output calibration)} There exists $c_{\Phi}>0$ and a fixed Head target $z=z_H^\star$ such that:
\begin{equation}\label{eq:phi-calib}
\mathcal{L}_H(\theta)-\mathcal{L}_H(\theta_H^\star)\ \le\ c_{\Phi}\,D_{\Phi}\!\big(h_{\theta}(\cdot), z_H^\star\big),
\;\;\forall \theta\in\mathcal{N}.
\end{equation}
\end{assumption}

Here we explain \noindent\emph{why Assumption \ref{ass:cltr} is mild.}
(i) Eq. \ref{eq:head-smooth} is a standard local smoothness bound; it always holds with some finite $L_H^{\mathrm{sm}}$ if $\mathcal{L}_H$ has locally Lipschitz gradient.
(ii) Eq. \ref{eq:psi-strong} holds for quadratic/EWC-type $\Psi(\theta)=\tfrac12(\theta-\theta_H^\star)^\top H_H(\theta-\theta_H^\star)$ with a damped positive-definite $H_H\succeq m_{\Psi}I$ (standard in practice). It also holds on piecewise strongly convex basins (e.g., deep networks with ReLU + weight decay \cite{milne2019piecewise}).
(iii) Eq. \ref{eq:phi-calib} is the usual distillation calibration (e.g., cross-entropy controlled by KL between predictions).

\noindent
We now propose the general guarantee for the effectiveness of CLTR through the following theorem.

\begin{theorembox}\label{theorem_general}
Suppose Assumption~\ref{ass:cltr} holds on $\mathcal{N}$. Define the thresholds
\begin{equation}
\beta_{\min}\ :=\ \frac{L_H^{\mathrm{sm}}}{2\,m_{\Psi}},
\qquad
\alpha_{\min}\ :=\ \frac{c_{\Phi}}{2},
\end{equation}
where $L_H^{\mathrm{sm}}$ is local smoothness constant of $\mathcal{L}_H$ at $\theta_H^\star$, $m_{\Psi}$ is strong-convexity constant of $D_{\Psi}(\cdot,\theta_H^\star)$, and $c_{\Phi}$ is calibration constant relating $\mathcal{L}_H$ to $D_{\Phi}$. If either $\beta\ge \beta_{\min}$ \textup{(Parameter route)} or $\alpha\ge \alpha_{\min}$ \textup{(Output route)}, then for any $\theta\in\mathcal{N}$,
\begin{equation}\label{eq:dominance}
 \mathcal{L}_{\mathrm{bal}}(\theta)\ \le\ \mathcal{L}_{\mathrm{CL}}(\theta)\ +\ \tfrac12\,\mathcal{L}_H(\theta_H^\star).
\end{equation}
In words: if either $\beta\ge \beta_{\min}$ or $\alpha\ge \alpha_{\min}$, then the CL objective $\mathcal{L}_{\mathrm{CL}}(\theta)$ upper-bounds the balanced loss $\mathcal{L}_{\mathrm{bal}}(\theta)$ up to an additive $\theta$-independent constant. For the full proof see Appendix~\ref{proof_theorem_general}.
\vspace{-2.5mm}
\end{theorembox}

Under the local conditions of Theorem~\ref{theorem_general}, the unified CL objective (Eq.~\ref{eq:wang-app}) upper-bounds the balanced LTR loss up to the additive constant $\frac{1}{2}\mathcal{L}_H(\theta_H^\ast)$, provided that either $\beta\geq\beta_{\min}$ or $\alpha\geq\alpha_{\min}$. As discussed in Appendix~\ref{app:min}, empirical estimates suggest that these thresholds can be satisfied by standard CL regularization strengths, often by orders of magnitude in the evaluated settings. Importantly, the additive term is determined by the residual Head loss and is small when Head training achieves low empirical loss. Therefore, if CL optimization yields $\hat{\theta}\in\mathcal{N}$ with a low $\mathcal{L}_{\mathrm{CL}}(\hat{\theta})$, the theorem certifies a correspondingly low balanced LTR loss. Moreover, decreasing $\mathcal{L}_{\mathrm{CL}}$ decreases the corresponding upper-bound certificate along iterates within $\mathcal{N}$, although this does not guarantee a monotonic decrease in the balanced loss itself. Thus, the significance of Theorem~\ref{theorem_general} lies in controlling the balanced loss through a decreasing surrogate, without requiring the upper bound to be numerically tight.

Furthermore, the CL objective minimizes the full Tail-task loss $\mathcal{L}_{task}=\mathcal{L}_T$, rather than the weighted Tail component $\frac{1}{2}\mathcal{L}_T$ of the balanced objective. This gives additional emphasis to Tail learning while the CL regularizers aim to preserve previously acquired Head knowledge, counteracting the Head bias induced during initial training. Hence, CLTR provides a principled surrogate for controlling the balanced LTR loss under the stated conditions. In the special convex setting analyzed in Appendix~\ref{eq_convex}, a particular choice of the CL objective components and coefficients yields an objective that coincides with the balanced loss up to a $\theta$-independent constant, thereby sharing identical minimizers.

\section{Experiments}\label{sec:exp}
\textbf{Datasets.}
First, we use the \textbf{MNIST-LT} \cite{lecun1998gradient} toy dataset with different $\IFNew$ values to study the behavior of the upper bound (Eq.~\ref{eq_theorem}) and its compliance with our Theorem \ref{theorem1}. Then we employed \textbf{CIFAR100-LT} \cite{cao2019learning} to numerically verify Theorem \ref{theorem_non}. Next, to evaluate the performance of CLTR in addressing the LTR problem, we employ three widely used LTR datasets: \textbf{CIFAR100-LT}, \textbf{CIFAR10-LT} \cite{cao2019learning}, and  \textbf{ImageNet-LT} \cite{liu2019large}. These datasets represent long-tailed versions of the original CIFAR100, CIFAR10, and ImageNet datasets, maintaining the same number of classes while the number of samples in each class decreases exponentially according to the $\IFNew$, where the first class has the maximum number of samples and the last class contains the least number of samples, as illustrated in Appendix \ref{app_dataset}. We also demonstrate the efficacy of CLTR in addressing LT-CIL using the CIFAR100-LT benchmark (Appendix \ref{app:cil}).
Finally, to further highlight the benefits of using CLTR, we carry out additional experiments using the naturally skewed \textbf{Caltech256} dataset \cite{griffin2007caltech} (Appendix \ref{app:caltech}).

\textbf{Implementation Details.}
Following the experimental setup of \cite{alshammari2022long}
and \cite{fu2022long}, we use ResNet-32 \cite{he2016deep}, ResNeXt-50 \cite{xie2017aggregated}, and CLIP ViT-B/32 \cite{radford2021learning} for CIFAR benchmarks, ImageNet benchmark, and foundation model experiments, respectively. The LTR methods selected for comparison are state-of-the-art solutions in the area. We also employ various standard and state-of-the-art CL methods in CLTR, namely LwF \cite{li2017learning}, EWC \cite{kirkpatrick2017overcoming}, Modified EWC \cite{molahasani2023continual}, GPM \cite{saha2021gradient}, FOSTER \cite{wang2022foster}, SGP \cite{Saha_Roy_2023}, TPL \cite{lin2023class}, and DualPrompt \cite{wang2022dualprompt}. Following proposition \ref{prop:partitioning}, we divided the dataset into 2 partitions for LwF, EWC, Modified EWC, GPM, and SGP
\begin{wrapfigure}{r}{0.4\linewidth}
\begin{minipage}[t]{\linewidth}

    \centering

    \includegraphics[width=1\linewidth]{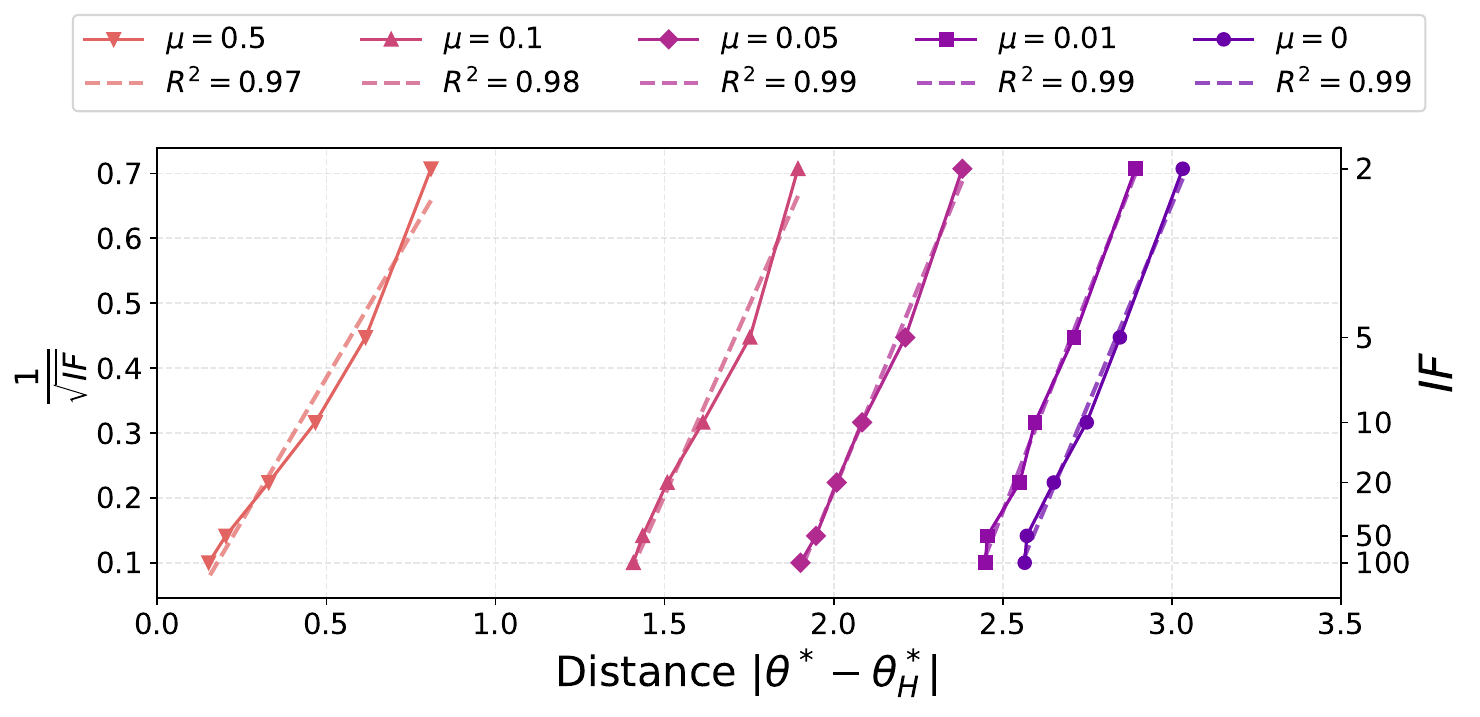}

\end{minipage}

\vspace{-5mm}

\begin{minipage}[t]{\linewidth}

    \centering

    \includegraphics[width = 1\linewidth]{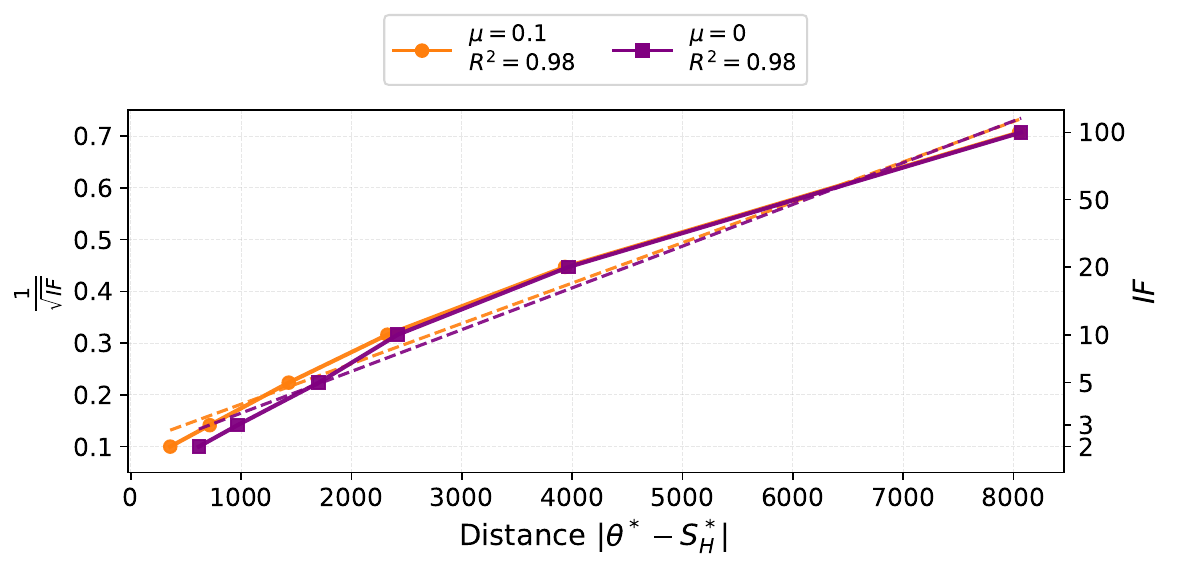}

\end{minipage}
\caption{Empirical validation for Theorems \ref{theorem1} and \ref{theorem_non}. Top: log-reg on MNIST-LT with varying imbalance factors and $L^2$ strengths $\mu$. Bottom: the same analysis for ResNet-18 on CIFAR-100, with and without weight decay.}
\label{fig_all}
\vspace{-7mm}
\end{wrapfigure}
and 4 partitions for FOSTER and TPL and the boundaries are determined by our theory. All trainings were conducted using an NVIDIA RTX 3090 GPU with 24GB VRAM. More details on the implementation specifics are provided in Appendix \ref{ID}.

\textbf{Evaluation.}
For the LTR datasets (MNIST-LT, CIFAR100-LT, CIFAR10-LT, ImageNet-LT), we first train the model on the long-tailed imbalanced training set and then evaluate it on the balanced test set, following the evaluation protocol of \cite{alshammari2022long}. All reported values represent classification accuracy. The results of our proposed approach are \colorbox{lightblue}{highlighted} in the Tables.

\begin{table*}[ht!]
\vspace{-2mm}
\caption{LTR benchmarks for (left) CIFAR100-LT, (middle) CIFAR10-LT, and (right) ImageNet-LT. \textbf{Bold} and \underline{underlined} values denote the best and second-best results in each column.}\label{tab:table1}
\scriptsize
\setlength{\tabcolsep}{3pt}
  \centering
  \begin{minipage}[t]{.3\textwidth}
    \centering
\resizebox{1.09\linewidth}{!}{
\begin{tabular}{llll}
\hline
\multicolumn{1}{c}{\multirow{2}{*}{Model}} & \multicolumn{3}{c}{$\IFNew$}                           \\
\multicolumn{1}{c}{}                       & \multicolumn{1}{c}{100}  & \multicolumn{1}{c}{50}   & 10    \\ \hline\hline
\multicolumn{1}{l}{Baseline \cite{cui2019class}}               & 38.3                    & 43.9                    & 55.7 \\
\multicolumn{1}{l}{Baseline + CB \cite{cui2019class}}          & 39.6                     & 45.3                    & 58.0 \\
\multicolumn{1}{l}{Focal loss \cite{lin2017focal}}                  & 38.4                    & 44.3                    & 55.8 \\
\multicolumn{1}{l}{Focal+CB \cite{cui2019class}}               & 39.6                     & 45.2                    & 58.0 \\
\multicolumn{1}{l}{$\tau$-norm \cite{38kang2019decoupling}}                 & 47.7                    & 52.5                    & 63.8  \\
\multicolumn{1}{l}{LDAM-DRW \cite{cao2019learning}}               & 42.0                    & 46.6                    & 58.7 \\
\multicolumn{1}{l}{BBN  \cite{zhou2020bbn}}                    & 42.6                    & 47.0                    & 59.1 \\
\multicolumn{1}{l}{LogitAdjust \cite{54menon2020long}}             & 42.0                    & 47.0                    & 57.7 \\
\multicolumn{1}{l}{LDAM+SSP \cite{78yang2020rethinking}}               & 43.4                    & 47.1                    & 58.9 \\
\multicolumn{1}{l}{De-confound \cite{tang2020long}}            & 44.1                     & 50.3                   & 59.6  \\
\multicolumn{1}{l}{SSD \cite{ li2021self}}                    & 46.0                       & 50.5                     & 62.3  \\
\multicolumn{1}{l}{DiVE \cite{32he2021distilling}}                   & 45.4                    & 51.1                    & 62.0    \\
\multicolumn{1}{l}{DRO-LT \cite{65samuel2021distributional}}                 & 47.3                    & 57.6                    & 63.4 \\
\multicolumn{1}{l}{WD \cite{alshammari2022long}}                     & 46.0 & 52.7 & 66.1 \\
\multicolumn{1}{l}{WD+Max \cite{alshammari2022long}}              & \textbf{53.4} & \underline{57.7} & \textbf{68.7} \\
\multicolumn{1}{l}{ABL \cite{jin2023optimal}}              & 46.8 & 52.1 & 61.8 \\
\multicolumn{1}{l}{BS + BSD \cite{ren2024balanced}}              & 50.6 & 55.4 &  64.0 \\
\multicolumn{1}{l}{PHASE \cite{li2025phase}}              & 49.1 & 52.8 &  61.2 \\
\rowcolor{lightblue}
\multicolumn{1}{l}{CLTR (LwF)}                    & 45.1               & 49.3     &      58.7 \\\rowcolor{lightblue}
\multicolumn{1}{l}{CLTR (EWC)}                    & 44.4     & 50.3     &     58.8  \\\rowcolor{lightblue}
\multicolumn{1}{l}{CLTR (Modified EWC)}            & 45.9                     & 51.0    &      60.7 \\\rowcolor{lightblue}
\multicolumn{1}{l}{CLTR (GPM)}                   & 48.3                    & 54.7    &      64.7         \\\rowcolor{lightblue}
\multicolumn{1}{l}{CLTR (FOSTER)}                    & 48.7  & 54.4  & 63.6 \\\rowcolor{lightblue}
\multicolumn{1}{l}{CLTR (SGP)}                    & \underline{50.7} & \textbf{58.0} & \underline{67.2}   \\\rowcolor{lightblue}
\multicolumn{1}{l}{CLTR (TPL)}                    & 48.4  & 54.0  & 62.1       \\\hline

\end{tabular}}

  \end{minipage}%
    \hfill
    \hspace{0.6cm}
  \begin{minipage}[t]{.3\textwidth}
    \centering
\resizebox{0.985\linewidth}{!}{\begin{tabular}{lcc}
\hline
\multicolumn{1}{c}{\multirow{2}{*}{Model}} & \multicolumn{2}{c}{$\IFNew$}                           \\
\multicolumn{1}{c}{}                       & \multicolumn{1}{c}{100}  & \multicolumn{1}{c}{50}       \\ \hline\hline
\multicolumn{1}{l}{Baseline \cite{cui2019class}}               & 69.8                    & 75.2                     \\
\multicolumn{1}{l}{Baseline + CB \cite{cui2019class}}          & 74.7                     & 79.3                  \\
Focal loss\cite{lin2017focal}                                          & 70.4                & 75.3               \\
PG Re-sampling \cite{13cui2018large}                                     & 67.1                & 75.0                 \\
Focal+CB\cite{cui2019class}                                           & 74.6                & 79.3               \\
LDAM-DRW\cite{cao2019learning}                                            & 77.0                  & 79.3               \\
BBN  \cite{zhou2020bbn}                                                & 79.8                & 82.2               \\
Manifold mixup \cite{cui2019class}                                    & 73.0                  & 78.1                 \\
CBA-LDAM \cite{cui2019class}                                            & 80.3                  & 82.2               \\
ELF (LDAM)+DRW \cite{cui2019class}                                      & 78.1                & 82.4               \\
De-confound \cite{tang2020long}                                         & 80.6                & 83.6               \\
Hybrid-SC \cite{105wang2021contrastive}             & 81.4                & 85.4               \\
MiSLAS \cite{131zhong2021improving}                 & 82.1                & 85.7               \\
BCL \cite{zhu2022balanced}                          & \underline{84.3 }               & \underline{87.2}               \\
WD+Max \cite{alshammari2022long} & 81.4& 85.2 \\
FNO \cite{kim2024fixed}                          & 82.6                & 85.1              \\
\multicolumn{1}{l}{BS + BSD \cite{ren2024balanced}}              & 79.7 & 82.7 \\
\multicolumn{1}{l}{PHASE \cite{li2025phase}}              & 78.9 & 82.7  \\
\rowcolor{lightblue}
\multicolumn{1}{l}{CLTR (LwF)
}                    & 76.3    & 78.6         \\\rowcolor{lightblue}
\multicolumn{1}{l}{CLTR (EWC)
}                    & 75.1  & 80.1         \\\rowcolor{lightblue}
\multicolumn{1}{l}{CLTR (Modified EWC)
}            & 77.8  &81.3         \\\rowcolor{lightblue}
\multicolumn{1}{l}{CLTR (GPM)
}                    & 81.2 & 84.8         \\ \rowcolor{lightblue}
\multicolumn{1}{l}{CLTR (FOSTER)
}  & 81.7  & 85.9  \\\rowcolor{lightblue}
\multicolumn{1}{l}{CLTR (SGP)
}                    & 83.0 &  85.5  \\\rowcolor{lightblue}
\multicolumn{1}{l}{CLTR (TPL)}                    & \textbf{84.7}  & \textbf{87.6}       \\\hline

\end{tabular}}
  \end{minipage}%
  \hfill
  \begin{minipage}[t]{.3\textwidth}
    \centering
\resizebox{0.91
\linewidth}{!}{\begin{tabular}{lc}
\hline
\multicolumn{1}{c}{\multirow{2}{*}{Model}} & \multicolumn{1}{c}{Top-1}                           \\
\multicolumn{1}{c}{}                       & \multicolumn{1}{c}{accuracy}      \\ \hline\hline
Baseline \cite{cui2019class}              & 44.4 \\
Baseline + CB \cite{cui2019class}         & 33.2 \\
KD \cite{33hinton2015distilling}            & 35.8 \\
Focal \cite{lin2017focal}                 & 30.5 \\
SR Re-sampling \cite{mahajan2018exploring}  & 46.8 \\
OLTR \cite{liu2019large}                     & 35.6 \\
cRT \cite{38kang2019decoupling}             & 49.6 \\
$\tau$-norm \cite{38kang2019decoupling}     & 49.4 \\
LFME \cite{xiang2020learning}               & 37.5 \\
De-confound \cite{tang2020long}           & 51.8 \\
Seesaw Loss \cite{wang2021seesaw}           & 50.4 \\
DiVE \cite{32he2021distilling}              & 53.1 \\
DisAlign \cite{zhang2021distribution}       & 52.9 \\
WD \cite{alshammari2022long}                & 48.6 \\
WD+Max \cite{alshammari2022long}            & \textbf{53.9} \\
ABL \cite{jin2023optimal}            & 53.2 \\
EWB-FDR \cite{hasegawa2024exploring}            & 53.2 \\
SynBalance (n=100) \cite{jiang2025synbalance}            & \underline{53.6} \\
\rowcolor{lightblue}
CLTR (LwF)
& 47.6\\\rowcolor{lightblue}
CLTR (EWC)
& 48.9 \\\rowcolor{lightblue}
CLTR (Modified EWC)
& 49.1\\\rowcolor{lightblue}
CLTR (GPM)
& 51.7\\\rowcolor{lightblue}
CLTR (FOSTER)
&  52.7    \\\rowcolor{lightblue}
CLTR (SGP)
& 53.2 \\ \rowcolor{lightblue}
CLTR (TPL)
&  \textbf{53.9}    \\ \hline

\end{tabular}} \end{minipage}%
\vspace{-4mm}
\end{table*}

\textbf{Empirical support for Theorems \ref{theorem1} and \ref{theorem_non}.}
 Prior works have shown that complementing theoretical guarantees with focused empirical verification can yield valuable insights into the practical relevance of the theory \cite{pourpanah2025federated,lee2019wide,wang2024forgetting}. Motivated by this, our experiments are structured to test the key behaviors and scaling predicted by our analysis.
To evaluate the validity of Theorem \ref{theorem1} on the upper bound for the distance between the learner's weights when trained on $\mathcal{D}$ and $\mathcal{D}_H$ ($\|\theta^* - \theta^*_H\|$), we first train a logistic regression model with cross-entropy loss on MNIST-LT with varying $\IFNew$ to ensure the convexity of the loss function \cite{rychlik2019proof}. We use different values for the regularizer coefficient as well ($\mu$).
Then, we compute the Euclidean distance between the two sets of weights and plot them against $\frac{1}{\sqrt{\IFNew}}$, as shown in Fig. \ref{fig_all} top. A regression line is fitted to the data, and the $R^2$ score is calculated. Consistent with Eq. \ref{eq_theorem}, the difference between the weights is bounded and scales linearly with $\frac{1}{\sqrt{\IFNew}}$ with high confidence ($R^2 \geq 0.97$).
We then extend this experiment to a deeper architecture, ResNet-18 (a feedforward deep network), trained on the CIFAR-100 dataset. We measure the distance $\|\theta^* - \theta^*_H\|$ with and without weight decay and report the results in Fig.~\ref{fig_all} bottom. In both settings, the fitted line achieves $R^2 \geq 0.98$, indicating that the bound scales linearly with $\frac{1}{\sqrt{\IFNew}}$ even in deep networks, where the loss is not convex, as predicted by Theorem~\ref{theorem_non}.

\textbf{Empirical support for Theorem~\ref{theorem_general}.} We compare the performance of CLTR with existing state-of-the-art LTR solutions on three benchmarks, CIFAR100-LT, CIFAR10-LT, and ImageNet-LT, as presented in Table~\ref{tab:table1}. We also include two baselines trained on imbalanced data, with and without a class-balanced loss. Beyond classification accuracy, we directly examine the upper-bound inequality of Theorem~\ref{theorem_general} throughout Tail training on CIFAR100-LT. Across 33 recorded checkpoints per method, the inequality holds for both CLTR (LwF) and CLTR (EWC), whereas naive sequential training violates it at every checkpoint. %
These results demonstrate that the predicted upper-bound relationship holds throughout the evaluated CLTR training trajectories, while consistently failing under naive sequential training. Full results are provided in Appendix~\ref{app:min}.

Following the prior works such as \cite{alshammari2022long}, we avoid direct comparisons with solutions with ``bells and whistles'' such as RIDE \cite{b1wang2020long}, ACE \cite{b2cai2021ace}, SSD \cite{ li2021self}, and PaCo \cite{17cui2021parametric}, which employ aggressive data augmentations, ensemble learning, multi-expert architectures, and self-supervised pretraining. For completeness, we report these methods in Appendix~\ref{app:bells_whistles} while keeping them separate from the main comparison due to protocol differences. It is worth mentioning that some previous LTR solutions like BBN \cite{zhou2020bbn} learn the Head and Tail separately in a multi-stage manner. They rely on various techniques to prevent performance loss on the Head while learning the Tail. However, unlike these methods, our approach only uses one model through the entire training process, and the results demonstrate that employing standard CL methods yields strong performance in the LTR benchmarks. We further demonstrate the strong performance of CLTR on the more challenging long-tailed class-incremental learning (LT-CIL) benchmarks and on the naturally skewed Caltech256 dataset in Appendices~\ref{app:cil} and~\ref{app:caltech}, respectively. We also provide additional discussion on the impact of the number of incremental steps (Appendix~\ref{app:step}), replay memory size (Appendix~\ref{app:replay}), runtime (Appendix~\ref{app:run}), weight imbalance (Appendix~\ref{app:imb}), backward/forward transfer as well as catastrophic forgetting (Appendix~\ref{app:back}), and CLTR performance on more advanced backbones (Appendix~\ref{app:vit}) to better elucidate the mechanisms underlying CLTR.
We further evaluate CLTR on the 20 Newsgroups-LT text classification benchmark, providing preliminary evidence that the framework extends beyond vision, with improved Tail and balanced accuracy over joint long-tailed training (Appendix~\ref{app:nlp}).

\begin{wraptable}{r}{0.35\linewidth}
\caption{Foundation-model adaptation on CIFAR100-LT. Bold denotes the best result without additional semantic supervision. Gray rows are contextual, non-protocol-identical comparisons.}
\label{tab:clip_ltr}
\scriptsize
\setlength{\tabcolsep}{4pt}
\centering
\resizebox{\linewidth}{!}{
\begin{tabular}{lccc}
\hline
\multirow{2}{*}{Method} & \multicolumn{3}{c}{IF} \\
& 100 & 50 & 10\\
\hline\hline
Zero-shot CLIP & 64.5 & 64.5 & 64.5\\
Linear probe & 60.0 & 67.2 & 75.6\\
Full fine-tuning & 73.4 & 74.8 & 81.0\\
Prompt learning & 60.4 & 63.1 & 66.9\\
Naive sequential & 48.0 & 51.2 & 53.1\\
\rowcolor{lightblue}
CLTR (DualPrompt) & \textbf{79.2} & \textbf{81.9} &  \textbf{84.0}\\
\hline
\multicolumn{4}{c}{\textcolor{gray}{\textit{External semantic supervision}}} \\
\textcolor{gray}{BALLAD \cite{ma2021simple}} & \textcolor{gray}{77.8} & \textcolor{gray}{NR} & \textcolor{gray}{NR}\\
\textcolor{gray}{MMS \cite{wang2023margin}} & \textcolor{gray}{75.2} & \textcolor{gray}{77.5} & \textcolor{gray}{82.0}\\
\textcolor{gray}{TASA \cite{li2025tasa}} & \textcolor{gray}{79.4} & \textcolor{gray}{80.6} & \textcolor{gray}{84.4}\\
\hline
\end{tabular}
}
\end{wraptable}
\textbf{Adapting Foundation Models to Long-Tailed Data.}
Modern LTR increasingly involves adapting pretrained foundation models to scarce and imbalanced data \cite{li2025tasa}. This setting is practically important because many real-world applications have access to strong pretrained foundation models but only heavily skewed task-specific data. CLTR is naturally suited to this regime. Table~\ref{tab:clip_ltr} evaluates this setting on CIFAR100-LT with a CLIP backbone. Standard adaptation baselines expose complementary failures: zero-shot CLIP cannot exploit task data, linear probing and prompt learning have limited capacity to correct Tail bias, full fine-tuning remains Head-biased under imbalanced updates, and naive sequential training catastrophically forgets earlier partitions. CLTR (DualPrompt) directly addresses this problem through employing CL methods (details in Appendix \ref{ID}) and achieves the best performance without additional semantic supervision, improving over full fine-tuning by $5.8$, $7.1$, and $3.0$ points. Gray rows report specialized VLM-LTR methods using external semantic supervision or dedicated cross-modal modules only for context; despite this stricter setting, CLTR is competitive with TASA at all $IF$ values.

\textbf{Limitations.} We identify the following limitations:
\textbf{Theoretical assumptions.} Theorems \ref{theorem1}--\ref{theorem 2} characterize parameter-space distance from the Head solution under partition-based simplifications of the cardinality distribution. While Assumption \ref{assumption_3} accommodates exponential decay via recursive partitioning, a closed-form treatment of fully continuous cardinality distributions remains open.
\textbf{Stability--plasticity completeness.} We characterize sufficient conditions on $\alpha$, $\beta$ for the upper-bound to hold; a necessary characterization of the optimal stability-plasticity trade-off in the LTR regime is left to future work.

\textbf{Scope.} CLTR is primarily evaluated on image classification benchmarks, including CIFAR100-LT, CIFAR10-LT, ImageNet-LT, Caltech256, and CIFAR100-LT-CIL. Our experiments with a CLIP backbone further demonstrate applicability to vision-language pretrained foundation models, while the 20 Newsgroups-LT experiment provides preliminary evidence of generalization to text classification. However, joint image-text adaptation under long-tailed distributions, as well as broader evaluation on NLP, detection, segmentation, and regression tasks, remains future work.
\textbf{Adaptive LTR-aware CL.} CLTR deliberately leverages off-the-shelf CL methods rather than introducing a new optimizer; its novelty lies in the theoretical LTR-to-CL connection and partitioning framework. While Proposition~\ref{prop:partitioning} provides a criterion for selecting a fixed number of partitions, a promising direction is to develop an LTR-aware CL method that estimates forgetting during training and dynamically adjusts partition boundaries and Head-preservation strength at each transition.

\section{Conclusion}\label{sec:conc}
We considered Long-Tailed Recognition (LTR) through the lens of Continual Learning (CL). We showed that, under the stated assumptions, optimizing a long-tailed objective yields parameters within an $\mathcal{O}(1/\sqrt{\text{IF}})$ neighborhood of the Head-only solution or minimizer set, in both convex and nonconvex settings. Building on this geometric insight, we established an objective-level reformulation of LTR as a sequential learning problem and proved that, under explicit local conditions, a unified CL objective upper-bounds the balanced LTR loss up to a constant determined by the residual Head loss. These findings provide a principled theoretical connection between LTR and CL. Motivated by this connection, we introduced CLTR, a framework that leverages off-the-shelf CL methods to incorporate Tail classes while preserving previously acquired Head knowledge. Extensive experiments on established LTR benchmarks, including foundation-model adaptation, demonstrated the effectiveness of this approach. We hope that viewing LTR through the lens of CL will inspire new algorithms and theoretical analyses that jointly advance both fields. Broader impact is discussed in Appendix~\ref{app:broad}.

\begin{ack}
We would like to thank Geotab Inc., the City of Kingston, and NSERC for supporting this work. The authors declare no competing  interests.
\end{ack}
\bibliography{ref}
\bibliographystyle{IEEEbib}

\newpage
\appendix
\onecolumn
\setcounter{table}{0}
\renewcommand{\thetable}{A\arabic{table}}
\setcounter{figure}{0}
\renewcommand{\thefigure}{A\arabic{figure}}
\setcounter{algorithm}{0}
\renewcommand{\thealgorithm}{A\arabic{algorithm}}

\newtheorem*{restatedtheorem}{Theorem}

\newcommand{\restatetheorem}[2]{%
  \begin{restatedtheorem}[Theorem \ref{#1} (restated)]%
  #2%
  \end{restatedtheorem}%
}

\newtheorem*{restatedremark}{Remark}

\newcommand{\restateremark}[2]{%
  \begin{restatedremark}[Remark \ref{#1} (Full version)]%
  #2%
  \end{restatedremark}%
}

\appendix
\section*{Appendix}

\section{Related Works}\label{app:related_works}
\subsection{Long-Tailed Recognition}
LTR methods broadly fall into: (i) \emph{data distribution re-balancing}, (ii) \emph{class-balanced losses}, and (iii) \emph{Head-to-Tail transfer learning} \cite{38kang2019decoupling}. Re-balancing includes Tail over-sampling \cite{chawla2002smote,han2005borderline}, Head under-sampling \cite{drummond2003c4}, class-balanced sampling \cite{shen2016relay,mahajan2018exploring}, and dynamic label distribution shift \cite{jin2023optimal}. Class-balanced losses reweight samples using class statistics \cite{cao2019learning,cui2019class,huang2019deep}, focal loss \cite{lin2017focal}, or Bayesian uncertainty \cite{khan2019striking}. Transfer learning leverages Head features to improve Tail performance \cite{yin2019feature, liu2019large, ren2024balanced}.
Vision-language models such as CLIP demonstrate promising performance in open-vocabulary settings \cite{radford2021learning}. However, when adapted to downstream tasks with highly skewed data, they exhibit a similar bias to conventional models: strong adaptation to head classes but poor performance on tail classes. To address this issue, recent studies have proposed methods for adapting vision-language models to the LTR regime. BALLAD introduces a simple two-stage CLIP-based framework that first continues contrastive pretraining on imbalanced data and then applies a balanced linear adapter to improve tail-class performance \cite{ma2021simple}. MMS aligns visual features with their corresponding text prototypes by adding class-aware prototype-distance margins to the softmax logits, thereby improving tail-class discrimination \cite{wang2023margin}. TASA further combines LLM-based semantic prototypes, dual-space image-text fusion, and MMS alignment to improve long-tailed vision-language classification across both head and tail classes \cite{li2025tasa}. However, these methods rely on external semantic supervision, such as text prototypes or LLM-generated descriptions, leaving the problem of adapting foundation models to LTR using only labeled data largely unaddressed.
Recently, multi-stage Head/Tail training has emerged as an effective LTR strategy: BBN improves representation learning \cite{zhou2020bbn}, while RIDE reduces model variance via diverse experts \cite{b1wang2020long}.  These methods implicitly adopt sequential training, but rely on heuristic LTR-specific architectures. In contrast, we provide a theoretical unification explaining \emph{why} sequential methods work: we show such specialized methods are subsets of a broader phenomenon, and by framing LTR as a CL problem, we enable \emph{any} off-the-shelf CL algorithm to address LTR without complex, LTR-specific architectural modifications. This also allows us to leverage a rich CL literature that, unlike existing multi-stage LTR solutions, tends to use a single network throughout training.

Although many works study LTR empirically, very few approach it theoretically \cite{ye2021procrustean,francazi2022theoretical}. These works show the Head is learned faster than the Tail, focusing on training dynamics; in contrast, our theory studies the \emph{convergence point} in LTR and unifies \emph{why} sequential strategies work, establishing that specialized LTR methods are subsets of this broader phenomenon.

\subsection{Continual Learning}
CL methods are commonly categorized as \emph{expansion-based}, \emph{regularization-based}, and \emph{memory-based}. Expansion-based methods allocate task-specific parameters \cite{sarwar2019incremental,li2019learn,yoon2019scalable}, but can grow unbounded as tasks accumulate; FOSTER mitigates this with a two-stage expand-then-compress paradigm \cite{wang2022foster}. Regularization-based methods constrain changes to important parameters via loss penalties \cite{saha2021gradient,saha2021space,farajtabar2020orthogonal,kirkpatrick2017overcoming,li2017learning}. Memory-based approaches use replay buffers to reduce forgetting \cite{riemer2018learning,chaudhry2018efficient,shim2021online}. For class-incremental learning without test-time task IDs, TPL predicts task IDs via likelihood ratios using replay data and task-specific models within a shared network \cite{lin2023class}. More recently, gradient surgery projects new-task gradients onto directions orthogonal to previous tasks to limit interference \cite{saha2021gradient, Saha_Roy_2023}, achieving state-of-the-art CL performance.

\newpage
\section{Notation}\label{app:notation}
Here, we summarize the key notations used throughout the paper. A few symbols are overloaded across sections; both meanings are recorded where relevant.
\begin{itemize}
\item $|\cdot|$ : Cardinality of a set (number of elements).
\item $\| \cdot \|$ : Euclidean ($L_2$) norm of a vector.
\item $\mathcal{O}(\cdot)$ : Asymptotic upper bound (Big-O).
\item $\mathrm{dist}(\theta,S)$ (also written $\textnormal{dist}(\theta,S)$) : Euclidean distance from a point $\theta$ to a set $S$ (e.g., $\mathrm{dist}(\theta^*,S_H)$ in Theorem~\ref{theorem_non}).

\item $\mathbb{R}^d$, $d$ : Input space and its dimension; each training input is written $\x_i\in\mathbb{R}^d$ (boldface $\x$ denotes a vector-valued input).
\item $y_i$ : Scalar class label of the $i^{\text{th}}$ sample; labels take values in $\{1,\ldots,k\}$ when the model has $k$ classes (in partition arguments one often writes $k=k_H+k_T$).
\item $f_{\theta}$ : Classifier/hypothesis used in mapping statements such as $f_{\theta}:x\mapsto y$; $h_{\theta}$ denotes the same model when emphasizing logits/outputs for Bregman terms.

\item $\mathcal{D} = \{(\x_i, y_i)_{i=1}^n\}$ : Training dataset with $n=|\mathcal{D}|$ samples (throughout the main text, $n$ denotes sample count).
\item $\mathcal{D}_c = \{(\x_i, y_i) \in \mathcal{D} \mid y_i = c\}$ : Samples of class $c$; $|\mathcal{D}_c|$ is the per-class training count.
\item $c_k$ : Number of Head classes (cutoff label index for Head vs.\ Tail).
\item $\mathcal{D}_H=\{(\x_i,y_i)\in \mathcal{D} \mid y_i\le c_k\}$ : Head subset.
\item $\mathcal{D}_T=\{(\x_i,y_i)\in \mathcal{D} \mid y_i>c_k\}$ : Tail subset.
\item $\mathcal{D}^i$ (or $\mathcal{D}^{i}$) : The $i^{\text{th}}$ cardinality-sorted partition when $\mathcal{D}$ is split into multiple subsets.
\item $N$ : Number of cardinality-sorted partitions in the LTR-as-CL view (Proposition~\ref{proposition}, Algorithm~\ref{alg:CLTR}); in Theorem~\ref{theorem 2} and its proof the same integer is denoted by lowercase $n$---\emph{not} the sample count $|\mathcal{D}|$.
\item $\mathcal{D}_{\mathcal{Y}_t}=\{(\x_i,y_i)\in\mathcal{D}\mid y_i\in\mathcal{Y}_t\}$ : Samples whose labels fall in label set $\mathcal{Y}_t$ at CL timestep $t$.
\item $\kappa$ : Number of samples per class in the balanced test protocol (Definition~\ref{def}).
\item $\IFNew$ : Global imbalance factor (Eq.~\ref{IF_def}).
\item $\IFNew_{\mathcal{D}^1, \mathcal{D}^2}$ (equivalently $\text{IF}_{\mathcal{D}^1, \mathcal{D}^2}$) : Ratio $|\mathcal{D}^1|/|\mathcal{D}^2|$ between two partitions.
\item $\IFNew_{\mathcal{D}^i}$ : Local imbalance when partition $\mathcal{D}^i$ is viewed as its own problem, i.e.\ the ratio of the largest to the smallest per-class sample count among classes assigned to that partition (coincides with $R_i$ in Appendix~\ref{app:partitioning} when $\mathcal{D}^i$ aggregates classes in $\mathcal{C}_i$). Used in Assumption~\ref{assumption_3} to express that each final partition is not long-tailed.
\item $c^{\max} =  \arg \max_c~|\mathcal{D}_c|$ : Majority (largest) class by cardinality.
\item $c^{\min} =  \arg \min_c~|\mathcal{D}_c|$ : Minority (smallest) class by cardinality.

\item $k_H$, $k_T$ : Numbers of Head and Tail classes under Assumption~\ref{assumption_1} (so $k_H+k_T=k$ for total classes $k$).
\item $n_H$, $n_T$ : Samples per class within Head and within Tail under Assumption~\ref{assumption_1}. Together with Definition~\ref{def}'s requirement that $c^{\max}$ lies in the Head and $c^{\min}$ in the Tail, this implies $\IFNew=|\mathcal{D}_{c^{\max}}|/|\mathcal{D}_{c^{\min}}|=n_H/n_T$; $\gamma:=|\mathcal{D}_H|/|\mathcal{D}|$ is the Head sample mass fraction.

\item $\ell((\x_i, y_i), \theta)$ : Per-example loss (also written $\ell_{(\x_i,y_i)}(\theta)$).
\item $\mathcal{L}(\mathcal{D}_c, \theta) = \frac{1}{|\mathcal{D}_c|}\sum_{(\x_i, y_i) \in \mathcal{D}_c} \ell((\x_i, y_i), \theta)$ : Empirical average loss on $\mathcal{D}_c$.
\item $\mathcal{L}_{\mathcal{D}_c}(\theta) = \mathcal{L}(\mathcal{D}_c, \theta)$ : Shorthand for the same quantity.
\item $\mathcal{L}(\theta) := \mathcal{L}(\mathcal{D},\theta)$ : Full-dataset empirical risk on $\mathcal{D}$; whenever proofs add $\frac{\mu}{2}\|\theta\|^2$, they work with $\mathcal{L}_{\mathrm{total}}$ instead.
\item $\mathcal{L}_H(\theta) := \mathcal{L}(\mathcal{D}_H,\theta)$, $\mathcal{L}_T(\theta) := \mathcal{L}(\mathcal{D}_T,\theta)$ : Head and Tail average losses.
\item $\mathcal{L}_{\mathrm{bal}}(\theta) := \tfrac12 \mathcal{L}_T(\theta) + \tfrac12 \mathcal{L}_H(\theta)$ : Balanced LTR objective.
\item $\mathcal{L}_{\mathrm{task}}(\theta)$ : CL task loss; in CLTR stage-2, $\mathcal{L}_{\mathrm{task}}\equiv \mathcal{L}_T$.
\item $\mathcal{L}_{\mathrm{CL}}(\theta)$ : Unified continual-learning surrogate from Eq.~\ref{eq:wang-app}.
\item $\mathcal{L}_{\mathrm{total}}(\mathcal{D},\theta) := \mathcal{L}(\mathcal{D},\theta)+\frac{\mu}{2}\|\theta\|^2$ : $L^2$-regularized full objective minimized in the convex proof of Theorem~\ref{theorem1} (Eq.~\ref{loss_t}). \emph{Caution:} later lines of that proof reuse the symbol $\mathcal{L}_T(\mathcal{D})$ for this same regularized quantity; it is \emph{not} the Tail average loss $\mathcal{L}(\mathcal{D}_T,\theta)$.
\item $\mathcal{L}_c(\theta)$ : Class-conditional average loss on $\mathcal{D}_c$ in the partitioning analysis (Appendix~\ref{app:partitioning}).
\item $\mathcal{L}_i^{\mathrm{task}}$, $\mathcal{L}_i^{\mathrm{bal}}$, $\mathcal{L}_i^{\mathrm{CL}}$ : Per-partition task, balanced, and CL objectives in Appendix~\ref{app:partitioning}; $\theta_i^\star$ and $\theta_i^{\mathrm{bal},\star}$ denote their respective minimizers.
\item $\delta$ : $\max_{\theta}|\mathcal{L}(\mathcal{D},\theta)-\mathcal{L}(\mathcal{D}_H,\theta)|$ in the convex proof; scales as $\mathcal{O}(1/\IFNew)$.
\item $\lambda$, $\lambda_H$ : Smallest eigenvalues of $\nabla^2 \mathcal{L}_{\mathrm{total}}(\mathcal{D},\cdot)$ and $\nabla^2 \mathcal{L}_{\mathrm{total}}(\mathcal{D}_H,\cdot)$ in Lemma~\ref{lemma_1}; more generally $\lambda_f,\lambda_g$ are smallest Hessian eigenvalues for auxiliary functions $f,g$ in Lemma~\ref{lemma_1}.

\item $D_{\Phi}(\cdot,\cdot)$, $D_{\Psi}(\cdot,\cdot)$ : Bregman divergences generated by convex $C^1$ maps $\Phi$ (output space) and $\Psi$ (parameter space).
\item $D_{\mathcal{L}_H}(\theta,\theta_H^\star)$ : Bregman divergence generated by the Head loss $\mathcal{L}_H$ (when used).
\item $z$ : Fixed Head reference for output regularization (e.g., stored logits); $z_H^\star$ is the calibrated Head target in Assumption~\ref{ass:cltr}.
\item $D_{\Phi,i}$, $D_{\Psi,i}$, $\alpha_i$, $\beta_i$, $c_{\Phi,i}$, $z_{i-1}$ : Stage-$i$ output/parameter regularizers and weights in Appendix~\ref{app:partitioning}; $z_{i-1}$ anchors divergences at the end of stage $i-1$.
\item $\theta_{i-1}$ : Parameters at the end of stage $i-1$ (initial point for stage $i$ CL objectives). Distinct from $\theta_i$ in the main-text figure, which denotes an initialization before long-tailed training, and from $\theta_i^\star$, the exact minimizer of $\mathcal{L}_i^{\mathrm{CL}}$.
\item $\Delta_i := \big[\mathcal{L}_i^{\mathrm{task}}(\theta_{i-1})-\mathcal{L}_i^{\mathrm{task}}(\theta_i^\star)\big]_+$ : Nonnegative ``current-task learning pressure'' after stage $i$ (Appendix~\ref{app:partitioning}).
\item $\mathcal{C}_{<i}:=\bigcup_{j<i}\mathcal{C}_j$ : Union of class-index sets from partitions preceding $i$.
\item $\mathcal{L}_{<i}^{\mathrm{bal}}(\theta)$ : Uniform average $\frac{1}{|\mathcal{C}_{<i}|}\sum_{c\in\mathcal{C}_{<i}}\mathcal{L}_c(\theta)$ over old classes (Appendix~\ref{app:partitioning}).
\item $\mathcal{L}_{i,\mathrm{old}}^{\mathrm{sm}}$, $m_{\Psi,i}$ (and uniform shorthands $L_{\mathrm{old}}^{\mathrm{sm}}$, $m_\Psi$) : Local smoothness of old-task balanced losses and strong-convexity constants for $D_{\Psi,i}$ in the multi-stage bounds (Appendix~\ref{app:partitioning}).
\item $\mathcal{L}_{\mathrm{bal}}^\star$ : Shorthand for the minimum attainable globally balanced risk in the idealized analysis of Appendix~\ref{app:partitioning}.
\item $\mathrm{OptErr}$ : Slack collecting finite-horizon optimization and approximation errors in Appendix~\ref{app:partitioning}.

\item $\frac{\mu}{2}\|\theta\|^2$ : $L^2$ regularization with coefficient $\mu>0$. \emph{Important:} in Appendix~\ref{app:exp_decay}, $\mu(\cdot)$ also denotes a \emph{measure} on $[0,1]$ (e.g., $d\mu(u)=w_0\IFNew^{-u}du$); it is always clear from integration vs.\ penalty context.

\item $\theta$ : Learnable parameters; $\theta^i$ (superscript) denotes parameters after the $i^{\text{th}}$ training iterate in the global optimization loop. The subscripted $\theta_i$ appearing in Fig.~\ref{fig:losslandscape} denotes an initialization before long-tailed training (not a partition-stage parameter).
\item $\theta^*$ / $\theta^\star$ : Minimizer of the full-data objective being analyzed in each result (context-dependent).
\item $\theta^*_H$ / $\theta_H^\star$ : Head-only minimizer on $\mathcal{D}_H$.
\item $\theta^*_T$ : Minimizer when training on $\mathcal{D}_T$ alone.
\item $\theta^*_{HT}$ : Target solution with good Head and Tail behavior (intersection regime $\psi_{HT}$).
\item $\theta^*_{\mathcal{D}^1}$, $\theta^*_1$ : Minimizer when training only on the largest partition $\mathcal{D}^1$ (Theorem~\ref{theorem 2}).
\item $\theta^*_{\mathcal{B}}$ for $\mathcal{B}\subseteq\mathcal{D}$ : Minimizer of the empirical risk restricted to subset $\mathcal{B}$ (union-of-partitions notation in Appendix proofs).
\item $\theta_{H,k}^\star$ : $k^{\text{th}}$ local Head minimizer in the non-convex setting (Theorem~\ref{theorem_non}).
\item $\theta_k$ : Generic iterate along an optimization trajectory (e.g., after $k$ steps).

\item $h_{\theta}(\cdot)$ : Network map to logits/predictions used inside $D_{\Phi}$.

\item $\psi_H$, $\psi_T$, $\psi_{HT}$ : Informal ``good-performance'' regions in weight space for Head, Tail, and both.

\item $\mathcal{Y}_t$ : Label set at CL step $t$.
\item $\mathcal{M}$ : Replay memory buffer.
\item $F$ : Fisher information matrix (e.g., EWC).
\item $H_H$ : Positive-definite matrix in quadratic/EWC-style $\Psi(\theta)=\tfrac12(\theta-\theta_H^\star)^\top H_H(\theta-\theta_H^\star)$.

\item $S_H$ : Set of minimizers of $\mathcal{L}_H$ on $\mathcal{D}_H$ (a singleton under strict convexity; possibly non-unique in Theorem~\ref{theorem_non}).
\item $C$ : Positive constant in the local Hölder--$\tfrac12$/KL-type bound $\mathrm{dist}(\theta^*,S_H)\le C\sqrt{\mathcal{L}_H(\theta^*)-\mathcal{L}_H(\theta^*_H)}$ from Theorem~\ref{theorem_non}'s proof (any fixed $\theta^*_H\in S_H$ on the right-hand side).
\item $M$ : In Theorems~\ref{theorem1} and~\ref{theorem_non}, $M:=\max_{\theta}\big|\mathcal{L}(\mathcal{D}_T,\theta)-\mathcal{L}(\mathcal{D}_H,\theta)\big|$ (finite under the proof's bounded-loss assumption).

\item $\mathcal{N} \subset \mathbb{R}^d$ : Neighborhood of $\theta_H^\star$ where Assumption~\ref{ass:cltr} holds.
\item $L_H^{\mathrm{sm}}$, $m_{\Psi}$, $c_{\Phi}$ : Local smoothness of $\mathcal{L}_H$, strong-convexity parameter for $D_{\Psi}(\cdot,\theta_H^\star)$, and output calibration constant.
\item $\alpha$, $\beta$ : Weights in $\mathcal{L}_{\mathrm{CL}}$ (Eq.~\ref{eq:wang-app}); $\alpha_{\min}$, $\beta_{\min}$ are the thresholds in Theorem~\ref{theorem_general}.

\item $\mathcal{A}$ : Continual-learning algorithm in Proposition~\ref{prop:partitioning}.
\item $\eta_{\mathcal{A}}\ge 0$ : Effective forgetting coefficient for algorithm $\mathcal{A}$ in Proposition~\ref{prop:partitioning}; plain $\eta$ denotes the uniform-stage surrogate grouped into $S(N)$ in Appendix~\ref{app:partitioning}.
\item $N^\star_{\mathcal{A}}$ : Any member of $\arg\min_N$ of the bracket in Eq.~\ref{tradeoff_eq} for algorithm $\mathcal{A}$ (optimal partition count under the stated tradeoff).
\item $N^\star$ : Same for the simplified score $S(N)$ when algorithm dependence is suppressed (Appendix~\ref{app:partitioning}).
\item $B$ : Uniform upper bound on per-class losses $\mathcal{L}_c(\theta)$ over the optimization region in Appendix~\ref{app:partitioning}.
\item $\mathcal{C}_i$ : Set of class indices assigned to cardinality partition $i$.
\item $R_i$ : Local imbalance for partition $i$. Discrete datasets: $R_i=\max_{c\in\mathcal{C}_i}n_c/\min_{c\in\mathcal{C}_i}n_c$ with $n_c=|\mathcal{D}_c|$. Under the continuous exponential profile $w(u)=w_0\IFNew^{-u}$ with boundaries $0=u_0<\cdots<u_N=1$, also $R_i=w(u_{i-1})/w(u_i)=\IFNew^{u_i-u_{i-1}}$ (Proposition~\ref{prop:partitioning}).
\item $p_c$, $q_c$, $r_c$ : Empirical class weights $p_c=n_c/\sum_{d\in\mathcal{C}_i}n_d$, uniform $q_c=1/|\mathcal{C}_i|$, and ratio $r_c=p_c/q_c$ inside partition $i$ (Appendix~\ref{app:partitioning}); $\operatorname{TV}(p,q)=\tfrac12\sum_{c\in\mathcal{C}_i}|p_c-q_c|$ is total variation.
\item $S(N)$ : Scalar tradeoff score $2B\,\frac{\IFNew^{1/(2N)}-1}{\IFNew^{1/(2N)}+1}+(N-1)\eta$ minimized to choose $N$ when stage-wise constants are uniform (Appendix~\ref{app:partitioning}).
\item $\mathcal{U}=[0,1]$ : Continuous class-index domain.
\item $w(u)=w_0\IFNew^{-u}$ : Decaying class-frequency profile ($w_0>0$ a scale).
\item $d\mu(u)=w(u)\,du$ : Induced measure on $\mathcal{U}$ (here $\mu$ is a measure, not the $L^2$ weight).
\item $L_A(\theta)$ : Normalized risk on measurable $A\subseteq[0,1]$, including shared regularizer $R(\theta)$.
\item $\ell_u(\theta)$ : Integrand loss at continuous index $u$.
\item $P_\tau$, $T_\tau$ : Head prefix $[0,\tau]$ and Tail suffix $(\tau,1]$.
\item $\varepsilon_\tau=\mu(T_\tau)/\mu(\mathcal{U})$ : Tail mass fraction.
\item $S_{P_\tau}$, $L_{P_\tau}^\star$ : Minimizer set of $L_{P_\tau}$ and its optimal value.
\item $\bar\theta_{P_\tau}$ : Any member of $S_{P_\tau}$ used as a witness in the exponential-decay proof (Appendix~\ref{app:exp_decay}).
\item $\theta_\mathcal{U}^\star$ : Minimizer of the continuum full-domain risk $L_{\mathcal{U}}$ in Appendix~\ref{app:exp_decay}.
\item $\theta^\star_{\mathcal{D}}$ : Full-dataset minimizer in the discrete remark statement (Remark~\ref{rem:exp_decay}); conceptually parallel to $\theta^\star$ and $\theta_\mathcal{U}^\star$.
\item $C_\tau$ : KL/Hölder-type constant for prefix $P_\tau$.
\item $M_\tau$ : Uniform bound on $|L_{T_\tau}(\theta)-L_{P_\tau}(\theta)|$ over $\theta\in\mathcal{S}$ in Appendix~\ref{app:exp_decay}.
\item $\IFNew_\tau := w(0)/w(\tau)$ : Boundary imbalance factor (equals $\IFNew^\tau$ under this profile).
\item $\mathcal{S}$ : Local sublevel set on which loss contrasts are uniformly bounded.
\end{itemize}

\newpage
\section{Proofs}\label{app:proof}
Here we provide proofs for all the theorems. For convenience, we restate each theorem at the start of its corresponding section before presenting the proof.

\subsection{Proof of Theorem \ref{theorem1}\label{theorem1_proof}}

\restatetheorem{theorem1}{
Given Assumption \ref{assumption_1},
if a model is trained in an LTR setting (Definition \ref{def}), then the weights of the model after training ($\theta^*$) will lie within the bounded neighborhood of the model's weight if solely trained on Head ($\theta^*_H$), with the scale of:
\begin{equation}
{\|\theta^* - \theta^*_{H} \|} = \mathcal{O}(\frac{1}{\sqrt{\IFNew}}).
\end{equation}
}

\textit{Proof Sketch.} We start by expressing the total loss as a weighted combination of the Head and Tail set losses, where the weights depend on the dataset imbalance factor $\IFNew$. As $\IFNew$ becomes larger, the contribution of the Tail set loss diminishes, and the total loss approaches the Head set loss.
Next, we bound the difference between the total loss and the Head set loss in terms of $\IFNew$. Using a lemma that connects loss differences to the proximity of their respective minimizers under Assumption \ref{assumption_1}, we translate this bound into a relationship between the model parameters obtained from the total dataset and those obtained from the Head set.
Finally, we show that the loss difference scales as $\mathcal{O}(1/\IFNew)$, leading to the parameter difference scaling of $\mathcal{O}(1/\sqrt{\IFNew})$.

\begin{proof}[Proof of Theorem \ref{theorem1}]
Following Assumption \ref{assumption_1}, the model is trained on the entire dataset $\mathcal{D}$ by minimizing the loss function $\mathcal{L}_{T}$, which can be formulated as:
\begin{equation} \label{loss_t}
\mathcal{L}_{total}(\mathcal{D})=\mathcal{L}(\mathcal{D}) + \frac{\mu}{2}\|\theta\|^2,
\end{equation}

The loss $\mathcal{L}(\mathcal{D})$ can be decomposed as:
\begin{equation} \label{loss}
\mathcal{L}(\mathcal{D})=\frac{1}{|\mathcal{D}|}\left(\sum_{(\x_i,y_i)\in\mathcal{D}_H}\ell(( \x_i,y_i))+\sum_{( \x_i,y_i)\in\mathcal{D}_T}\ell(( \x_i,y_i))\right),
\end{equation}
Using $\mathcal{L}(\mathcal{D}_H) = \frac{1}{|\mathcal{D}_H|}\sum_{( \x_i,y_i)\in\mathcal{D}_H}\ell(( \x_i,y_i))$
and
$\mathcal{L}(\mathcal{D}_T) = \frac{1}{|\mathcal{D}_T|}\sum_{( \x_i,y_i)\in\mathcal{D}_T}\ell(( \x_i,y_i))$, we can rewrite $\mathcal{L}(\mathcal{D})$ as:
\begin{equation} \label{loss_3}
\mathcal{L}(\mathcal{D})= \frac{|\mathcal{D}_H|}{|\mathcal{D}|}\mathcal{L}(\mathcal{D}_H)+\frac{|\mathcal{D}_T|}{|\mathcal{D}|}\mathcal{L}(\mathcal{D}_T).
\end{equation}
Let $\gamma = \frac{|\mathcal{D}_H|}{|\mathcal{D}|}$. Since $|\mathcal{D}| = |\mathcal{D}_H|+|\mathcal{D}_T|$, we can derive that $1-\gamma = \frac{|\mathcal{D}_T|}{|\mathcal{D}|}$. Plugging $\gamma$ into Eq. \ref{loss_3} yields:
\begin{equation} \label{loss_4}
\mathcal{L}(\mathcal{D})= \gamma\mathcal{L}(\mathcal{D}_H)+(1-\gamma)\mathcal{L}(\mathcal{D}_T).
\end{equation}
We now express $\gamma$ in terms of the imbalance factor $\IFNew$ in Definition \ref{def} under the LTR setting.
Let $k_H$ and $k_T$ denote the number of Head and Tail classes, respectively ($k_H+k_T=k$).
Under Assumption \ref{assumption_1} (LTR setting with approximately constant per-class sizes within Head and within Tail),
let $n_H$ be the number of samples per Head class and $n_T$ be the number of samples per Tail class.
Then $|\mathcal{D}_H| = k_H n_H$ and $|\mathcal{D}_T| = k_T n_T$.
Moreover, by Definition \ref{def}, $\IFNew = \frac{n_H}{n_T}$, hence
\begin{equation}\label{gamma_if}
\gamma=\frac{|\mathcal{D}_H|}{|\mathcal{D}_H|+|\mathcal{D}_T|}
=\frac{k_H n_H}{k_H n_H + k_T n_T}
=\frac{k_H \IFNew}{k_H \IFNew + k_T},
\end{equation}
and therefore
\begin{equation}\label{one_minus_gamma_if}
1-\gamma=\frac{k_T}{k_H \IFNew + k_T}.
\end{equation}
Since in LTR we have $\IFNew \gg 1$ (Definition \ref{def}), $\gamma \to 1$ and $1-\gamma = \mathcal{O}(1/\IFNew)$.
Consequently, $\mathcal{L}(\mathcal{D})$ approaches $\mathcal{L}(\mathcal{D}_H)$ for all $\theta$ values.

Let $\delta$ represent the maximum difference between the losses:
\begin{equation} \label{loss_5}
 \delta = \max_{\theta}| \mathcal{L}(\mathcal{D}) - \mathcal{L}(\mathcal{D}_{H}) | .
\end{equation}
From Eq. \ref{loss_4}, it follows that as $\IFNew \to \infty$, $\delta \to 0$. To find the asymptotic relationship of $\delta$ with $\IFNew$, we derive:
\begin{equation} \label{loss_5_0}
| \mathcal{L}(\mathcal{D}) - \mathcal{L}(\mathcal{D}_{H}) | = |(1-\gamma)(\mathcal{L}(\mathcal{D}_T) - \mathcal{L}(\mathcal{D}_{H}))|,
\end{equation}
Replacing $1-\gamma$ with $\frac{k_T}{k_H \IFNew + k_T}$ from Eq. \ref{one_minus_gamma_if} and simplifying:
\begin{equation} \label{loss_5_0_1}
| \mathcal{L}(\mathcal{D}) - \mathcal{L}(\mathcal{D}_{H}) | = \frac{k_T}{k_H \IFNew + k_T}| \mathcal{L}(\mathcal{D}_T) - \mathcal{L}(\mathcal{D}_{H}) |.
\end{equation}
By applying maximization over $\theta$, we derive:
\begin{equation} \label{loss_5_0_1_1}
\max_{\theta}| \mathcal{L}(\mathcal{D}) - \mathcal{L}(\mathcal{D}_{H}) |
= \frac{k_T}{k_H \IFNew + k_T}\max_{\theta}\left| \mathcal{L}(\mathcal{D}_T) - \mathcal{L}(\mathcal{D}_{H}) \right|.
\end{equation}
Since $\IFNew$ is independent of $\theta$:
\begin{equation} \label{loss_5_0_1_2}
\max_{\theta}| \mathcal{L}(\mathcal{D}) - \mathcal{L}(\mathcal{D}_{H}) | = M\frac{k_T}{k_H \IFNew + k_T},
\end{equation}
where $M = \max_{\theta} (| \mathcal{L}(\mathcal{D}_T) - \mathcal{L}(\mathcal{D}_{H}) |) $. We assume $M<\infty$ (e.g., $\ell$ is bounded/clipped or the optimization domain is restricted). Since $\mathcal{L}(\mathcal{D}_T)$ and $\mathcal{L}(\mathcal{D}_H)$ are average loss over Head and Tail, respectively, they are independent of $\IFNew$. Moreover, number of classes in the Head ($k_H$) and number of classes in the Tail ($k_T$) are also independent of $\IFNew$. Consequently, replacing $\delta$ from Eq. \ref{loss_5} in Eq. \ref{loss_5_0_1_2} yields:
\begin{equation} \label{loss_5_0_1_3}
\delta = \mathcal{O}\left(\frac{1}{\IFNew}\right).
\end{equation}

Adding and subtracting the regularizer $\frac{\mu}{2}\|\theta\|^2$ in Eq. \ref{loss_5} yields:
\begin{equation} \label{loss_5_1}
 \delta = \max_{\theta}| \mathcal{L}_T(\mathcal{D}) - \mathcal{L}_T(\mathcal{D}_{H}) | .
\end{equation}

We now use the following lemma to relate the difference in losses to the difference in the parameter vectors:

\begin{lemma}\label{lemma_1}
Let $f,g:\mathbb{R}^d\to\mathbb{R}$ satisfy $|f(x)-g(x)|\le\delta$ for all $x$, and let
$x_f\in\arg\min f$, $x_g\in\arg\min g$. Assume quadratic growth around the minimizers with
constants $\lambda_f,\lambda_g\ge 0$, $\lambda_f+\lambda_g>0$:
$
f(y)\ge f(x_f)+\tfrac{\lambda_f}{2}\|y-x_f\|^2,\qquad
g(y)\ge g(x_g)+\tfrac{\lambda_g}{2}\|y-x_g\|^2\qquad\forall y .
$
(This holds in particular if $f$ is $\lambda_f$-strongly convex and $g$ is $\lambda_g$-strongly convex.)
Then $\|x_f-x_g\|^2\le \frac{4\delta}{\lambda_f+\lambda_g}$.
\end{lemma}

Applying Lemma \ref{lemma_1} to Eq. \ref{loss_5_1}, we get:
\begin{equation} \label{eq11}
\|\theta^* - \theta^*_H\|^2 \leq \frac{4\delta}{\lambda + \lambda_H},
\end{equation}
where $\lambda$ and $\lambda_H$ are the minimum eigenvalues of the Hessians of $\mathcal{L}_T(\mathcal{D})$ and $\mathcal{L}_T(\mathcal{D}_H)$, respectively.

To find $\lambda$ and $\lambda_H$, note that:
\begin{equation}
\nabla^2 \mathcal{L}_T(\mathcal{D}) = \nabla^2 \mathcal{L}(\mathcal{D}) + \mu \mI, \quad \nabla^2 \mathcal{L}_T(\mathcal{D}_H) = \nabla^2 \mathcal{L}(\mathcal{D}_H) + \mu \mI,
\end{equation}
Since $\nabla^2 \mathcal{L}(\mathcal{D}) \succeq 0$ and $\nabla^2 \mathcal{L}(\mathcal{D}_H) \succeq 0$ due to the fact that $\mathcal{L}(.)$ is convex, we have $\lambda, \lambda_H \geq \mu$.

Substituting this back into Eq. \ref{eq11} yields:
\begin{equation}\label{way_to_calc}
\|\theta^* - \theta^*_H\|^2 \leq \frac{2\delta}{\mu}.
\end{equation}
From Eq. \ref{loss_5_0_1_3}, $\delta = \mathcal{O}(\frac{1}{\IFNew})$, so:
\begin{equation}
\|\theta^* - \theta^*_H\|^2 = \mathcal{O}\left(\frac{1}{\IFNew}\right).
\end{equation}
According to Definition \ref{def}, $\IFNew \gg 1$, consequently:
\begin{equation}\label{eq13}
\|\theta^* - \theta^*_H\| = \mathcal{O}(\frac{1}{\sqrt{\IFNew}}).
\end{equation}
Eq. \ref{eq13} and Eq. \ref{way_to_calc} complete the proof.
\end{proof}

\subsection{Proof of Lemma \ref{lemma_1}}\label{lemma_1_proof}

\textit{Proof Sketch.} We evaluate the quadratic growth of $f$ at the minimizer of $g$ and the quadratic growth of $g$ at the minimizer of $f$. Adding the two inequalities bounds $\|x_f-x_g\|^2$ by two differences between $f$ and $g$, each of which is at most $\delta$. Finally, we show that strong convexity implies the required quadratic growth.

\begin{proof}
By assumption, $f$ and $g$ grow at least quadratically around their minimizers $x_f$ and $x_g$, respectively:
\begin{equation}\label{lem1_growth_f}
f(y) \geq f(x_f)+\frac{\lambda_f}{2}\|y-x_f\|^2, \qquad \forall y\in\mathbb{R}^d,
\end{equation}
\begin{equation}\label{lem1_growth_g}
g(y) \geq g(x_g)+\frac{\lambda_g}{2}\|y-x_g\|^2, \qquad \forall y\in\mathbb{R}^d.
\end{equation}
Evaluating Eq. \ref{lem1_growth_f} at $y=x_g$ and Eq. \ref{lem1_growth_g} at $y=x_f$, we obtain:
\begin{equation}\label{lem1_f_at_xg}
f(x_g)-f(x_f) \geq \frac{\lambda_f}{2}\|x_g-x_f\|^2,
\end{equation}
\begin{equation}\label{lem1_g_at_xf}
g(x_f)-g(x_g) \geq \frac{\lambda_g}{2}\|x_f-x_g\|^2.
\end{equation}
Since $\|x_g-x_f\|=\|x_f-x_g\|$, adding Eq. \ref{lem1_f_at_xg} and Eq. \ref{lem1_g_at_xf} together and regrouping the terms by their evaluation points yields:
\begin{equation}\label{lem1_sum}
\frac{\lambda_f+\lambda_g}{2}\|x_f-x_g\|^2 \leq \big(f(x_g)-g(x_g)\big)+\big(g(x_f)-f(x_f)\big).
\end{equation}
Using $|f(x)-g(x)|\leq\delta$ for all $x$, each of the two terms on the right-hand side of Eq. \ref{lem1_sum} is bounded by $\delta$:
\begin{equation}\label{lem1_delta}
f(x_g)-g(x_g)\leq\delta, \qquad g(x_f)-f(x_f)\leq\delta.
\end{equation}
Substituting Eq. \ref{lem1_delta} into Eq. \ref{lem1_sum}, we get:
\begin{equation}\label{lem1_main}
\frac{\lambda_f+\lambda_g}{2}\|x_f-x_g\|^2 \leq 2\delta.
\end{equation}
Since $\lambda_f+\lambda_g>0$, Eq. \ref{lem1_main} can be rewritten as:
\begin{equation}
\|x_f-x_g\|^2 \leq \frac{4\delta}{\lambda_f+\lambda_g}.
\end{equation}

It remains to show that the quadratic growth in Eq. \ref{lem1_growth_f} holds whenever $f$ is $\lambda$-strongly convex (the argument for $g$ is identical). By the definition of $\lambda$-strong convexity, for all $x,y\in\mathbb{R}^d$ and $t\in[0,1]$:
\begin{equation}\label{lem1_strong_def}
f\big((1-t)x+ty\big) \leq (1-t)f(x)+tf(y)-t(1-t)\frac{\lambda}{2}\|y-x\|^2.
\end{equation}
Setting $x=x_f$ in Eq. \ref{lem1_strong_def} and using the optimality of $x_f$, i.e. $f(x_f)\leq f\big((1-t)x_f+ty\big)$, we have for all $t\in(0,1)$:
\begin{equation}\label{lem1_strong_min}
f(x_f) \leq (1-t)f(x_f)+tf(y)-t(1-t)\frac{\lambda}{2}\|y-x_f\|^2.
\end{equation}
Subtracting $(1-t)f(x_f)$ from both sides of Eq. \ref{lem1_strong_min} and dividing by $t>0$ yields:
\begin{equation}\label{lem1_strong_div}
f(x_f)+(1-t)\frac{\lambda}{2}\|y-x_f\|^2 \leq f(y).
\end{equation}
Letting $t\to0^{+}$ in Eq. \ref{lem1_strong_div}, we obtain:
\begin{equation}
f(x_f)+\frac{\lambda}{2}\|y-x_f\|^2 \leq f(y),
\end{equation}
which is Eq. \ref{lem1_growth_f} with $\lambda_f=\lambda$. This completes the proof.
\end{proof}

\subsection{Proof of Theorem \ref{theorem_non}}\label{theorem_non_proof}
\restatetheorem{theorem_non}{
Given Assumption \ref{assumption_non}, there exists a bound for the distance between the weights of the model after training ($\theta^*$) and one of the minima of the loss landscape when training on the head solely, scaling inversely with the square root of the imbalance factor:
\begin{equation}
\textnormal{{\text{dist}}}(\theta^*,S_H)= \mathcal{O}(\frac{1}{\sqrt{\IFNew}}).
\end{equation}}

\textit{Proof Sketch.}
The proof first establishes a Hölder–1/2 error bound based on the loss's KL property and piecewise strong convexity (Assumption \ref{assumption_non}), which links the parameter distance $\mathrm{dist}(\theta^*,S_H)$ to the square root of the Head-set loss difference. To bound this loss difference in terms of $\IFNew$, we decompose it by adding and subtracting the total loss $\mathcal{L}$ at the optimal points $\theta^*$ and $\theta^*_H$. Using the optimality of $\theta^*$ for $\mathcal{L}$, we simplify the inequality by removing a non-positive term. Finally, we apply the bound from Eq. \ref{loss_5_0_1_2}, which limits the deviation between $\mathcal{L}$ and $\mathcal{L}_H$ to $\mathcal{O}(1/\IFNew)$, to the remaining terms. This shows the loss difference $\mathcal{L}_H(\theta^*) - \mathcal{L}_H(\theta^*_H)$ scales as $\mathcal{O}(1/\IFNew)$, and substituting this back into the Hölder bound (Eq. \ref{holder}) gives the final parameter distance scaling of $\mathcal{O}(1/\sqrt{\IFNew})$.

\begin{proof}
Following assumption \ref{assumption_non}, $\mathcal{L}$ has the KL property with an exponent of 1/2. Since the loss is regularized with $L^2$ regularization, the total loss function is proved to be piecewise strongly convex \cite{milne2019piecewise}. Hence, a Hölder–1/2 error bound holds for the loss function $\mathcal{L}_H$:
\begin{equation} \label{holder}
 \mathrm{dist}(\theta^*,S_H) \le C\sqrt{\mathcal{L}_H(\theta^*)-\mathcal{L}_H(\theta^*_H)}\qquad\text{for all }\theta^*_H\in S_H,
\end{equation}
where $C$ is a constant. Let $\theta^*_H$ be any element in $S_H$. Since $\theta^*$ is the minimizer of $\mathcal{L}$, by optimality we have:
\begin{equation}
\mathcal{L}(\theta^*) \le \mathcal{L}(\theta_H^*)
\end{equation}
We seek to bound the term $\mathcal{L}_H(\theta^*) - \mathcal{L}_H(\theta^*_H)$. We can add and subtract terms to decompose it:
\begin{equation}
\mathcal{L}_H(\theta^*) - \mathcal{L}_H(\theta^*_H) = (\mathcal{L}_H(\theta^*) - \mathcal{L}(\theta^*)) + (\mathcal{L}(\theta^*) - \mathcal{L}(\theta_H^*)) + (\mathcal{L}(\theta_H^*) - \mathcal{L}_H(\theta^*_H))
\end{equation}
From the optimality of $\theta^*$, we know the middle term is non-positive: $(\mathcal{L}(\theta^*) - \mathcal{L}(\theta_H^*)) \le 0$. We can thus bound the expression by dropping this term:
\begin{equation}
\mathcal{L}_H(\theta^*) - \mathcal{L}_H(\theta^*_H) \le (\mathcal{L}_H(\theta^*) - \mathcal{L}(\theta^*)) + (\mathcal{L}(\theta_H^*) - \mathcal{L}_H(\theta^*_H))
\end{equation}
We now apply the bound from Eq. \ref{loss_5_0_1_2} ($\max_{\theta}| \mathcal{L}(\mathcal{D}) - \mathcal{L}(\mathcal{D}_{H}) | = M\frac{k_T}{k_H \IFNew + k_T}$) to both terms on the right-hand side:
\begin{align}
(\mathcal{L}_H(\theta^*) - \mathcal{L}(\theta^*)) &\le |\mathcal{L}_H(\theta^*) - \mathcal{L}(\theta^*)| \le M\frac{k_T}{k_H \IFNew + k_T} \\
(\mathcal{L}(\theta_H^*) - \mathcal{L}_H(\theta^*_H)) &\le |\mathcal{L}(\theta_H^*) - \mathcal{L}_H(\theta^*_H)| \le M\frac{k_T}{k_H \IFNew + k_T}
\end{align}
Substituting these bounds back gives:
\begin{equation} \label{l_diff}
 \mathcal{L}_H(\theta^*) - \mathcal{L}_H(\theta_H^*) \le M\frac{k_T}{k_H \IFNew + k_T} + M\frac{k_T}{k_H \IFNew + k_T} = 2M\frac{k_T}{k_H \IFNew + k_T}
\end{equation}
Replacing the result from Eq. \ref{l_diff} into the error bound Eq. \ref{holder} yields:
\begin{equation}
 \mathrm{dist}(\theta^*,S_H) \le C\sqrt{\mathcal{L}_H(\theta^*)-\mathcal{L}_H(\theta^*_H)} \le C \sqrt{2M\frac{k_T}{k_H \IFNew + k_T}}
\end{equation}
Consequently:
\begin{equation}
\mathrm{dist}(\theta^*,S_H) = \mathcal{O}\left(\frac{1}{\sqrt{\IFNew}}\right),
\end{equation}
which completes the proof.
\end{proof}

\subsection{Proof of Theorem \ref{theorem 2}}\label{theorem_2_proof}

\restatetheorem{theorem 2}{Following Assumption~\ref{assumption_3}, let $\mathcal{D}=\bigcup_{i=1}^{n}\mathcal{D}^i$ be divided into $n\geq 2$ disjoint partitions, ordered such that $|\mathcal{D}^1|\geq\cdots\geq|\mathcal{D}^n|$.
Taking $\mathcal{D}^1$ as the Head subset $\mathcal{D}_H$, define the aggregated Tail as $\mathcal{D}_T=\bigcup_{i=2}^{n}\mathcal{D}^i$}
and the Head minimizer set as
$S_{\mathcal{D}^1} = \arg\min_{\theta} F_{\mathcal{D}^1}(\theta),$
where $F_A(\theta)=\mathcal{L}(A,\theta)+\mathcal{R}(\theta)$ denotes the training objective with a shared regularizer. Let $\theta^*$ minimize $F_{\mathcal{D}}$. Under the local error-bound and bounded loss-contrast conditions of Theorem~\ref{theorem_non}, applied to the Head and aggregated Tail, and assuming $\theta^*$ lies within the corresponding neighborhood, the following bound holds for fixed $n$:
\begin{align} \operatorname{dist} \left(\theta^*,S_{\mathcal{D}^1}\right)
&\leq C\sqrt{
2M\frac{|\mathcal{D}_T|}{|\mathcal{D}|}
}\leq C\sqrt{
2M(n-1)
\frac{|\mathcal{D}^2|}{|\mathcal{D}^1|}
}=
\mathcal{O}_n\left(
\frac{1}{
\sqrt{\IFNew_{\mathcal{D}^1,\mathcal{D}^2}}
}
\right),
\end{align}
where $\IFNew_{\mathcal{D}^1,\mathcal{D}^2} =|\mathcal{D}^1|/|\mathcal{D}^2|$, and $C,M>0$ are constants defined by the local error bound and loss-contrast condition, respectively.

When the Head objective has a unique minimizer $\theta^*_{\mathcal{D}^1}$, the set-valued distance reduces to $\|\theta^*-\theta^*_{\mathcal{D}^1}\|$.

\textit{Proof Sketch.} We extend the perturbation argument of Theorem~\ref{theorem_non} to multiple partitions by treating the largest partition $\mathcal{D}^1$ as the Head and aggregating all remaining partitions into a single Tail set $\mathcal{D}_T=\bigcup_{i=2}^{n}\mathcal{D}^i$. Using the optimality of the full-dataset solution and a bounded Head--Tail loss contrast, we show that the increase in Head loss is proportional to the relative size of the aggregated Tail. The local H\"older-$1/2$ error bound then converts this loss difference into a bound on the distance to the Head minimizer set. Finally, the cardinality ordering of the partitions yields the inverse-square-root scaling with respect to the imbalance factor between the two largest partitions.

\begin{proof}[Proof of Theorem~\ref{theorem 2}]
Let the long-tailed dataset be divided into $n$ disjoint partitions ordered by cardinality:
\begin{equation}
\mathcal{D}=\bigcup_{i=1}^{n}\mathcal{D}^i, \qquad |\mathcal{D}^1|\geq|\mathcal{D}^2|\geq\cdots\geq|\mathcal{D}^n|.
\end{equation}
We treat $\mathcal{D}^1$ as the Head partition and aggregate the remaining partitions into the Tail, $\mathcal{D}_T=\bigcup_{i=2}^{n}\mathcal{D}^i$. The relative Tail mass is defined as:
\begin{equation}\label{eq:tail_mass}
\varepsilon=\frac{|\mathcal{D}_T|}{|\mathcal{D}|}=\frac{\sum_{i=2}^{n}|\mathcal{D}^i|}{\sum_{i=1}^{n}|\mathcal{D}^i|}.
\end{equation}
For any nonempty subset $A\subseteq\mathcal{D}$, let $F_A(\theta)=\mathcal{L}(A,\theta)+\mathcal{R}(\theta)$ denote the training objective, where $\mathcal{L}(A,\theta)$ is the average empirical loss and $\mathcal{R}(\theta)$ is the shared regularization term. Since the partitions are disjoint, the full-dataset objective can be decomposed as:
\begin{equation}\label{eq:full_decomposition}
F_{\mathcal{D}}(\theta)=(1-\varepsilon)F_{\mathcal{D}^1}(\theta)+\varepsilon F_{\mathcal{D}_T}(\theta).
\end{equation}
Equivalently,
\begin{equation}\label{eq:perturbed_objective}
F_{\mathcal{D}}(\theta)=F_{\mathcal{D}^1}(\theta)+\varepsilon\Delta(\theta),
\end{equation}
where $\Delta(\theta)=F_{\mathcal{D}_T}(\theta)-F_{\mathcal{D}^1}(\theta)$ represents the Head--Tail loss contrast.

Define the Head minimizer set as $S_{\mathcal{D}^1}=\arg\min_{\theta}F_{\mathcal{D}^1}(\theta)$ and its minimum loss as $F_{\mathcal{D}^1}^{*}=\min_{\theta}F_{\mathcal{D}^1}(\theta)$. Following the local geometric conditions employed in Theorem~\ref{theorem_non}, assume that there exists a neighborhood $\mathcal{U}$ of $S_{\mathcal{D}^1}$ containing the full-dataset minimizer $\theta^*$, on which the H\"older-$1/2$ error bound holds:
\begin{equation}\label{eq:head_error_bound}
\operatorname{dist}(\theta,S_{\mathcal{D}^1})\leq C\sqrt{F_{\mathcal{D}^1}(\theta)-F_{\mathcal{D}^1}^{*}}, \qquad \forall\theta\in\mathcal{U},
\end{equation}
for some constant $C>0$. We further assume that the Head--Tail loss contrast is uniformly bounded on this neighborhood:
\begin{equation}\label{eq:loss_contrast}
\sup_{\theta\in\mathcal{U}}|\Delta(\theta)|\leq M,
\end{equation}
where $M<\infty$.

Let $\bar{\theta}\in S_{\mathcal{D}^1}$ be a Head minimizer. Since $\theta^*$ minimizes $F_{\mathcal{D}}$, its optimality implies:
\begin{equation}\label{eq:full_optimality}
F_{\mathcal{D}}(\theta^*)\leq F_{\mathcal{D}}(\bar{\theta}).
\end{equation}
Substituting Eq.~\ref{eq:perturbed_objective} into Eq.~\ref{eq:full_optimality}, we obtain:
\begin{equation}
F_{\mathcal{D}^1}(\theta^*)+\varepsilon\Delta(\theta^*)\leq F_{\mathcal{D}^1}(\bar{\theta})+\varepsilon\Delta(\bar{\theta}).
\end{equation}
Rearranging and using $F_{\mathcal{D}^1}(\bar{\theta})=F_{\mathcal{D}^1}^{*}$, we have:
\begin{align}\label{eq:head_loss_bound}
F_{\mathcal{D}^1}(\theta^*)-F_{\mathcal{D}^1}^{*}
&\leq \varepsilon\left[\Delta(\bar{\theta})-\Delta(\theta^*)\right]\nonumber\\
&\leq 2M\varepsilon,
\end{align}
where the last inequality follows from Eq.~\ref{eq:loss_contrast}. Applying the local error bound in Eq.~\ref{eq:head_error_bound} at $\theta^*$ gives:
\begin{align}\label{eq:aggregate_bound}
\operatorname{dist}(\theta^*,S_{\mathcal{D}^1})
&\leq C\sqrt{F_{\mathcal{D}^1}(\theta^*)-F_{\mathcal{D}^1}^{*}}\nonumber\\
&\leq C\sqrt{2M\varepsilon}.
\end{align}

We now bound the relative Tail mass using the cardinality ordering of the partitions. Since $|\mathcal{D}^i|\leq|\mathcal{D}^2|$ for all $i\geq 2$, we have:
\begin{equation}
|\mathcal{D}_T|=\sum_{i=2}^{n}|\mathcal{D}^i|\leq(n-1)|\mathcal{D}^2|.
\end{equation}
Consequently,
\begin{align}\label{eq:epsilon_bound}
\varepsilon=\frac{|\mathcal{D}_T|}{|\mathcal{D}|}
\leq\frac{(n-1)|\mathcal{D}^2|}{|\mathcal{D}^1|}.
\end{align}
Substituting Eq.~\ref{eq:epsilon_bound} into Eq.~\ref{eq:aggregate_bound}, we obtain:
\begin{equation}
\operatorname{dist}(\theta^*,S_{\mathcal{D}^1})\leq C\sqrt{2M(n-1)\frac{|\mathcal{D}^2|}{|\mathcal{D}^1|}}.
\end{equation}
Defining the imbalance factor between the two largest partitions as $\IFNew_{\mathcal{D}^1,\mathcal{D}^2}=|\mathcal{D}^1|/|\mathcal{D}^2|$, the final bound becomes:
\begin{align}
\operatorname{dist}(\theta^*,S_{\mathcal{D}^1})
&\leq C\sqrt{\frac{2M(n-1)}{\IFNew_{\mathcal{D}^1,\mathcal{D}^2}}}\nonumber\\
&=\mathcal{O}_n\left(\frac{1}{\sqrt{\IFNew_{\mathcal{D}^1,\mathcal{D}^2}}}\right),
\end{align}
for fixed $n$ and constants $C,M$ independent of the imbalance factor.

Finally, when the Head objective admits a unique minimizer $\theta^*_{\mathcal{D}^1}$, the minimizer set reduces to a singleton, and therefore:
\begin{equation}
\operatorname{dist}(\theta^*,S_{\mathcal{D}^1})=\|\theta^*-\theta^*_{\mathcal{D}^1}\|.
\end{equation}
Thus, the result extends the Head-dominance analysis to multiple partitions without requiring uniqueness of the Head minimizer, while preserving the inverse-square-root scaling with respect to the partition imbalance factor.
\end{proof}

\subsection{Exponential-Decay Extension of Theorem~\ref{theorem 2}}
\label{app:exp_decay}
We now give the full statement and proof of Remark~\ref{rem:exp_decay}.

\restateremark{rem:exp_decay}{
The partition-based statement in Theorem~\ref{theorem 2} also extends to the standard exponential long-tail model used in CIFAR10-LT and CIFAR100-LT. Let sorted class frequencies follow
$w(u)=w_0\IFNew^{-u}$ for $u\in[0,1]$, where $\IFNew=w(0)/w(1)$. For a Head prefix $P_\tau=[0,\tau]$ and Tail suffix $T_\tau=(\tau,1]$, the Tail mass fraction is $
\varepsilon_\tau
=
\frac{\mu(T_\tau)}{\mu([0,1])}
=
\frac{\IFNew^{-\tau}-\IFNew^{-1}}{1-\IFNew^{-1}}.
$ Under the same local error-bound/KL-$1/2$ condition used in Theorem~\ref{theorem_non}, the full-data optimum satisfies

\begin{equation}
\operatorname{dist}(\theta^\star_{\mathcal D},S_{P_\tau})
\le
C_\tau
\sqrt{
2M_\tau
\frac{\IFNew^{-\tau}-\IFNew^{-1}}
{1-\IFNew^{-1}}
}
=
\mathcal{O}(\IFNew^{-\tau/2}).
\end{equation}

Equivalently, defining the boundary imbalance factor
$\IFNew_\tau=w(0)/w(\tau)=\IFNew^\tau$, we recover the same inverse-square-root form

\begin{equation}
\operatorname{dist}(\theta^\star_{\mathcal D},S_{P_\tau})
=
\mathcal{O}(\IFNew_\tau^{-1/2}).
\end{equation}

Thus, the geometric conclusion of Theorem~\ref{theorem 2} is not an artifact of uniform within-partition cardinalities; it also holds for exponentially decaying long-tailed datasets.
}
\begin{proof}
Let $\mathcal{U}=[0,1]$ denote the full sorted class-index domain. This continuous domain is the measure-theoretic analogue of the full dataset $\mathcal{D}$; hence $L_{\mathcal{U}}$ and $\theta^\star_{\mathcal{U}}$ correspond to the full-data loss and full-data optimum in the exponential-decay model. Recall that the Head prefix and Tail suffix are defined as
$P_\tau=[0,\tau]$ and $T_\tau=(\tau,1]$, so that $\mathcal{U}=P_\tau\cup T_\tau$ and $P_\tau\cap T_\tau=\emptyset$.

We begin by writing the training measure induced by the exponential class-frequency profile. Since
$w(u)=w_0\IFNew^{-u}$, define
\begin{equation}
d\mu(u)=w(u)du=w_0\IFNew^{-u}du.
\end{equation}
For any measurable set $A\subseteq[0,1]$, define the normalized risk over $A$ as
\begin{equation}
L_A(\theta)
=
\frac{1}{\mu(A)}
\int_A \ell_u(\theta)d\mu(u)
+
R(\theta),
\end{equation}
where $R(\theta)$ is the same regularization term for all subsets.

The first step is to decompose the full-data risk into a weighted combination of the Head-prefix risk and the Tail-suffix risk. By definition,
\begin{equation}
L_\mathcal{U}(\theta)
=
\frac{1}{\mu(\mathcal{U})}
\int_\mathcal{U} \ell_u(\theta)d\mu(u)
+
R(\theta).
\end{equation}
Since $\mathcal{U}=P_\tau\cup T_\tau$ and the two sets are disjoint, we have
\begin{equation}
\int_\mathcal{U} \ell_u(\theta)d\mu(u)
=
\int_{P_\tau}\ell_u(\theta)d\mu(u)
+
\int_{T_\tau}\ell_u(\theta)d\mu(u).
\end{equation}
Therefore,
\begin{equation}
L_\mathcal{U}(\theta)
=
\frac{1}{\mu(\mathcal{U})}
\left(
\int_{P_\tau}\ell_u(\theta)d\mu(u)
+
\int_{T_\tau}\ell_u(\theta)d\mu(u)
\right)
+
R(\theta).
\end{equation}
Using the definitions of $L_{P_\tau}$ and $L_{T_\tau}$, we can rewrite the two integrals as
\begin{equation}
\int_{P_\tau}\ell_u(\theta)d\mu(u)
=
\mu(P_\tau)\left(L_{P_\tau}(\theta)-R(\theta)\right),
\end{equation}
and
\begin{equation}
\int_{T_\tau}\ell_u(\theta)d\mu(u)
=
\mu(T_\tau)\left(L_{T_\tau}(\theta)-R(\theta)\right).
\end{equation}
Substituting these identities into the expression for $L_\mathcal{U}(\theta)$ gives
\begin{equation}
\begin{aligned}
L_\mathcal{U}(\theta)
&=
\frac{\mu(P_\tau)}{\mu(\mathcal{U})}
\left(L_{P_\tau}(\theta)-R(\theta)\right)
+
\frac{\mu(T_\tau)}{\mu(\mathcal{U})}
\left(L_{T_\tau}(\theta)-R(\theta)\right)
+
R(\theta).
\end{aligned}
\end{equation}
Since $\mu(\mathcal{U})=\mu(P_\tau)+\mu(T_\tau)$, the regularization terms combine as
\begin{equation}
-
\frac{\mu(P_\tau)}{\mu(\mathcal{U})}R(\theta)
-
\frac{\mu(T_\tau)}{\mu(\mathcal{U})}R(\theta)
+
R(\theta)
=
0.
\end{equation}
Thus,
\begin{equation}
L_\mathcal{U}(\theta)
=
\frac{\mu(P_\tau)}{\mu(\mathcal{U})}L_{P_\tau}(\theta)
+
\frac{\mu(T_\tau)}{\mu(\mathcal{U})}L_{T_\tau}(\theta).
\end{equation}
Define the Tail mass fraction as
\begin{equation}
\varepsilon_\tau
=
\frac{\mu(T_\tau)}{\mu(\mathcal{U})}.
\end{equation}
Then
\begin{equation}
\frac{\mu(P_\tau)}{\mu(\mathcal{U})}
=
1-\varepsilon_\tau,
\end{equation}
and hence
\begin{equation}
L_\mathcal{U}(\theta)
=
(1-\varepsilon_\tau)L_{P_\tau}(\theta)
+
\varepsilon_\tau L_{T_\tau}(\theta).
\end{equation}

We now bound how far the full-data optimum can be from the Head-prefix minimizer set. Let
\begin{equation}
S_{P_\tau}
=
\arg\min_\theta L_{P_\tau}(\theta),
\qquad
L_{P_\tau}^\star
=
\min_\theta L_{P_\tau}(\theta),
\end{equation}
and let
\begin{equation}
\theta_\mathcal{U}^\star\in\arg\min_\theta L_\mathcal{U}(\theta).
\end{equation}
Choose any $\bar\theta_{P_\tau}\in S_{P_\tau}$. Then
\begin{equation}
L_{P_\tau}(\bar\theta_{P_\tau})=L_{P_\tau}^\star.
\end{equation}
We want to bound the Head-prefix excess risk
\begin{equation}
L_{P_\tau}(\theta_\mathcal{U}^\star)-L_{P_\tau}(\bar\theta_{P_\tau}).
\end{equation}
Add and subtract $L_\mathcal{U}(\theta_\mathcal{U}^\star)$ and $L_\mathcal{U}(\bar\theta_{P_\tau})$:
\begin{equation}
\begin{aligned}
L_{P_\tau}(\theta_\mathcal{U}^\star)-L_{P_\tau}(\bar\theta_{P_\tau})
&=
\left[
L_{P_\tau}(\theta_\mathcal{U}^\star)-L_\mathcal{U}(\theta_\mathcal{U}^\star)
\right] \\
&\quad+
\left[
L_\mathcal{U}(\theta_\mathcal{U}^\star)-L_\mathcal{U}(\bar\theta_{P_\tau})
\right] \\
&\quad+
\left[
L_\mathcal{U}(\bar\theta_{P_\tau})-L_{P_\tau}(\bar\theta_{P_\tau})
\right].
\end{aligned}
\end{equation}
Since $\theta_\mathcal{U}^\star$ minimizes $L_\mathcal{U}$, we have
\begin{equation}
L_\mathcal{U}(\theta_\mathcal{U}^\star)-L_\mathcal{U}(\bar\theta_{P_\tau})
\le 0.
\end{equation}
Therefore,
\begin{equation}
\begin{aligned}
L_{P_\tau}(\theta_\mathcal{U}^\star)-L_{P_\tau}(\bar\theta_{P_\tau})
&\le
\left[
L_{P_\tau}(\theta_\mathcal{U}^\star)-L_\mathcal{U}(\theta_\mathcal{U}^\star)
\right] \\
&\quad+
\left[
L_\mathcal{U}(\bar\theta_{P_\tau})-L_{P_\tau}(\bar\theta_{P_\tau})
\right].
\end{aligned}
\end{equation}

Next, using the decomposition of $L_\mathcal{U}$, for any $\theta$ we have
\begin{equation}
\begin{aligned}
L_\mathcal{U}(\theta)-L_{P_\tau}(\theta)
&=
(1-\varepsilon_\tau)L_{P_\tau}(\theta)
+
\varepsilon_\tau L_{T_\tau}(\theta)
-
L_{P_\tau}(\theta) \\
&=
\varepsilon_\tau
\left(
L_{T_\tau}(\theta)-L_{P_\tau}(\theta)
\right).
\end{aligned}
\end{equation}
Equivalently,
\begin{equation}
L_{P_\tau}(\theta)-L_\mathcal{U}(\theta)
=
\varepsilon_\tau
\left(
L_{P_\tau}(\theta)-L_{T_\tau}(\theta)
\right).
\end{equation}
By the local bounded loss-contrast condition on the relevant sublevel set $\mathcal S$, there exists $M_\tau<\infty$ such that
\begin{equation}
\sup_{\theta\in\mathcal S}
\left|
L_{T_\tau}(\theta)-L_{P_\tau}(\theta)
\right|
\le
M_\tau.
\end{equation}
Assuming $\theta_\mathcal{U}^\star,\bar\theta_{P_\tau}\in\mathcal S$, this implies
\begin{equation}
\left|
L_\mathcal{U}(\theta)-L_{P_\tau}(\theta)
\right|
\le
\varepsilon_\tau M_\tau
\end{equation}
for both $\theta=\theta_\mathcal{U}^\star$ and $\theta=\bar\theta_{P_\tau}$. Hence,
\begin{equation}
L_{P_\tau}(\theta_\mathcal{U}^\star)-L_\mathcal{U}(\theta_\mathcal{U}^\star)
\le
\varepsilon_\tau M_\tau,
\end{equation}
and
\begin{equation}
L_\mathcal{U}(\bar\theta_{P_\tau})-L_{P_\tau}(\bar\theta_{P_\tau})
\le
\varepsilon_\tau M_\tau.
\end{equation}
Substituting these two bounds into the previous inequality gives
\begin{equation}
L_{P_\tau}(\theta_\mathcal{U}^\star)-L_{P_\tau}(\bar\theta_{P_\tau})
\le
2\varepsilon_\tau M_\tau.
\end{equation}
Since $\bar\theta_{P_\tau}\in S_{P_\tau}$, we have
\begin{equation}
L_{P_\tau}(\bar\theta_{P_\tau})=L_{P_\tau}^\star.
\end{equation}
Therefore,
\begin{equation}
L_{P_\tau}(\theta_\mathcal{U}^\star)-L_{P_\tau}^\star
\le
2\varepsilon_\tau M_\tau.
\end{equation}

We now invoke the same local error-bound/KL-$1/2$ condition used in Theorem~\ref{theorem_non}. Namely, there exists a constant $C_\tau>0$ such that
\begin{equation}
\operatorname{dist}(\theta,S_{P_\tau})
\le
C_\tau
\sqrt{
L_{P_\tau}(\theta)-L_{P_\tau}^\star
}
\end{equation}
for all $\theta$ in the relevant local sublevel set. Applying this inequality at $\theta=\theta_\mathcal{U}^\star$ yields
\begin{equation}
\operatorname{dist}(\theta_\mathcal{U}^\star,S_{P_\tau})
\le
C_\tau
\sqrt{
L_{P_\tau}(\theta_\mathcal{U}^\star)-L_{P_\tau}^\star
}.
\end{equation}
Using the excess-risk bound above, we obtain
\begin{equation}
\operatorname{dist}(\theta_\mathcal{U}^\star,S_{P_\tau})
\le
C_\tau
\sqrt{
2\varepsilon_\tau M_\tau
}.
\end{equation}

It remains to compute $\varepsilon_\tau$ under the exponential decay measure. For any interval $[a,b]\subseteq[0,1]$,
\begin{equation}
\mu([a,b])
=
\int_a^b w_0\IFNew^{-u}du.
\end{equation}
Since $\IFNew^{-u}=e^{-u\log\IFNew}$, we have
\begin{equation}
\begin{aligned}
\mu([a,b])
&=
w_0\int_a^b e^{-u\log\IFNew}du \\
&=
w_0
\left[
-\frac{1}{\log\IFNew}
e^{-u\log\IFNew}
\right]_a^b \\
&=
\frac{w_0}{\log\IFNew}
\left(
\IFNew^{-a}-\IFNew^{-b}
\right).
\end{aligned}
\end{equation}
Applying this formula to the full domain $\mathcal{U}=[0,1]$ gives
\begin{equation}
\mu(\mathcal{U})
=
\mu([0,1])
=
\frac{w_0}{\log\IFNew}
\left(
1-\IFNew^{-1}
\right).
\end{equation}
Applying the same formula to the Tail suffix $T_\tau=(\tau,1]$ gives
\begin{equation}
\mu(T_\tau)
=
\frac{w_0}{\log\IFNew}
\left(
\IFNew^{-\tau}-\IFNew^{-1}
\right).
\end{equation}
Therefore,
\begin{equation}
\varepsilon_\tau
=
\frac{\mu(T_\tau)}{\mu(\mathcal{U})}
=
\frac{
\frac{w_0}{\log\IFNew}
\left(
\IFNew^{-\tau}-\IFNew^{-1}
\right)
}{
\frac{w_0}{\log\IFNew}
\left(
1-\IFNew^{-1}
\right)
}.
\end{equation}
Canceling the common factor $\frac{w_0}{\log\IFNew}$ gives
\begin{equation}
\varepsilon_\tau
=
\frac{\IFNew^{-\tau}-\IFNew^{-1}}
{1-\IFNew^{-1}}.
\end{equation}
Substituting this expression for $\varepsilon_\tau$ into the distance bound yields
\begin{equation}
\operatorname{dist}(\theta_\mathcal{U}^\star,S_{P_\tau})
\le
C_\tau
\sqrt{
2M_\tau
\frac{\IFNew^{-\tau}-\IFNew^{-1}}
{1-\IFNew^{-1}}
}.
\end{equation}
This proves the first claimed bound.

We now derive the asymptotic form. For a fixed $\tau\in(0,1)$ and large $\IFNew$, we have
\begin{equation}
1-\IFNew^{-1}\to 1.
\end{equation}
Moreover, since $\tau<1$, the term $\IFNew^{-1}$ decays faster than $\IFNew^{-\tau}$, so
\begin{equation}
\IFNew^{-\tau}-\IFNew^{-1}
=
\mathcal{O}(\IFNew^{-\tau}).
\end{equation}
Hence,
\begin{equation}
\frac{\IFNew^{-\tau}-\IFNew^{-1}}
{1-\IFNew^{-1}}
=
\mathcal{O}(\IFNew^{-\tau}).
\end{equation}
Substituting this into the previous bound gives
\begin{equation}
\operatorname{dist}(\theta_\mathcal{U}^\star,S_{P_\tau})
=
\mathcal{O}
\left(
\sqrt{\IFNew^{-\tau}}
\right)
=
\mathcal{O}(\IFNew^{-\tau/2}).
\end{equation}

Finally, define the boundary imbalance factor as
\begin{equation}
\IFNew_\tau
=
\frac{w(0)}{w(\tau)}.
\end{equation}
Since $w(0)=w_0$ and $w(\tau)=w_0\IFNew^{-\tau}$, we get
\begin{equation}
\IFNew_\tau
=
\frac{w_0}{w_0\IFNew^{-\tau}}
=
\IFNew^\tau.
\end{equation}
Therefore,
\begin{equation}
\IFNew^{-\tau/2}
=
(\IFNew^\tau)^{-1/2}
=
\IFNew_\tau^{-1/2}.
\end{equation}
Thus,
\begin{equation}
\operatorname{dist}(\theta_\mathcal{U}^\star,S_{P_\tau})
=
\mathcal{O}(\IFNew_\tau^{-1/2}).
\end{equation}
This shows that the same inverse-square-root geometry from Theorem~\ref{theorem 2} also holds under the exponential long-tail model.
Finally, the same perturbation argument can be applied sequentially to multiple cardinality-sorted exponential partitions; summing the resulting prefix-to-prefix bounds via the triangle inequality yields a telescoping extension analogous to Theorem~\ref{theorem 2}.
\end{proof}

\subsection{Proof of Proposition~\ref{prop:partitioning}}
\label{app:partitioning}

We now provide the full proof of the partitioning bound. The proof has three steps. First, we bound the mismatch between the imbalanced training risk and the balanced risk inside each partition. Second, we bound the forgetting introduced by one continual-learning transition. Third, we combine these terms and derive the equal-local-imbalance rule for exponentially decaying class counts.

\begin{proof}
Let the classes be sorted by decreasing cardinality. For a partition $\mathcal C_i$, define the corresponding dataset partition as
\begin{equation}
\mathcal D^i
=
\{(x,y)\in\mathcal D:y\in\mathcal C_i\}.
\end{equation}
Let $n_c=|\mathcal D_c|$ denote the number of samples in class $c$. The local imbalance factor inside partition $i$ is
\begin{equation}
R_i
=
\frac{\max_{c\in\mathcal C_i}n_c}
{\min_{c\in\mathcal C_i}n_c}.
\end{equation}

We first compare the imbalanced training risk and the balanced risk inside a fixed partition. Define the empirical class distribution inside partition $i$ by
\begin{equation}
p_c
=
\frac{n_c}{\sum_{d\in\mathcal C_i}n_d},
\qquad
c\in\mathcal C_i.
\end{equation}
Define the balanced class distribution inside the same partition by
\begin{equation}
q_c
=
\frac{1}{|\mathcal C_i|},
\qquad
c\in\mathcal C_i.
\end{equation}
The task loss used for training on partition $i$ can be written as
\begin{equation}
\mathcal L_i^{\mathrm{task}}(\theta)
=
\sum_{c\in\mathcal C_i}p_c\mathcal L_c(\theta),
\end{equation}
where $\mathcal L_c(\theta)$ denotes the class-conditional loss. The balanced risk inside partition $i$ is
\begin{equation}
\mathcal L_i^{\mathrm{bal}}(\theta)
=
\sum_{c\in\mathcal C_i}q_c\mathcal L_c(\theta).
\end{equation}
Therefore,
\begin{equation}
\mathcal L_i^{\mathrm{task}}(\theta)-\mathcal L_i^{\mathrm{bal}}(\theta)
=
\sum_{c\in\mathcal C_i}(p_c-q_c)\mathcal L_c(\theta).
\end{equation}
Assume that class losses are locally bounded on the optimization region, i.e.,
\begin{equation}
0\le \mathcal L_c(\theta)\le B
\end{equation}
for all $c$ and all relevant $\theta$. Then
\begin{equation}
\left|
\mathcal L_i^{\mathrm{task}}(\theta)-\mathcal L_i^{\mathrm{bal}}(\theta)
\right|
\le
B
\sum_{c\in\mathcal C_i}|p_c-q_c|.
\end{equation}
Using the convention
\begin{equation}
\operatorname{TV}(p,q)
=
\frac{1}{2}
\sum_{c\in\mathcal C_i}|p_c-q_c|,
\end{equation}
we obtain
\begin{equation}
\left|
\mathcal L_i^{\mathrm{task}}(\theta)-\mathcal L_i^{\mathrm{bal}}(\theta)
\right|
\le
2B\operatorname{TV}(p,q).
\end{equation}
Equivalently, if total variation is defined without the factor $1/2$, the same bound is written with $B\operatorname{TV}(p,q)$. In the statement of the proposition, we use the normalized form. We now bound this total variation in terms of the local imbalance factor $R_i$. Let
\begin{equation}
r_c
=
\frac{p_c}{q_c}
=
|\mathcal C_i|p_c.
\end{equation}
Since $p_c$ is proportional to $n_c$ and $q_c$ is uniform, we have
\begin{equation}
\frac{\max_{c\in\mathcal C_i}r_c}
{\min_{c\in\mathcal C_i}r_c}
=
\frac{\max_{c\in\mathcal C_i}n_c}
{\min_{c\in\mathcal C_i}n_c}
=
R_i.
\end{equation}
Moreover,
\begin{equation}
\sum_{c\in\mathcal C_i}q_cr_c
=
\sum_{c\in\mathcal C_i}p_c
=
1.
\end{equation}
For distributions whose likelihood ratio $r_c=p_c/q_c$ has range at most $R_i$, the total variation satisfies the standard bound
\begin{equation}
\operatorname{TV}(p,q)
\le
\frac{\sqrt{R_i}-1}{\sqrt{R_i}+1}.
\end{equation}
For completeness, we recall the short argument. Since $\mathbb E_q[r_c]=1$ and $\max r_c/\min r_c\le R_i$, the largest possible absolute deviation from $1$ is attained by a two-level likelihood ratio, with one level below $1$ and one level above $1$. Balancing the two sides around $1$ gives lower and upper levels proportional to $1/\sqrt{R_i}$ and $\sqrt{R_i}$, which yields
\begin{equation}
\operatorname{TV}(p,q)
\le
\frac{\sqrt{R_i}-1}{\sqrt{R_i}+1}.
\end{equation}
Thus, for all relevant $\theta$,
\begin{equation}
\left|
\mathcal L_i^{\mathrm{task}}(\theta)-\mathcal L_i^{\mathrm{bal}}(\theta)
\right|
\le
2B
\frac{\sqrt{R_i}-1}{\sqrt{R_i}+1}.
\end{equation}
Absorbing the factor convention for total variation into $B$, we write the partition bias term as
\begin{equation}
B
\frac{\sqrt{R_i}-1}{\sqrt{R_i}+1}.
\end{equation}

Let
\begin{equation}
\theta_i^\star
\in
\arg\min_\theta \mathcal L_i^{\mathrm{task}}(\theta),
\qquad
\theta_i^{\mathrm{bal},\star}
\in
\arg\min_\theta \mathcal L_i^{\mathrm{bal}}(\theta).
\end{equation}
Then
\begin{equation}
\begin{aligned}
\mathcal L_i^{\mathrm{bal}}(\theta_i^\star)
&\le
\mathcal L_i^{\mathrm{task}}(\theta_i^\star)
+
B
\frac{\sqrt{R_i}-1}{\sqrt{R_i}+1}
\\
&\le
\mathcal L_i^{\mathrm{task}}(\theta_i^{\mathrm{bal},\star})
+
B
\frac{\sqrt{R_i}-1}{\sqrt{R_i}+1}
\\
&\le
\mathcal L_i^{\mathrm{bal}}(\theta_i^{\mathrm{bal},\star})
+
2B
\frac{\sqrt{R_i}-1}{\sqrt{R_i}+1}.
\end{aligned}
\end{equation}
The first and third inequalities use the risk-mismatch bound, and the second inequality uses the optimality of $\theta_i^\star$ for $\mathcal L_i^{\mathrm{task}}$. Therefore, the balanced-risk penalty caused by training on an imbalanced partition is bounded by
\begin{equation}
2B
\frac{\sqrt{R_i}-1}{\sqrt{R_i}+1}.
\end{equation}
Taking the worst partition gives the within-partition LTR bias term
\begin{equation}
2B
\max_{i\le N}
\frac{\sqrt{R_i}-1}{\sqrt{R_i}+1}.
\end{equation}

We now bound the forgetting introduced by one CL transition. At stage $i\ge2$, CLTR optimizes (Note that $i$ in this part of the proof refers to the CL training step, whereas in the previous part it denotes the class index.)
\begin{equation}
\mathcal L_i^{\mathrm{CL}}(\theta)
=
\mathcal L_i^{\mathrm{task}}(\theta)
+
\alpha_iD_{\Phi,i}(h_\theta,z_{i-1})
+
\beta_iD_{\Psi,i}(\theta,\theta_{i-1}).
\end{equation}
Assume first that the stage is optimized exactly, so that
\begin{equation}
\theta_i^\star
\in
\arg\min_\theta \mathcal L_i^{\mathrm{CL}}(\theta).
\end{equation}
Because both divergences vanish at $\theta_{i-1}$, we have
\begin{equation}
D_{\Phi,i}(h_{\theta_{i-1}},z_{i-1})=0,
\qquad
D_{\Psi,i}(\theta_{i-1},\theta_{i-1})=0.
\end{equation}
By optimality of $\theta_i^\star$,
\begin{equation}
\mathcal L_i^{\mathrm{CL}}(\theta_i^\star)
\le
\mathcal L_i^{\mathrm{CL}}(\theta_{i-1})
=
\mathcal L_i^{\mathrm{task}}(\theta_{i-1}).
\end{equation}
Expanding the left-hand side gives
\begin{equation}
\mathcal L_i^{\mathrm{task}}(\theta_i^\star)
+
\alpha_iD_{\Phi,i}(h_{\theta_i^\star},z_{i-1})
+
\beta_iD_{\Psi,i}(\theta_i^\star,\theta_{i-1})
\le
\mathcal L_i^{\mathrm{task}}(\theta_{i-1}).
\end{equation}
Rearranging,
\begin{equation}
\alpha_iD_{\Phi,i}(h_{\theta_i^\star},z_{i-1})
+
\beta_iD_{\Psi,i}(\theta_i^\star,\theta_{i-1})
\le
\mathcal L_i^{\mathrm{task}}(\theta_{i-1})
-
\mathcal L_i^{\mathrm{task}}(\theta_i^\star).
\end{equation}
Define the current-task learning pressure as
\begin{equation}
\Delta_i
=
\left[
\mathcal L_i^{\mathrm{task}}(\theta_{i-1})
-
\mathcal L_i^{\mathrm{task}}(\theta_i^\star)
\right]_+.
\end{equation}
Then
\begin{equation}
\alpha_iD_{\Phi,i}(h_{\theta_i^\star},z_{i-1})
+
\beta_iD_{\Psi,i}(\theta_i^\star,\theta_{i-1})
\le
\Delta_i.
\end{equation}
Since both divergence terms are nonnegative, we obtain
\begin{equation}
D_{\Phi,i}(h_{\theta_i^\star},z_{i-1})
\le
\frac{\Delta_i}{\alpha_i},
\end{equation}
and
\begin{equation}
D_{\Psi,i}(\theta_i^\star,\theta_{i-1})
\le
\frac{\Delta_i}{\beta_i}.
\end{equation}

Let $\mathcal C_{<i}=\bigcup_{j<i}\mathcal C_j$ denote the set of previously learned classes, and define the old-task balanced loss as
\begin{equation}
\mathcal L_{<i}^{\mathrm{bal}}(\theta)
=
\frac{1}{|\mathcal C_{<i}|}
\sum_{c\in\mathcal C_{<i}}\mathcal L_c(\theta).
\end{equation}
Assume the output-route control condition
\begin{equation}
\mathcal L_{<i}^{\mathrm{bal}}(\theta)-\mathcal L_{<i}^{\mathrm{bal}}(\theta_{i-1})
\le
c_{\Phi,i}D_{\Phi,i}(h_\theta,z_{i-1})
\end{equation}
on the relevant local region. Applying this at $\theta=\theta_i^\star$ gives
\begin{equation}
\mathcal L_{<i}^{\mathrm{bal}}(\theta_i^\star)-\mathcal L_{<i}^{\mathrm{bal}}(\theta_{i-1})
\le
c_{\Phi,i}D_{\Phi,i}(h_{\theta_i^\star},z_{i-1})
\le
\Delta_i
\frac{c_{\Phi,i}}{\alpha_i}.
\end{equation}

Similarly, assume the parameter-route control condition. Specifically, suppose $\mathcal L_{<i}^{\mathrm{bal}}$ is locally $\mathcal L_{i,\mathrm{old}}^{\mathrm{sm}}$-smooth around $\theta_{i-1}$ and that $D_{\Psi,i}$ is locally $m_{\Psi,i}$-strongly convex:
\begin{equation}
D_{\Psi,i}(\theta,\theta_{i-1})
\ge
\frac{m_{\Psi,i}}{2}
\|\theta-\theta_{i-1}\|^2.
\end{equation}
Then the same argument used in Theorem~\ref{theorem_general} gives
\begin{equation}
\mathcal L_{<i}^{\mathrm{bal}}(\theta)-\mathcal L_{<i}^{\mathrm{bal}}(\theta_{i-1})
\le
\frac{\mathcal L_{i,\mathrm{old}}^{\mathrm{sm}}}{m_{\Psi,i}}
D_{\Psi,i}(\theta,\theta_{i-1}).
\end{equation}
Applying this at $\theta=\theta_i^\star$ gives
\begin{equation}
\mathcal L_{<i}^{\mathrm{bal}}(\theta_i^\star)-\mathcal L_{<i}^{\mathrm{bal}}(\theta_{i-1})
\le
\frac{\mathcal L_{i,\mathrm{old}}^{\mathrm{sm}}}{m_{\Psi,i}}
D_{\Psi,i}(\theta_i^\star,\theta_{i-1})
\le
\Delta_i
\frac{\mathcal L_{i,\mathrm{old}}^{\mathrm{sm}}}{m_{\Psi,i}\beta_i}.
\end{equation}

Either the output route or the parameter route is sufficient to control old-task loss increase. Therefore, the sharper bound is
\begin{equation}
\mathcal L_{<i}^{\mathrm{bal}}(\theta_i^\star)-\mathcal L_{<i}^{\mathrm{bal}}(\theta_{i-1})
\le
\Delta_i
\min\left\{
\frac{c_{\Phi,i}}{\alpha_i},
\frac{\mathcal L_{i,\mathrm{old}}^{\mathrm{sm}}}{m_{\Psi,i}\beta_i}
\right\}.
\end{equation}
This is the forgetting cost of transition $i-1\to i$. Summing over all transitions gives the cumulative forgetting bound
\begin{equation}
\sum_{i=2}^{N}
\Delta_i
\min\left\{
\frac{c_{\Phi,i}}{\alpha_i},
\frac{\mathcal L_{i,\mathrm{old}}^{\mathrm{sm}}}{m_{\Psi,i}\beta_i}
\right\}.
\end{equation}

Combining the within-partition LTR bias term and the cumulative forgetting term yields
\begin{equation}
\mathcal L_{\mathrm{bal}}(\theta_N)-\mathcal L_{\mathrm{bal}}^\star
\le
2B\max_{i\le N}
\frac{\sqrt{R_i}-1}{\sqrt{R_i}+1}
+
\sum_{i=2}^{N}
\Delta_i
\min\left\{
\frac{c_{\Phi,i}}{\alpha_i},
\frac{\mathcal L_{i,\mathrm{old}}^{\mathrm{sm}}}{m_{\Psi,i}\beta_i}
\right\}
+
\mathrm{OptErr}.
\end{equation}
Here $\mathrm{OptErr}$ collects optimization errors, approximation errors, and constants arising from local boundedness assumptions. This proves the first claim.

We now derive the boundary rule for exponentially decaying class counts. In the continuous approximation, let the sorted class-frequency profile be
\begin{equation}
w(u)=w_0\IFNew^{-u},
\qquad
u\in[0,1].
\end{equation}
Let the partition boundaries be
\begin{equation}
0=u_0<u_1<\cdots<u_N=1.
\end{equation}
The local imbalance factor in partition $i$ is
\begin{equation}
R_i
=
\frac{w(u_{i-1})}{w(u_i)}.
\end{equation}
Substituting $w(u)=w_0\IFNew^{-u}$ gives
\begin{equation}
R_i
=
\frac{w_0\IFNew^{-u_{i-1}}}{w_0\IFNew^{-u_i}}
=
\IFNew^{u_i-u_{i-1}}.
\end{equation}
For fixed $N$, the forgetting term has a fixed number of transitions. Thus, the partition-boundary-dependent part of the first term is
\begin{equation}
\max_{i\le N}R_i
=
\max_{i\le N}\IFNew^{u_i-u_{i-1}}.
\end{equation}
Since $\IFNew^x$ is increasing in $x$, minimizing $\max_iR_i$ is equivalent to minimizing
\begin{equation}
\max_{i\le N}(u_i-u_{i-1})
\end{equation}
subject to
\begin{equation}
\sum_{i=1}^{N}(u_i-u_{i-1})=1.
\end{equation}
The minimizer is obtained by equalizing all interval lengths:
\begin{equation}
u_i-u_{i-1}
=
\frac{1}{N}
\end{equation}
for all $i=1,\ldots,N$. Hence
\begin{equation}
u_i=\frac{i}{N}.
\end{equation}
Substituting this into the expression for $R_i$ gives
\begin{equation}
R_i
=
\IFNew^{1/N}
\end{equation}
for every $i$. Therefore,
\begin{equation}
R_1=R_2=\cdots=R_N=\IFNew^{1/N}.
\end{equation}
This proves that, for exponentially decaying class counts and fixed $N$, the optimal boundaries equalize the local imbalance factors.

Finally, suppose the CL constants are approximately uniform across stages:
\begin{equation}
\Delta_i\approx\Delta,
\qquad
c_{\Phi,i}\approx c_\Phi,
\qquad
L_{i,\mathrm{old}}^{\mathrm{sm}}\approx L_{\mathrm{old}}^{\mathrm{sm}},
\qquad
m_{\Psi,i}\approx m_\Psi,
\qquad
\alpha_i\approx\alpha,
\qquad
\beta_i\approx\beta.
\end{equation}
Then
\begin{equation}
\sum_{i=2}^{N}
\Delta_i
\min\left\{
\frac{c_{\Phi,i}}{\alpha_i},
\frac{L_{i,\mathrm{old}}^{\mathrm{sm}}}{m_{\Psi,i}\beta_i}
\right\}
\approx
(N-1)\Delta
\min\left\{
\frac{c_\Phi}{\alpha},
\frac{L_{\mathrm{old}}^{\mathrm{sm}}}{m_\Psi\beta}
\right\}.
\end{equation}
Using the fixed-$N$ optimal boundary rule $R_i=\IFNew^{1/N}$ and representing $\Delta\min\left\{
\frac{c_\Phi}{\alpha},
\frac{L_{\mathrm{old}}^{\mathrm{sm}}}{m_\Psi\beta}
\right\}$ as effective forgetting coefficient $\eta$, the partition-selection score becomes
\begin{equation}
S(N)
=
2B
\frac{\IFNew^{1/(2N)}-1}{\IFNew^{1/(2N)}+1}
+
(N-1)\eta.
\end{equation}
Thus,
\begin{equation}
N^\star
\in
\arg\min_N
S(N).
\end{equation}
This expression captures the bias-forgetting tradeoff: increasing $N$ decreases the local LTR-bias term but increases the cumulative CL-forgetting term. Therefore, the practical rule is to use the largest number of partitions that the chosen CL method can handle without excessive forgetting.
\end{proof}

\subsection{Proof of Theorem \ref{theorem_general}}\label{proof_theorem_general}
\restatetheorem{theorem_general}{
Suppose Assumption~\ref{ass:cltr} holds on $\mathcal{N}$. Define the thresholds
$
\beta_{\min}\ :=\ \frac{L_H^{\mathrm{sm}}}{2\,m_{\Psi}},
\qquad
\alpha_{\min}\ :=\ \frac{c_{\Phi}}{2},
$
where $L_H^{\mathrm{sm}}$ is local smoothness constant of $\mathcal{L}_H$ at $\theta_H^\star$, $m_{\Psi}$ is strong-convexity constant of $D_{\Psi}(\cdot,\theta_H^\star)$, and $c_{\Phi}$ is calibration constant relating $\mathcal{L}_H$ to $D_{\Phi}$. If either $\beta\ge \beta_{\min}$ \textup{(Parameter route)} or $\alpha\ge \alpha_{\min}$ \textup{(Output route)}, then for any $\theta\in\mathcal{N}$,
\begin{equation}
 \mathcal{L}_{\mathrm{bal}}(\theta)\ \le\ \mathcal{L}_{\mathrm{CL}}(\theta)\ +\ \tfrac12\,\mathcal{L}_H(\theta_H^\star).
\end{equation}
In words: if either $\beta\ge \beta_{\min}$ or $\alpha\ge \alpha_{\min}$, then the CL objective $\mathcal{L}_{\mathrm{CL}}(\theta)$ upper-bounds the balanced loss $\mathcal{L}_{\mathrm{bal}}(\theta)$ up to an additive $\theta$-independent constant.}

\textit{Proof Sketch.}  The proof first rewrites the balanced loss as the sum of a Tail term, a Head-loss difference term, and a constant by adding and subtracting $\mathcal{L}_H(\theta_H^\star)$. The Tail term $\tfrac12 \mathcal{L}_T(\theta)$ is upper-bounded by the task loss $\mathcal{L}_{\mathrm{task}}(\theta)$ and hence by $\mathcal{L}_{\mathrm{CL}}(\theta)$. For the parameter route, local smoothness of $\mathcal{L}_H$ and strong convexity of $D_{\Psi}$ are combined to bound the Head-loss difference in terms of $D_{\Psi}$, which appears in $\mathcal{L}_{\mathrm{CL}}$ when $\beta\ge\beta_{\min}$. For the output route, the calibration inequality bounds the same Head-loss difference in terms of $D_{\Phi}$, controlled by $\mathcal{L}_{\mathrm{CL}}$ when $\alpha\ge\alpha_{\min}$. Adding these bounds and the constant yields the dominance relation in Eq. \ref{eq:dominance}.

\begin{proof}
We first prove the dominance inequality Eq. \ref{eq:dominance}. Recall
\begin{equation}
\mathcal{L}_{\mathrm{bal}}(\theta):=\tfrac12 \mathcal{L}_T(\theta)+\tfrac12 \mathcal{L}_H(\theta),
\end{equation}
and add/subtract $\mathcal{L}_H(\theta_H^\star)$:
\begin{equation}\label{eq:split-app}
\mathcal{L}_{\mathrm{bal}}(\theta)
=\tfrac12 \mathcal{L}_T(\theta)+\tfrac12\big(\mathcal{L}_H(\theta)-\mathcal{L}_H(\theta_H^\star)\big)+\tfrac12 \mathcal{L}_H(\theta_H^\star).
\end{equation}
Since $\mathcal{L}_{\mathrm{task}}\equiv \mathcal{L}_T$ and the loss is non-negative, we have
\begin{equation}\label{first_ineq-app}
\tfrac12 \mathcal{L}_T(\theta)\ \le\ \mathcal{L}_{\mathrm{task}}(\theta),
\end{equation}
and combining Eq. \ref{eq:split-app} and Eq. \ref{first_ineq-app} gives
\begin{equation}\label{first_ineq2-app}
\mathcal{L}_{\mathrm{bal}}(\theta)
\ \le\ \mathcal{L}_{\mathrm{task}}(\theta)+\tfrac12\big(\mathcal{L}_H(\theta)-\mathcal{L}_H(\theta_H^\star)\big)+\tfrac12 \mathcal{L}_H(\theta_H^\star).
\end{equation}

We now bound the term $\tfrac12\big(\mathcal{L}_H(\theta)-\mathcal{L}_H(\theta_H^\star)\big)$ in the two cases of the theorem.

\paragraph{Parameter route.}
Assume $\beta\ge\beta_{\min}$ and use Eq. \ref{eq:head-smooth} and Eq. \ref{eq:psi-strong}. For all $\theta\in\mathcal{N}$,
\begin{align}
\mathcal{L}_H(\theta)-\mathcal{L}_H(\theta_H^\star)
&\le \tfrac{L_H^{\mathrm{sm}}}{2}\,\|\theta-\theta_H^\star\|^2 \\
&\le \tfrac{L_H^{\mathrm{sm}}}{2}\,\frac{2}{m_{\Psi}}\,D_{\Psi}(\theta,\theta_H^\star) \\
&\le \frac{L_H^{\mathrm{sm}}}{m_{\Psi}} D_{\Psi}(\theta,\theta_H^\star),
\end{align}
where the second inequality uses $\|\theta-\theta_H^\star\|^2 \le \frac{2}{m_{\Psi}}D_{\Psi}(\theta,\theta_H^\star)$ from Eq. \ref{eq:psi-strong}. Multiplying by $\tfrac12$ yields
\begin{equation}\label{eq:head-dom-psi-app}
\tfrac12\big(\mathcal{L}_H(\theta)-\mathcal{L}_H(\theta_H^\star)\big)
\ \le\ \frac{L_H^{\mathrm{sm}}}{2m_{\Psi}}\,D_{\Psi}(\theta,\theta_H^\star).
\end{equation}
Since $\beta\ge \beta_{\min}={L_H^{\mathrm{sm}}}/{(2m_{\Psi})}$, we deduce
\begin{equation}\label{half_beta-app}
\tfrac12\big(\mathcal{L}_H(\theta)-\mathcal{L}_H(\theta_H^\star)\big)
\ \le\ \beta\,D_{\Psi}(\theta,\theta_H^\star).
\end{equation}
Substituting Eq. \ref{half_beta-app} into Eq. \ref{first_ineq2-app} gives
\begin{equation}\label{last_param-app}
\mathcal{L}_{\mathrm{bal}}(\theta)
\ \le\ \mathcal{L}_{\mathrm{task}}(\theta)+\beta\,D_{\Psi}(\theta,\theta_H^\star)+\tfrac12 \mathcal{L}_H(\theta_H^\star).
\end{equation}
By definition,
$
\mathcal{L}_{\mathrm{CL}}(\theta)
=\mathcal{L}_{\mathrm{task}}(\theta)+\alpha D_{\Phi}\!\big(h_{\theta}(\cdot), z\big)+\beta D_{\Psi}(\theta,\theta_H^\star)
\ \ge\ \mathcal{L}_{\mathrm{task}}(\theta)+\beta D_{\Psi}(\theta,\theta_H^\star),
$
since $\alpha D_{\Phi}\ge 0$. Combining this with Eq. \ref{last_param-app} yields
\begin{equation}\label{proof_param-app}
\mathcal{L}_{\mathrm{bal}}(\theta)
\ \le\ \mathcal{L}_{\mathrm{CL}}(\theta)+\tfrac12 \mathcal{L}_H(\theta_H^\star),
\end{equation},
which is Eq. \ref{eq:dominance} in the parameter route.

\paragraph{Output route.}
Assume $\alpha\ge\alpha_{\min}$ and use Eq. \ref{eq:phi-calib}. For all $\theta\in\mathcal{N}$,
\begin{equation}\label{eq:head-dom-phi-app}
\mathcal{L}_H(\theta)-\mathcal{L}_H(\theta_H^\star)
\ \le\ c_{\Phi}\,D_{\Phi}\!\big(h_{\theta}(\cdot),z_H^\star\big).
\end{equation}
Multiplying by $\tfrac12$ gives
\begin{equation}\label{eq:head-dom-phi-half-app}
\tfrac12\big(\mathcal{L}_H(\theta)-\mathcal{L}_H(\theta_H^\star)\big)
\ \le\ \tfrac{c_{\Phi}}{2}\,D_{\Phi}\!\big(h_{\theta}(\cdot),z_H^\star\big).
\end{equation}
Since $\alpha\ge\alpha_{\min}=c_{\Phi}/2$, we have
\begin{equation}\label{half_alpha-app}
\tfrac12\big(\mathcal{L}_H(\theta)-\mathcal{L}_H(\theta_H^\star)\big)
\ \le\ \alpha\,D_{\Phi}\!\big(h_{\theta}(\cdot),z_H^\star\big).
\end{equation}
Substituting Eq. \ref{half_alpha-app} into Eq. \ref{first_ineq2-app} yields
\begin{equation}\label{last_route-app}
\mathcal{L}_{\mathrm{bal}}(\theta)
\ \le\ \mathcal{L}_{\mathrm{task}}(\theta)+\alpha\,D_{\Phi}\!\big(h_{\theta}(\cdot),z_H^\star\big)+\tfrac12 \mathcal{L}_H(\theta_H^\star).
\end{equation}
Again,
$
\mathcal{L}_{\mathrm{CL}}(\theta)
=\mathcal{L}_{\mathrm{task}}(\theta)+\alpha D_{\Phi}\!\big(h_{\theta}(\cdot), z\big)+\beta D_{\Psi}(\theta,\theta_H^\star)
\ \ge\ \mathcal{L}_{\mathrm{task}}(\theta)+\alpha D_{\Phi}\!\big(h_{\theta}(\cdot),z_H^\star\big),
$
so Eq. \ref{last_route-app} implies
\begin{equation}\label{proof_route-app}
\mathcal{L}_{\mathrm{bal}}(\theta)
\ \le\ \mathcal{L}_{\mathrm{CL}}(\theta)+\tfrac12 \mathcal{L}_H(\theta_H^\star),
\end{equation}
which is again Eq. \ref{eq:dominance}, now for the output route.

Combining Eq. \ref{proof_param-app} and Eq. \ref{proof_route-app} shows that if either $\beta \ge \beta_{\min}$ or $\alpha\ge \alpha_{\min}$, then for all $\theta\in\mathcal{N}$,
\begin{equation}\label{eq:dominance-final-app}
\mathcal{L}_{\mathrm{bal}}(\theta)\ \le\ \mathcal{L}_{\mathrm{CL}}(\theta)+\tfrac12 \mathcal{L}_H(\theta_H^\star),
\end{equation}
which completes the proof.
\end{proof}

\subsection{Stronger Statement (Special Case): Exact Minimizer Equivalence}\label{eq_convex}
For completeness, we recall a special case (useful for exposition) where CL \emph{exactly} optimizes the balanced objective.

\begin{proposition}[Exact equality up to a constant]\label{prop:exact-balanced}
Let $\Psi=\mathcal{L}_H$ and assume $\mathcal{L}_H$ is convex on $\mathcal{N}$. Set $\alpha=0$, $\beta=\tfrac12$, $\mathcal{L}_{\mathrm{task}}=\tfrac12 \mathcal{L}_T$, and $z$ arbitrary. Then
$
\mathcal{L}_{\mathrm{CL}}(\theta)=\tfrac12 \mathcal{L}_T(\theta)+\tfrac12 D_{\mathcal{L}_H}(\theta,\theta_H^\star)
=\mathcal{L}_{\mathrm{bal}}(\theta)-\tfrac12 \mathcal{L}_H(\theta_H^\star),
$
so $\mathcal{L}_{\mathrm{CL}}$ and $\mathcal{L}_{\mathrm{bal}}$ have identical gradients and the same minimizers.
\end{proposition}

\begin{proof}
Since $\nabla \mathcal{L}_H(\theta_H^\star)=0$, $D_{\mathcal{L}_H}(\theta,\theta_H^\star)=\mathcal{L}_H(\theta)-\mathcal{L}_H(\theta_H^\star)$. Add and subtract $\tfrac12 \mathcal{L}_H(\theta_H^\star)$ and regroup terms.
\end{proof}

\section{Implementation Details}\label{ID}
All our experiments were conducted using the PyTorch framework. We build upon publicly available implementations of Learning without Forgetting (LwF), Elastic Weight Consolidation (EWC), Gradient Projection Memory (GPM), Scaled Gradient Projection (SGP), FOSTER, TPL, and DualPrompt, adapting their training protocols where necessary for long-tailed recognition. The corresponding method-specific modifications and hyperparameters are detailed below.
\footnote{The code for the algorithms was obtained and modified from various open-source repositories.\\
\textit{ \href{https://github.com/ngailapdi/LWF}{https://github.com/ngailapdi/LWF}
\\
\href{https://github.com/shivamsaboo17/Overcoming-Catastrophic-forgetting-in-Neural-Networks}
{https://github.com/shivamsaboo17/Overcoming-Catastrophic-forgetting-in-Neural-Networks}
\\
\href{https://github.com/sahagobinda/GPM}{https://github.com/sahagobinda/GPM} \\
\href{https://github.com/sahagobinda/SGP}{https://github.com/sahagobinda/SGP} \\
\href{https://github.com/G-U-N/ECCV22-FOSTER}{https://github.com/G-U-N/ECCV22-FOSTER} \\
\href{https://github.com/linhaowei1/TPL}{https://github.com/linhaowei1/TPL} \\
\href{https://github.com/google-research/l2p}{https://github.com/google-research/l2p}
} }
The specifics of each algorithm's implementation are summarized in Table~\ref{ta1}. The parameters for each algorithm, such as learning rate (LR), optimizer, momentum, LR scheduler, CL loss weight, number of epochs, and number of steps, are detailed there. We mainly follow the original hyperparameter settings recommended for each method, including their training schedules (e.g., number of epochs), and only deviate when strictly necessary (e.g., for algorithm-specific grid search), in order to ensure fair and faithful comparisons.

\begin{table}[t]
\centering
\caption{Implementation details of the considered algorithms for the LTR benchmark.}\label{ta1}
\small
\resizebox{\linewidth}{!}{
\begin{tabular}{cccccccc}
\hline
\textbf{Algorithm} & \textbf{LR} & \textbf{Opt.} & \textbf{Momentum} & \textbf{LR Scheduler} & \textbf{CL Loss Weight} & \textbf{Epochs} & \textbf{Steps} \\ \hline
LwF                & 0.001                  & SGD                & 0.9               & --                    & 0.01                    & 5   & 2 \\
EWC                & 0.01                   & SGD                & 0.9               & --                    & 10                      & 90  & 2 \\
Modified EWC       & 0.01                   & SGD                & 0.9               & --                    & 1000                    & 90  & 2 \\
GPM                & 0.001                  & SGD                & 0                 & Cosine Anneal LR      & --                      & 100 & 2 \\
SGP                & 0.001                  & SGD                & 0.9                 & Cosine Anneal LR      & --                      & 100 & 2 \\
FOSTER (phase 1)   & 0.1                    & SGD                & 0.9               & Cosine Anneal LR      & --                      & 170 & 4 \\
FOSTER (phase 2)   & 0.1                    & SGD                & 0                 & Cosine Anneal LR      & --                      & 130 & 4 \\
TPL                & 0.01                   & SGD                & 0.8               & Cosine Anneal LR      & --                      & 50  & 4 \\
\hline
\end{tabular}}
\end{table}

Note that for SGP, the algorithm-specific hyperparameters are obtained through grid search as follows: $gpm_{\text{eps}} = 0.9$ and $gpm_{\text{eps-inc}} = 0.004$.

We evaluate six adaptation strategies for a frozen ViT-B/32 CLIP backbone (OpenAI pretrained weights) on CIFAR100-LT under imbalance factors $\IFNew\in\{10,50,100\}$.

\textbf{Initialization Strategy.} Algorithm~\ref{alg:CLTR} permits initialization using either the Head partition $\mathcal{D}^{1}$ or the full long-tailed dataset $\mathcal{D}$. This flexibility is motivated by Theorems~\ref{theorem1}, \ref{theorem_non}, and \ref{theorem 2}, which establish that, under their respective assumptions, the full-dataset minimizer lies within an $O(1/\sqrt{IF})$ neighborhood of the Head-only solution or minimizer set. Thus, full-dataset training offers a theoretically motivated Head-dominated initialization.

\textbf{Zero-shot CLIP} requires no training: we compute cosine similarity between image embeddings and text embeddings produced by the fixed template ``\texttt{a photo of a \{class\}}'', and classify by nearest neighbor.

\textbf{Linear probe} freezes the ViT-B/32 encoder and trains only a linear $512\!\to\!100$ classification head with AdamW ($\text{lr}=10^{-3}$, $\lambda=10^{-2}$, 50 epochs, cosine schedule).

\textbf{Full fine-tuning} optimizes all ViT-B/32 parameters jointly with a linear head using AdamW ($\text{lr}=3\!\times\!10^{-5}$, $\lambda=10^{-2}$, 20 epochs, cosine schedule with 200-step warmup).

\textbf{Prompt learning} follows the clip-zs-prompting style~\cite{bahng2022exploring}: a learnable $30$-pixel visual-prompt border is prepended to every image while the ViT encoder and text branch remain frozen, and is trained with SGD ($\text{lr}=40$, momentum $0.9$, cosine schedule, 1000-step warmup, 10 epochs).

\textbf{Naive sequential} applies the same full fine-tuning procedure sequentially: first on the Head partition (60 classes), then on the Tail partition (40 classes), with no replay buffer, intentionally inducing catastrophic forgetting.

\paragraph{CLTR (DualPrompt).} We adapt the DualPrompt architecture~\cite{wang2022dualprompt} using the OpenAI-pretrained CLIP ViT-B/32 backbone. The visual encoder is frozen except for its LayerNorm parameters. We employ G-prompts (length 5, layers 0--1) and E-prompts (pool size 10, length 5, top-$k=1$, layers 2--4), together with a trainable cosine classifier initialized from CLIP text embeddings of class names. Optimization uses AdamW with cosine learning-rate scheduling and a 200-step warmup, balanced-softmax classification loss, and prompt-similarity regularization. During Tail adaptation, the method incorporates exemplar replay from the Head partition and knowledge distillation from a frozen Head-stage teacher. At inference, logits are averaged over the original and horizontally flipped images, followed by validation-calibrated logit adjustment. Table~\ref{tab:fm_impl} summarizes the key hyperparameters.

\begin{table}[h]
\centering
\caption{Implementation details for foundation-model adaptation experiments on CIFAR100-LT.}
\label{tab:fm_impl}
\small
\begin{tabular}{lcccccc}
\hline
\textbf{Method} & \textbf{Backbone} & \textbf{Trainable params} & \textbf{Opt.} & \textbf{LR} & \textbf{Epochs} & \textbf{Tasks} \\
\hline
Zero-shot CLIP   & ViT-B/32 (frozen) & 0                  & --    & --                   & -- & -- \\
Linear probe     & ViT-B/32 (frozen) & head only          & AdamW & $10^{-3}$            & 50 & 1  \\
Full fine-tuning & ViT-B/32          & all                & AdamW & $3\!\times\!10^{-5}$ & 20 & 1  \\
Prompt learning  & ViT-B/32 (frozen) & visual prompts     & SGD   & 40                   & 10 & 1  \\
Naive sequential & ViT-B/32          & all                & AdamW & $3\!\times\!10^{-5}$ & 20 & 2  \\
\shortstack[c]{DualPrompt \\ $\quad$} & \shortstack[c]{ViT-B/32 \\(LN adapted)} & \shortstack[c]{G/E prompts + LN\\+ cosine head} & AdamW & $3\!\times 10^{-3}$ & 40--60 & 2 \\
\hline
\end{tabular}
\end{table}

\textbf{Formal Verification and Resources.}
Lean~4 formalizations of our main theoretical results, together with implementations of selected top-performing CLTR configurations (SGP and DualPrompt-based), are available at \url{https://github.com/MahdiyarMM/CLTR}.

\section{Datasets}\label{app_dataset}
\begin{figure}[t]
    \centering
    \begin{minipage}[b]{0.32\linewidth}
        \centering
        \includegraphics[width=0.95\linewidth]{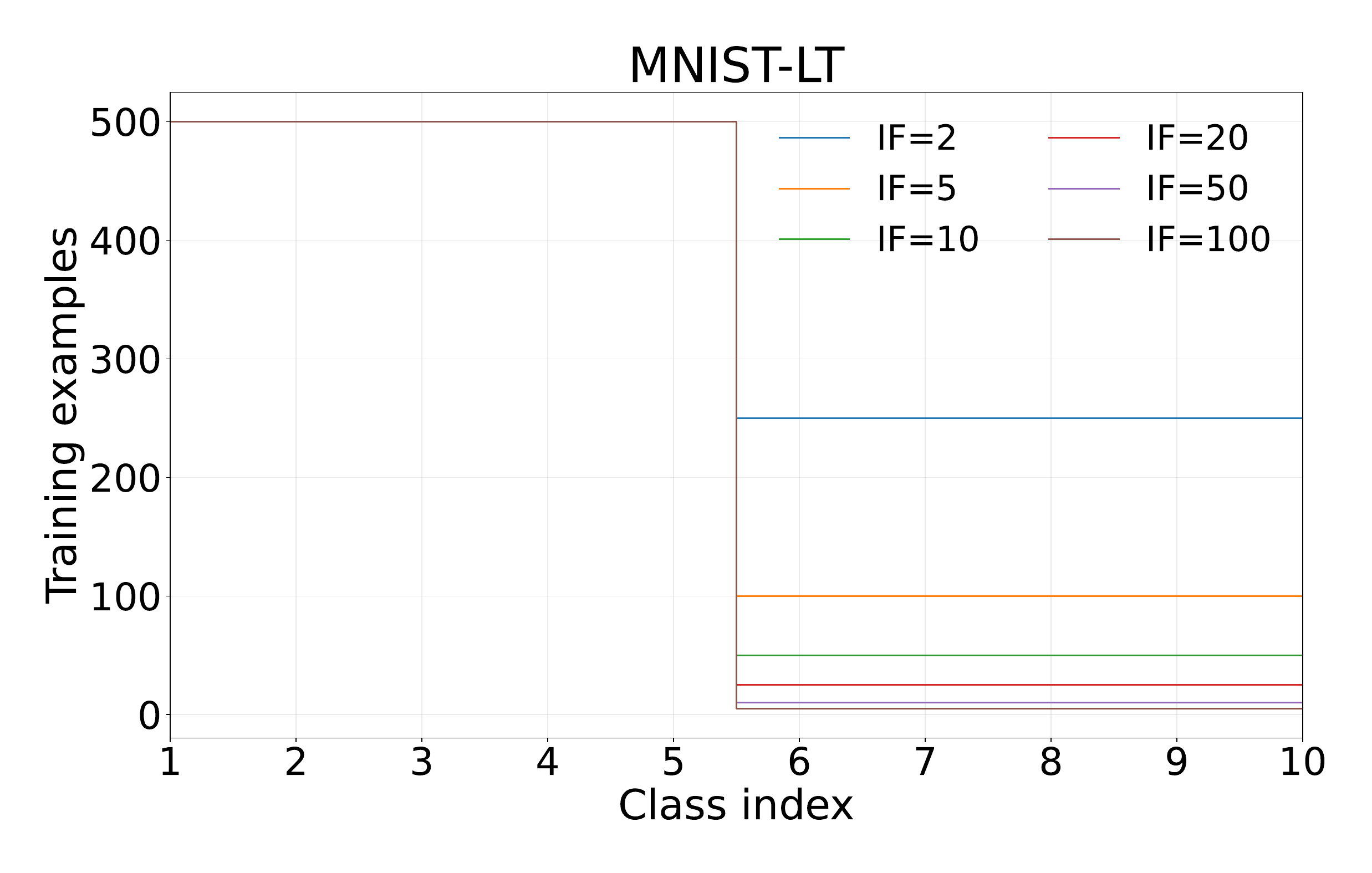}
        (a)
    \end{minipage}
    \hfill
    \begin{minipage}[b]{0.32\linewidth}
        \centering
        \includegraphics[width=0.95\linewidth]{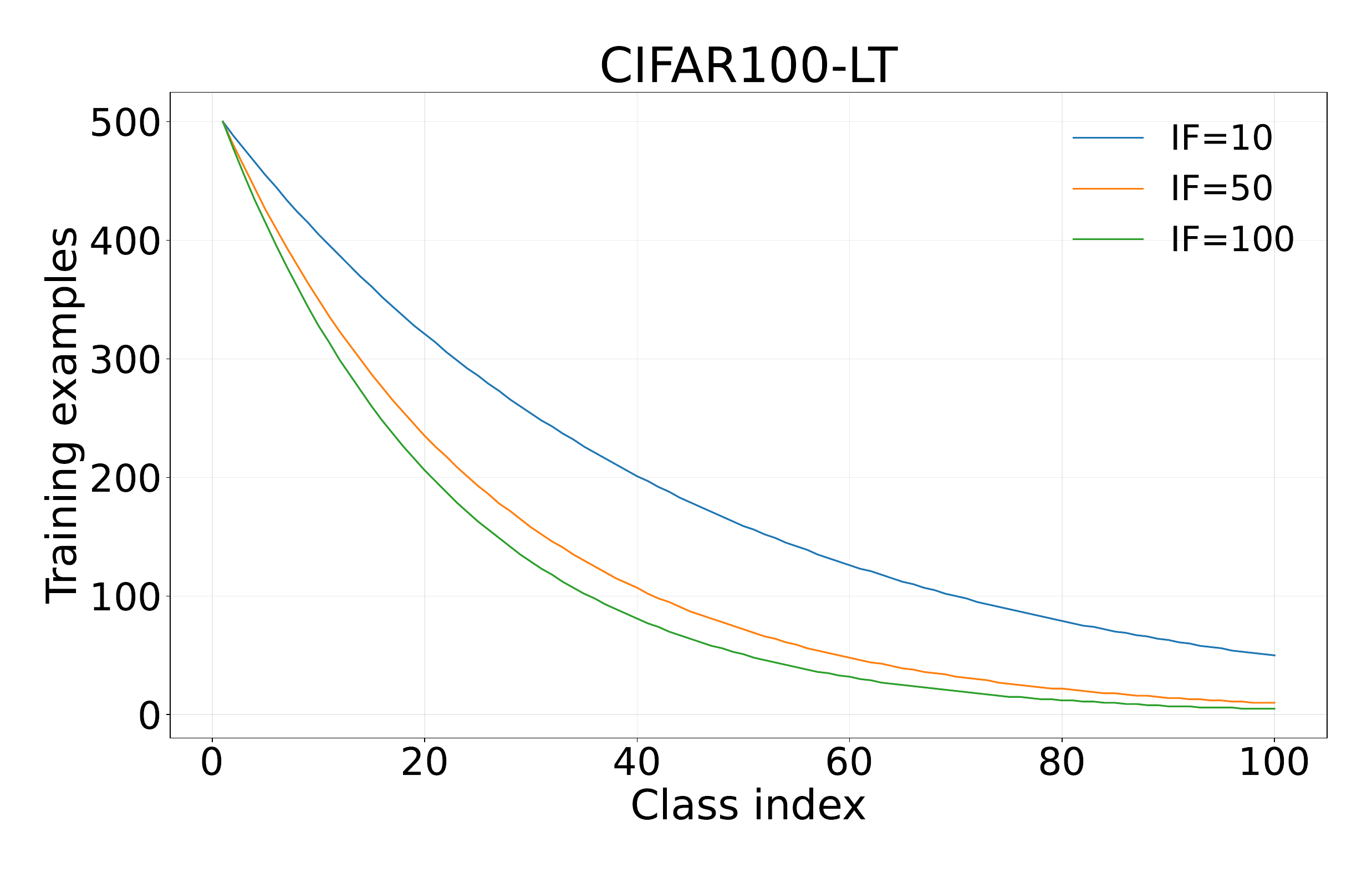}
        (b)
    \end{minipage}
    \hfill
    \begin{minipage}[b]{0.32\linewidth}
        \centering
        \includegraphics[width=0.95\linewidth]{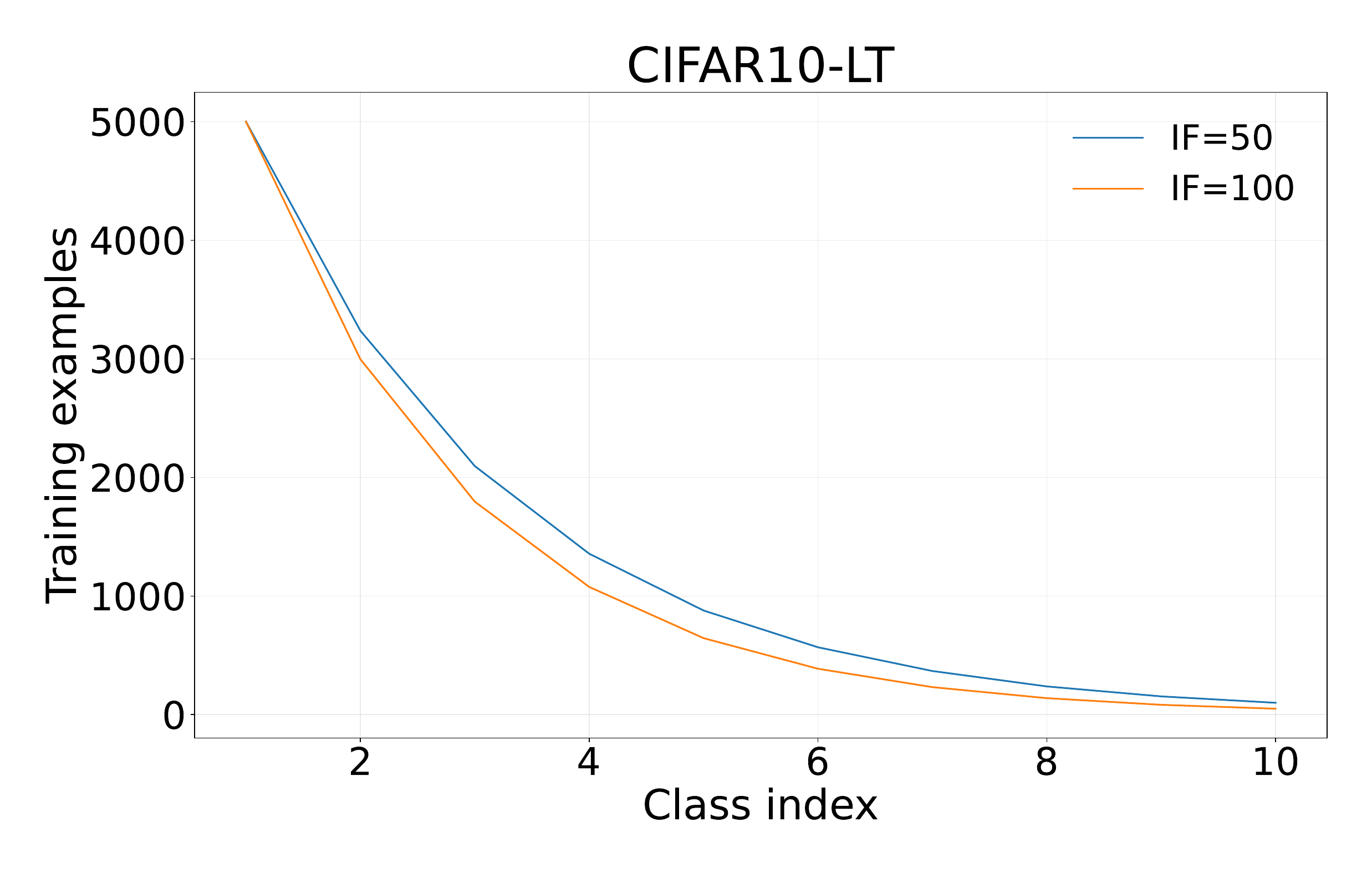}
        (c)
    \end{minipage}
    \vspace{1em} %
    \hspace{2cm}
    \begin{minipage}[b]{0.32\linewidth}
        \centering
        \includegraphics[width=0.95\linewidth]{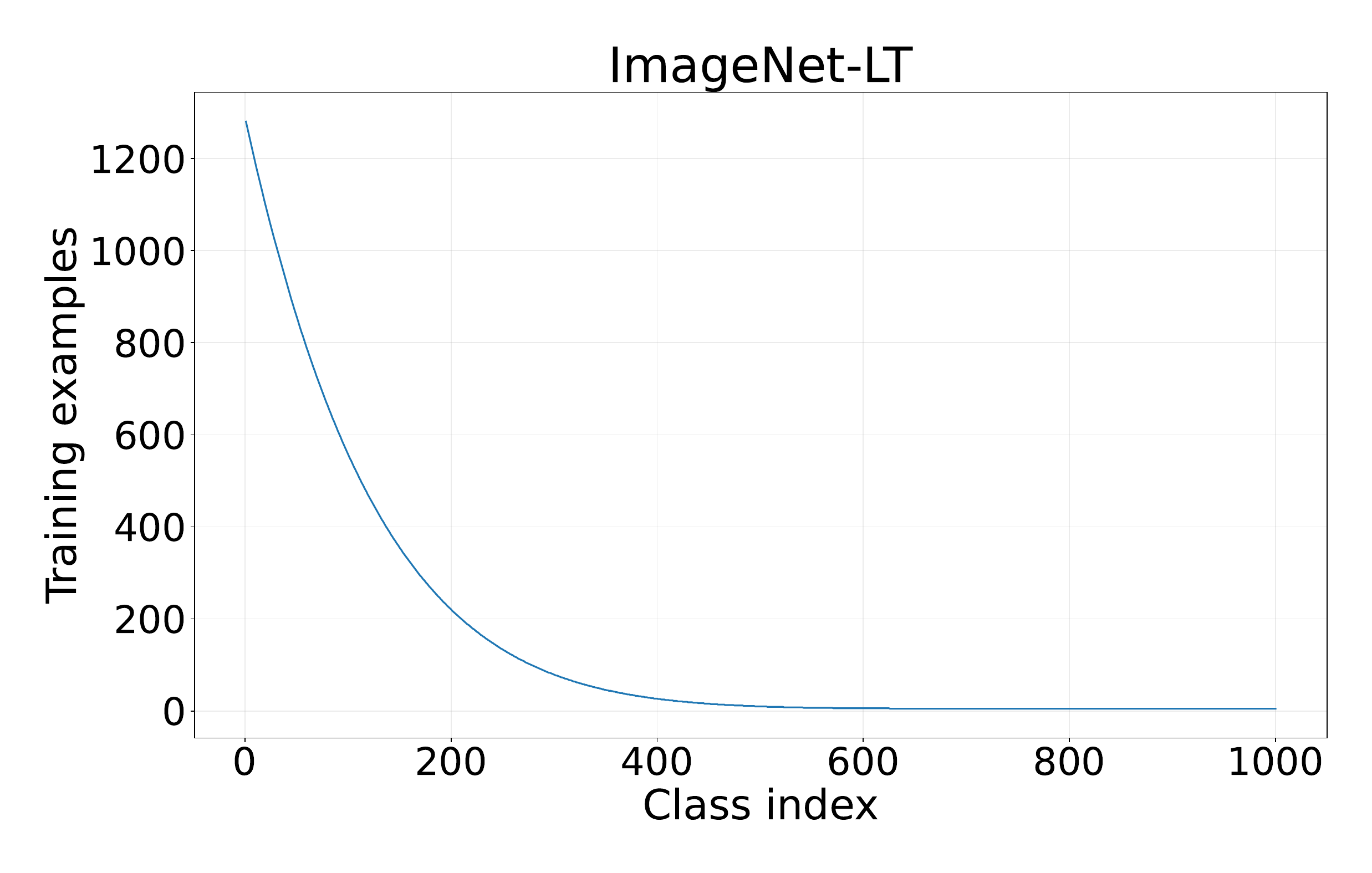}
        (d)
    \end{minipage}
    \hfill
    \begin{minipage}[b]{0.32\linewidth}
        \centering
        \includegraphics[width=0.95\linewidth]{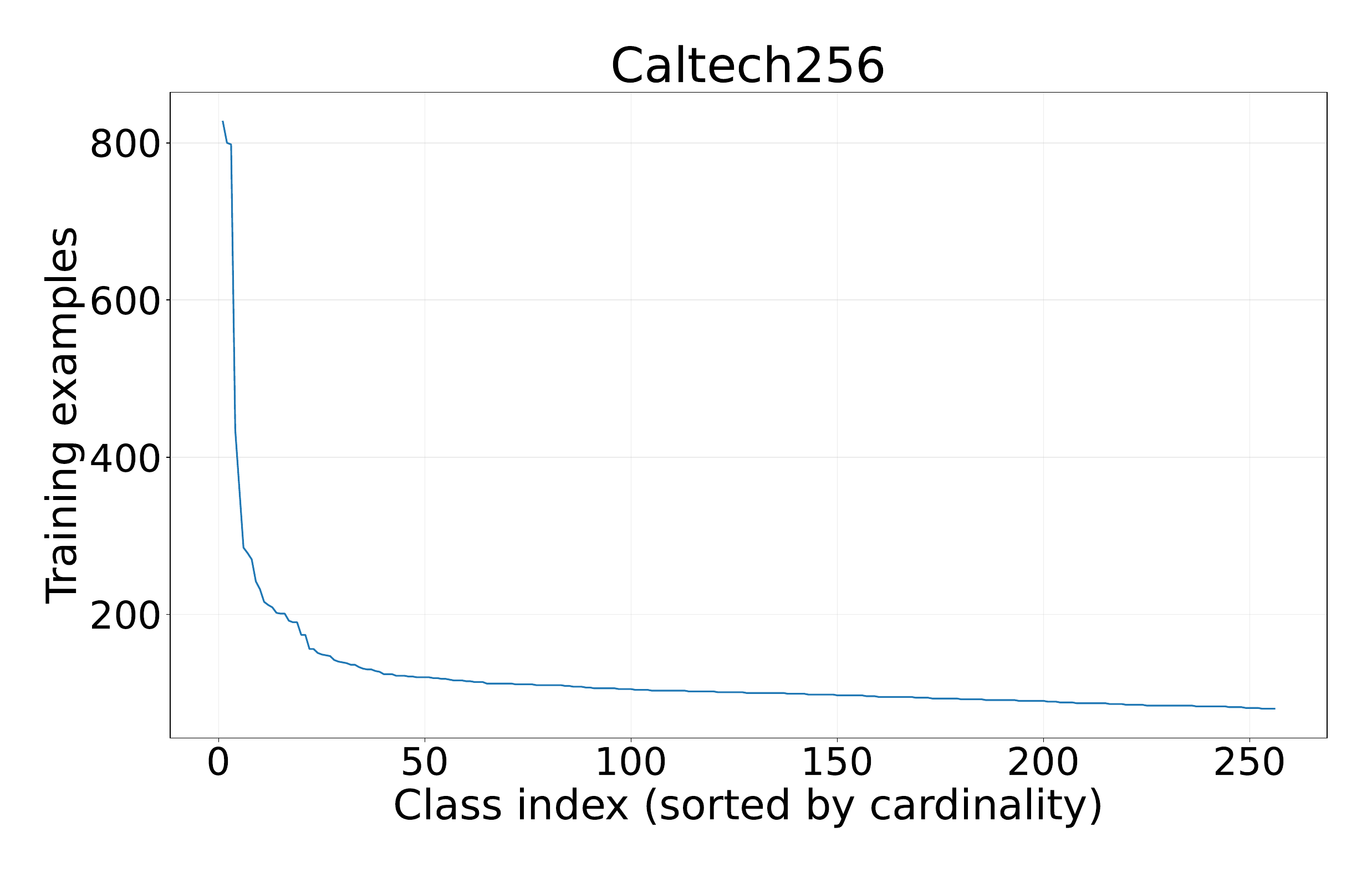}
        (e)
    \end{minipage}
    \hspace{2cm}

    \caption{Class cardinality of (a) MNIST-LT, (b) CIFAR100-LT, (c) CIFAR10-LT, (d) ImageNet-LT, and (e) Caltech256.}
    \label{fig:a1}
\end{figure}

Fig. \ref{fig:a1} illustrates the distribution of samples among different classes and the division of the dataset into the Head and Tail sections. For exponentially decaying class distributions, we follow Proposition~\ref{prop:partitioning} to partition each dataset into Head and Tail subsets.

\section{Additional Discussions} \label{app:disc}

\subsection{Comparison with Methods Using Additional Techniques}
\label{app:bells_whistles}

Table~\ref{tab:appendix_bells_whistles} provides an additional comparison with LTR methods that use extra techniques such as aggressive data augmentation, ensemble learning, multi-expert architectures, or self-supervised pretraining. Since these methods do not follow the same protocol as our main comparison, we report them separately in the appendix for reference rather than including them in the main benchmark table.

\begin{table*}[ht!]
\vspace{-2mm}
\caption{Performance comparison on CIFAR100-LT and ImageNet-LT. The blue rows show CLTR variants. The gray block reports methods with additional bells and whistles and is included for context, since these methods may not follow the same protocol.}
\label{tab:appendix_bells_whistles}
\scriptsize
\setlength{\tabcolsep}{4pt}
\centering
\resizebox{0.72\linewidth}{!}{
\begin{tabular}{lcccc}
\hline
\multirow{2}{*}{Method}
& \multirow{2}{*}{ImageNet-LT}
& \multicolumn{3}{c}{CIFAR100-LT} \\
& & $\IFNew=100$ & $\IFNew=50$ & $\IFNew=10$ \\
\hline\hline

\rowcolor{lightblue}
CLTR (LwF)
& 47.6 & 45.1 & 49.3 & 58.7 \\

\rowcolor{lightblue}
CLTR (EWC)
& 48.9 & 44.4 & 50.3 & 58.8 \\

\rowcolor{lightblue}
CLTR (Modified EWC)
& 49.1 & 45.9 & 51.0 & 60.7 \\

\rowcolor{lightblue}
CLTR (GPM)
& 51.7 & 48.3 & 54.7 & 64.7 \\

\rowcolor{lightblue}
CLTR (FOSTER)
& 52.7 & 48.7 & 54.4 & 63.6 \\

\rowcolor{lightblue}
CLTR (SGP)
& \underline{53.2} & \underline{50.7} & \textbf{58.0} & \underline{67.2} \\

\rowcolor{lightblue}
CLTR (TPL)
& \textbf{53.9} & 48.4 & 54.0 & 62.1 \\

\hline
\multicolumn{5}{c}{\textcolor{gray}{\textit{Methods with bells and whistles}}} \\

\textcolor{gray}{RIDE \cite{b1wang2020long}}
& \textcolor{gray}{56.1}
& \textcolor{gray}{49.1}
& \textcolor{gray}{NR}
& \textcolor{gray}{NR} \\

\textcolor{gray}{ACE \cite{b2cai2021ace}}
& \textcolor{gray}{56.6}
& \textcolor{gray}{49.4}
& \textcolor{gray}{50.7}
& \textcolor{gray}{NR} \\

\textcolor{gray}{SSD \cite{ li2021self}}
& \textcolor{gray}{56.0}
& \textcolor{gray}{46.0}
& \textcolor{gray}{50.5}
& \textcolor{gray}{62.3} \\

\textcolor{gray}{PaCo \cite{17cui2021parametric}}
& \textcolor{gray}{54.0}
& \textcolor{gray}{52.0}
& \textcolor{gray}{56.0}
& \textcolor{gray}{64.2} \\

\hline
\end{tabular}}
\end{table*}

\subsection{Long-Tailed Class-Incremental Learning}\label{app:cil}
Few prior works have attempted to address the problem of class-incremental learning when the data is heavily imbalanced. A novel replay method called Partitioning Reservoir Sampling (PRS) is proposed in \cite{kim2020imbalanced}. This method dedicates a sufficient amount of memory to Tail classes in order to avoid catastrophic forgetting in minority classes.
In \cite{liu2022long}, this problem is addressed in two different setups, ordered and shuffled. In the ordered scenario the number of samples in each new task is less than in previous tasks, while in the shuffled scenario, the size of classes is completely random. They propose a two-stage learning method utilizing a learnable weight scaling layer for reducing the bias due to data imbalance. Finally, in \cite{liu2022open}, OLTR++ is proposed which is a unified algorithm that integrates imbalanced classification, few-shot learning, open-set recognition, and active learning through dynamic meta-embedding and memory association. Note that none of the above works attempt to employ CL as a solution for LTR scenarios.

Here, we compare the performance of CLTR in addressing the LT-CIL problem with the prior state-of-the-art solutions in the area.
Following the experimental setup in \cite{liu2022long,hou2019learning,douillard2020podnet}, the models are first trained on the largest 50 classes (Head), then, the remaining classes are learned incrementally in 5 or 10 consecutive tasks (Tail) with an equal number of new classes in each new task, from the largest subset to the smallest subset of the dataset.
 We apply our method in this setting on the CIFAR100-LT dataset and compare its performance with the prior works, as presented in Table \ref{tab:cil}. The results demonstrate that CLTR outperforms prior methods in both 5- and 10-task settings by considerable margins of 2.5\% and 1.6\%, respectively.%

\begin{table}[ht!]
\caption{The performance of  CLTR on Ordered LT-CIL Benchmark for CIFAR100-LT}\label{tab:cil}
\scriptsize
\setlength{\tabcolsep}{3pt}
  \centering
  \begin{minipage}[t]{\linewidth}
    \centering
\resizebox{0.4\linewidth}{!}{\begin{tabular}{lcc}
\hline
\addlinespace[0.2em]
\multirow{2}{*}{Method} & \multicolumn{2}{c}{Tasks} \\
& 5 & 10   \\ \hline\hline
EEIL \cite{castro2018end} &38.5 &37.5 \\
EEIL+2sLWS \cite{liu2022long} &39.0& 37.6 \\
LUCIR \cite{hou2019learning} &42.7 &42.2 \\

PODNET \cite{douillard2020podnet} &44.1 &44.0 \\
PODNET+2sLWS \cite{liu2022long} &44.4 &44.4            \\
LUCIR+2sLWS \cite{liu2022long} &\underline{45.9} &\underline{45.7}\\
\rowcolor{lightblue}
CLTR (TPL)            & \textbf{48.4}              & \textbf{47.3}      \\\hline
\end{tabular}}
  \end{minipage}%
  \hfill
\end{table}

\subsection{Naturally Skewed Dataset}\label{app:caltech}
In LTR benchmarks, datasets are modified to exhibit a skewed distribution of samples among various classes. However, such imbalanced class distributions are naturally observed in real-world data as well \cite{alshammari2022long}. To evaluate the efficacy of CL techniques on non-LTR benchmark datasets, we utilize the Caltech256 dataset \cite{griffin2007caltech}, which consists of 256 distinct classes representing everyday objects. The largest class comprises 827 samples, while the smallest class contains only 80 samples, exhibiting an $\IFNew$ of over 10.
Here, we employ the CLTR and compare its performance to the state-of-the-art methods on this dataset for object classification. The results are presented in Table \ref{tab:caltech}.
We observe that CLTR outperforms the previous method on this dataset, demonstrating the strong potential of using CL in dealing with long-tailed real-world datasets.

\begin{table}[ht!]
\caption{The performance of CLTR on Caltech256.}\label{tab:caltech}
\scriptsize
\setlength{\tabcolsep}{3pt}
  \centering
  \begin{minipage}{\linewidth}
    \centering

\resizebox{0.4\linewidth}{!}{\begin{tabular}{lll}
\hline
\addlinespace[0.2em]
\multirow{2}{*}{Method} & \multicolumn{2}{c}{Backbone} \\
& Inc.V4 & Res.101   \\ \hline\hline
$L^2-\textit{\text{FE}}$ \cite{li2019delta}                         & 84.1               & 85.3             \\
$L^2$ \cite{li2019delta}                         & 85.8               & 87.2             \\
$L^2-\textit{\text{SP}}$ \cite{li2019delta}                       & 85.3               & 87.2            \\
DELTA \cite{li2019delta}                  & \underline{86.8}               & \underline{88.7}             \\
GBN \cite{liu2021transtailor}          & -                    & 86.9             \\
TransTailor \cite{liu2021transtailor}          & -                    & 87.3             \\\rowcolor{lightblue}
CLTR (SGP)               & \textbf{88.6 }             & \textbf{89.8}      \\\hline
\end{tabular}}
  \end{minipage}%
  \hfill
\end{table}

\subsection{Training Order}\label{app:order}

Our theoretical analysis motivates a specific ordering for CLTR: the model should first learn the largest partition, i.e., the Head classes, and then progressively incorporate smaller Tail partitions while preventing forgetting. This ordering follows directly from Theorems~\ref{theorem1}--\ref{theorem 2}, which show that standard training on a long-tailed dataset yields parameters close to the solution obtained by training on the Head classes alone. Therefore, the Head solution provides a natural anchor point for the subsequent CL stage. Learning the Tail after the Head allows the CL objective to move the model toward Tail-sensitive solutions while preserving Head knowledge.

To verify this prediction empirically, we evaluate CLTR(SGP) on CIFAR100-LT with $\IFNew=100$ under three task orders: Head$\rightarrow$Tail, Tail$\rightarrow$Head, and Random. The results are reported in Table~\ref{tab:order}. These results confirm that the ordering is not an arbitrary implementation detail: it is a key part of the CLTR formulation and is consistent with the geometry predicted by our theory.

\begin{table}[ht!]
\caption{Effect of task order for CLTR(SGP) on CIFAR100-LT with $\IFNew=100$. Head$\rightarrow$Tail follows the cardinality-sorted ordering motivated by our theory and achieves the best performance.}
\label{tab:order}
\vspace{0.7em}
\scriptsize
\setlength{\tabcolsep}{6pt}
\centering
\resizebox{0.45\linewidth}{!}{
\begin{tabular}{lc}
\hline
\addlinespace[0.2em]
Training order & Top-1 accuracy \\
\hline\hline
Tail$\rightarrow$Head & 36.0 \\
Random & 32.3 \\
\rowcolor{lightblue}
Head$\rightarrow$Tail (CLTR order) & \textbf{50.7} \\
\hline
\end{tabular}
}
\end{table}

\subsection{Practical estimation of \texorpdfstring{$\beta_{\min}$}{beta min} and \texorpdfstring{$\alpha_{\min}$}{alpha min}}\label{app:min}
To illustrate that the conditions in Theorem~\ref{theorem_general} are mild and typically satisfied by standard CL methods, we provide a concrete empirical estimation of $\beta_{\min}$ and $\alpha_{\min}$ in a real-world setting. We consider a ResNet-18 trained on the Head classes of CIFAR-100-LT, matching our experimental setup.

\textbf{Estimating $\beta_{\min}$}:
Recall that
$\beta_{\min} = \frac{L_H^{\mathrm{sm}}}{2\,m_{\Psi}},$
where $L_H^{\mathrm{sm}}$ is the local smoothness constant of the Head loss and $m_{\Psi}$ is the strong-convexity parameter of the parameter-space divergence $D_\Psi$.
We approximate $L_H^{\mathrm{sm}}$ by the top eigenvalue of the Hessian of $L_H$ at $\theta_H^\star$ using Hessian–vector products and power iteration.
For the trained ResNet-18, we obtain
$L_H^{\mathrm{sm}} \approx 0.2.$
When $D_\Psi$ is the standard $\ell_2$ divergence (the default choice in many EWC-style methods), the corresponding Bregman divergence is 2-strongly convex, implying $m_\Psi \approx 1$.
This gives the practical threshold
$\beta_{\min} \approx \frac{0.2}{2} = 0.1.$
This value is substantially \emph{below} the regularization strengths typically used in off-the-shelf CL methods (e.g., $\beta = 10$–$10^3$ in EWC variants), indicating that the theoretical requirement is easily satisfied in practice.

\textbf{Estimating $\alpha_{\min}$}:
In the unified CL formulation, the output-space regularizer takes the form $\alpha\,D_{\Phi}(h_{\theta}, z_H^\star)$, where $D_\Phi$ is a Bregman divergence in logit space.
In Theorem~\ref{theorem_general}, $\alpha_{\min}$ is defined as
$\alpha_{\min} = \frac{c_{\Phi}}{2},$
where $c_{\Phi}$ is the smallest constant satisfying
$L_H(\theta)-L_H(\theta_H^\star) \le c_\Phi\, D_{\Phi}(h_\theta, z_H^\star) \quad \text{for all }\theta\text{ in a local neighborhood.}$
For Bregman divergences $D_\Phi$ induced by a strongly convex potential in logit space (e.g., squared-logit or KL divergences), this inequality is equivalent to a standard smoothness–strong-convexity relation in the logit domain. In particular, if $L_H$ is $L_{\text{out}}$-smooth with respect to logits and $D_\Phi$ is $m_\Phi$-strongly convex, then
$c_\Phi = \frac{L_{\text{out}}}{m_\Phi}.$
Substituting this expression into $\alpha_{\min} = c_\Phi/2$ yields the empirically estimable form
$\alpha_{\min} = \frac{L_{\text{out}}}{2 m_\Phi},$
which we use below.
$L_{\text{out}}$ quantifies the local smoothness of $L_H$ with respect to logits and $m_\Phi$ is the strong-convexity constant of $D_\Phi$.
To estimate $L_{\text{out}}$, we compute the largest eigenvalue of the logit-space Hessian $\nabla_z^2 L_H$ at $\theta_H^\star$, again using power iteration on batches of Head samples. Since $L_H$ is a batch-averaged loss in our theory, this batch-averaged Hessian is the correct object to analyze.
In our ResNet-18 / CIFAR-100-LT Head-class setting, we observe
$L_{\text{out}} \approx 2 \times 10^{-4}.$
For squared-logit or KL-based output divergences, the associated Bregman divergence is 1-strongly convex in the relevant output domain, giving $m_\Phi \approx 1$. Hence,
$\alpha_{\min} \approx \frac{2 \times 10^{-4}}{2} = 10^{-4}.$
Again, this threshold is far smaller than the regularization strengths commonly used in output-regularized CL methods (e.g., LwF typically uses $\alpha$ between $10^{-2}$ and $1$). Thus, even modest output-space regularization comfortably exceeds $\alpha_{\min}$.

Both estimated thresholds, $\beta_{\min} \approx 0.1$ and $\alpha_{\min} \approx 10^{-4}$, lie \emph{orders of magnitude below} the $\alpha$ and $\beta$ values routinely used in practical CL methods. This empirical check supports the conclusion that the theoretical conditions of Theorem~\ref{theorem_general} are not restrictive and are naturally satisfied by off-the-shelf continual learning algorithms.

\paragraph{Empirical Verification of the Upper Bound.} To investigate whether the upper-bound inequality in Theorem~\ref{theorem_general} remains satisfied during Tail learning, we evaluate its empirical margin at intermediate training checkpoints. Because the radius of the neighborhood $\mathcal{N}$ is not directly identifiable, parameter distances alone cannot certify neighborhood membership. Instead, for random seed $s$ and checkpoint $t$, we measure:
\begin{equation}\label{eq:cltr_bound_margin}
M_{s,t}=\mathcal{L}_{\mathrm{CL}}(\theta_{s,t})+\frac{1}{2}\mathcal{L}_H(\theta_{H,s}^{\ast})-\mathcal{L}_{\mathrm{bal}}(\theta_{s,t}),
\end{equation}
where $M_{s,t}\geq 0$ indicates that the upper-bound inequality holds at the corresponding checkpoint. We evaluate CIFAR100-LT ($IF=100$, 10,847 training samples) using ResNet-32 trained from scratch, with three random seeds $s\in\{0,1,2\}$. During the 100-epoch Tail stage, we record checkpoints at epochs $\{1,11,21,31,41,51,61,71,81,91,100\}$, yielding 33 measurements per method. Table~\ref{tab:bound_margin} reports the worst-case margin $\min_{s,t}M_{s,t}$ across all seeds and checkpoints, rather than an average.

\begin{table}[t]
\centering
\caption{Empirical verification of the upper-bound inequality on CIFAR100-LT ($IF=100$) across 33 measurements per method.}
\label{tab:bound_margin}
\small
\begin{tabular}{lcc}
\hline
Method & Min. margin & Satisfied \\
\hline
Naive sequential & $-12.15$ & $0/33$ \\
CLTR (LwF) & $+2.25$ & $33/33$ \\
CLTR (EWC) & $+0.07$ & $33/33$ \\
\hline
\end{tabular}
\end{table}

Both CLTR variants satisfy the evaluated inequality at all recorded checkpoints, while naive sequential training, for which the CL regularization coefficients are zero, violates it throughout. The worst-case EWC margin is close to zero, whereas the LwF certificate is looser. Furthermore, the measured Head-stage residual loss is $0.0170\pm0.0009$, corresponding to an additive constant of only $0.0085\pm0.00045$. Thus, the observed satisfaction of the inequality is not attributable to a large residual Head-loss constant. These results provide empirical support for the upper-bound certificate along the evaluated trajectories but do not establish that every iterate remains inside the neighborhood required by Theorem~\ref{theorem_general}.

\subsection{Multiple Incremental Steps} \label{app:step}
\begin{figure}[t]
        \begin{minipage}[t]{.43\linewidth}
        \includegraphics[width = 1.17\linewidth]{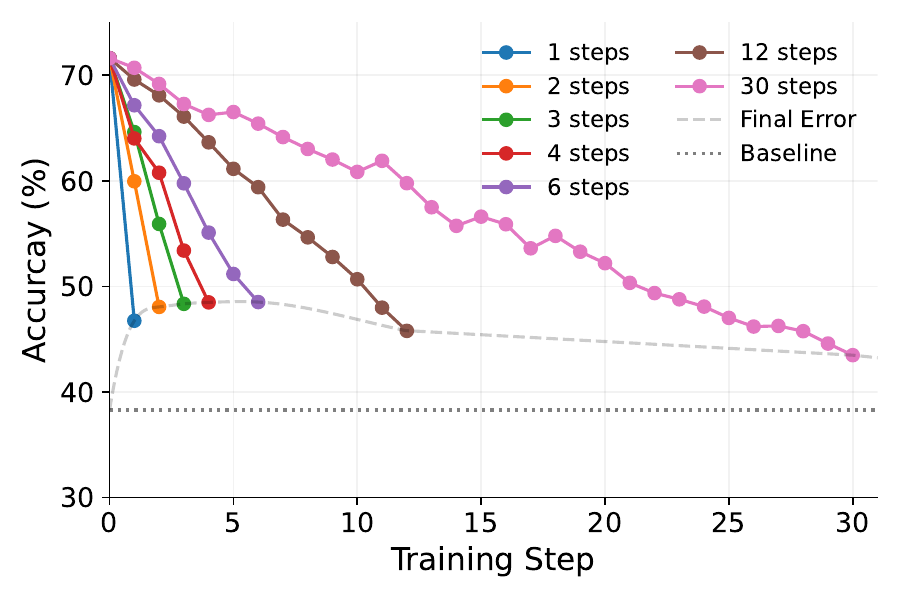}
    \hfill
    \end{minipage}
    \hspace{0.5cm}
    \begin{minipage}[t]{.43\linewidth}
        \centering
        \includegraphics[width = 1.17\linewidth]{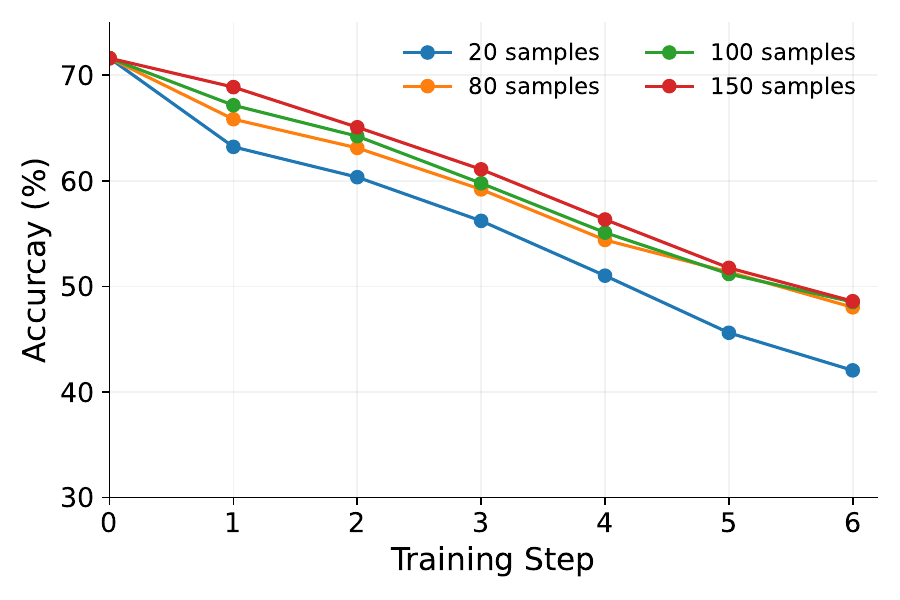}
\end{minipage}
\vspace{-4mm}
\caption{CLTR performance for various (left) number of incremental steps and  (right) replay memory ($\mathcal{M}$) size.}\label{fig:buffer_tradeoff}
\vspace{-3mm}
\end{figure}

Recall that Theorem \ref{theorem 2} highlights CLTR's capability for extending beyond two incremental steps. Increasing the number of partitions leads to smaller $\IFNew$ within each partition at the cost of an increase in forgetting. To explore the effect of varying partition numbers on CLTR's final performance, we adhere to the experimental protocol outlined in \cite{liu2022long}. Initially, the model is trained on the first 60 classes, followed by sequential learning of the remaining classes, divided into different numbers of partitions. The results of this experiment are presented in Fig. \ref{fig:buffer_tradeoff} (left). Our results reveal an optimal value for CLTR (FOSTER) (4 steps), yet the performance margin remains slim even with up to 30 tasks. This highlights the effectiveness of CL methods employed within CLTR. The optimal value for the number of incremental steps for each CL algorithm can be found in Appendix \ref{ID}.

To further evaluate the boundary-selection criterion of Proposition~\ref{prop:partitioning}, we vary the Head--Tail class boundary while keeping the remaining training configuration fixed. Specifically, we evaluate CLTR (SGP) with two partitions on CIFAR100-LT ($IF=100$) using three class splits: 20:80, 50:50, and 80:20. As shown in Table~\ref{tab:boundary_sensitivity}, the 50:50 split approximately equalizes the local imbalance factors and achieves the highest accuracy among the evaluated boundaries, outperforming the alternative splits by 3.7 and 5.0 percentage points. These results empirically support the boundary-selection criterion in Proposition~\ref{prop:partitioning}.

\subsection{Replay Memory} \label{app:replay} Several CL algorithms incorporate mechanisms to retain partial information from the previous task, aiming to mitigate catastrophic forgetting. For example, EWC maintains prior model parameters along with their Fisher values, whereas both GPM and SGP safeguard the Core Gradient Space of previous tasks. FOSTER, on the other hand, utilizes a replay memory for this purpose. Within the LTR context, the presence of a buffer memory doesn't require additional storage, as access to the full dataset is already available. Nonetheless, to prevent hindering the model's capacity to learn Tail distributions, we deliberately avoid replaying all Head samples when learning the Tail (evidenced by Eq. \ref{loss_4}). Accordingly, our analysis extends to how the number of Head samples replayed while learning the Tail impacts the model's performance, as illustrated in Fig. \ref{fig:buffer_tradeoff} (right). The replay memory's size serves as a mediator between forgetting previous information and worsening class imbalance, e.g. a larger replay memory reduces forgetting but increases imbalance. Therefore, identifying an optimal balance in this trade-off is crucial. Our results demonstrate the significance of an appropriate replay memory size; however, there exists a threshold beyond which additional samples per class do not further
improve the performance and
the performance levels off.

\begin{table}[t]
\centering
\caption{Sensitivity of CLTR (SGP) to Head--Tail class boundaries on CIFAR100-LT ($IF=100$) with two partitions.}
\label{tab:boundary_sensitivity}
\small
\begin{tabular}{lcc}
\hline
Head:Tail split & Local IFs & Top-1 (\%) \\
\hline
20:80 & [2.43, 39.40] & 46.0 \\
\textbf{50:50} & \textbf{[9.8, 9.6]} & \textbf{49.7} \\
80:20 & [41.67, 2.40] & 44.7 \\
\hline
\end{tabular}
\end{table}

\subsection{Runtime}\label{app:run}

We evaluate the computational overhead of CLTR against standard LTR training on CIFAR100-LT using ResNet-32, a batch size of 128, and a single NVIDIA RTX 3090 GPU. All methods use identical data-loading and evaluation settings, and the measurements include method-specific computations such as Fisher information estimation for EWC and singular value decomposition (SVD) for SGP. We report wall-clock training time averaged over three runs, total training FLOPs, and peak GPU memory in Table~\ref{tab:runtime}. Using WD+MaxNorm as a computationally efficient LTR reference, CLTR introduces a moderate, method-dependent overhead. Specifically, LwF, EWC, and SGP increase wall-clock training time by 14.3\%, 25.0\%, and 34.9\%, respectively, while increasing training FLOPs by 11.2--16.4\%. The larger runtime overhead of SGP is primarily associated with its SVD computation; performing SVD on the GPU reduces its overhead from 34.9\% to 23.8\%. Peak GPU memory increases by at most 40 MB across the evaluated CLTR variants. These results demonstrate that CLTR incurs additional training costs associated with the selected CL algorithm, rather than being computationally equivalent to standard LTR training. However, for the evaluated methods, the overhead remains relatively modest, and inference complexity is unchanged when the same backbone architecture is retained.  The measurements in Table~\ref{tab:runtime} are obtained under a consistent CIFAR100-LT/ResNet-32 benchmarking setup on an RTX 3090, providing a comparative assessment of computational overhead across methods. The reported wall-clock times correspond to this benchmark configuration, while experiment-specific training schedules, including that of the released CIFAR-SGP implementation, are documented separately in Appendix~D.

\begin{table}[t]
\centering
\caption{Computational overhead on CIFAR100-LT using ResNet-32 and a single RTX 3090 GPU. Wall-clock time is averaged over three runs. Relative values are normalized to WD+MaxNorm.}
\label{tab:runtime}
\small
\setlength{\tabcolsep}{4pt}
\begin{tabular}{lccccc}
\hline
Method & Time (s) & Rel. & FLOPs (T) & Rel. & Mem. (MB) \\
\hline
LTR (WD) & $205.5\pm1.7$ & 0.986 & 227.0 & 1.000 & 334 \\
LTR (WD+MaxNorm) & $208.3\pm7.5$ & 1.000 & 227.0 & 1.000 & 334 \\
CLTR (no CL term) & $238.6\pm1.9$ & 1.145 & 252.1 & 1.111 & 334 \\
CLTR (LwF) & $238.2\pm1.4$ & 1.143 & 264.4 & 1.164 & 374 \\
CLTR (EWC) & $260.4\pm13.0$ & 1.250 & 254.3 & 1.120 & 340 \\
CLTR (SGP) & $280.9\pm2.7$ & 1.349 & 252.5 & 1.112 & 343 \\
CLTR (SGP, GPU SVD) & $257.9\pm8.3$ & 1.238 & 252.5 & 1.112 & 351 \\
\hline
\end{tabular}
\end{table}

\subsection{Weight Imbalance}\label{app:imb}  
\begin{figure}[t]
    \centering
    \includegraphics[width=0.35\textwidth]{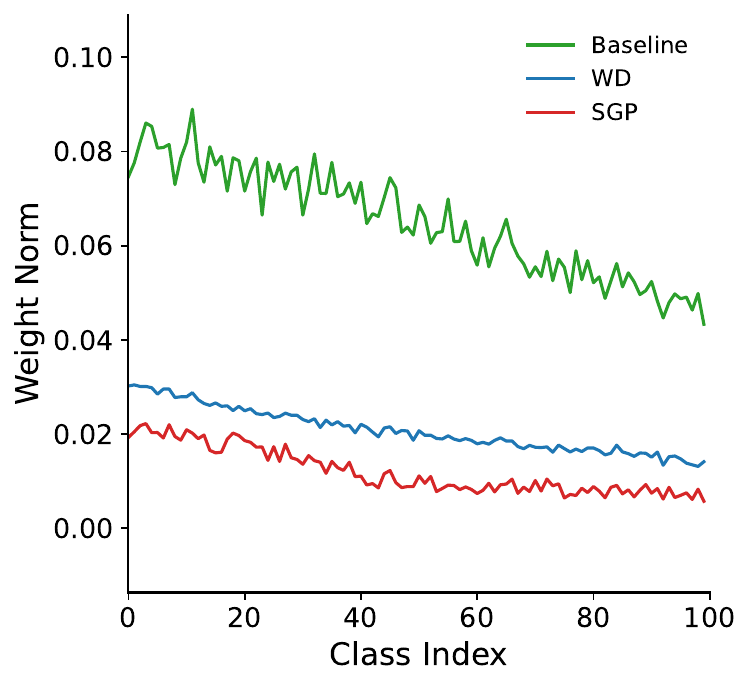}
    \caption{Per-class weight norms of the baseline, SGP, and WD.}\label{norm}
\end{figure}

An interesting phenomenon observed when training models on highly imbalanced data is the presence of artificially large weights in neurons corresponding to the Head classes \cite{alshammari2022long}.
The LTR solution, WD, addresses this problem by penalizing weight growth using weight decay.
One way to assess the network's ability to handle LTR is by analyzing the bias in per-class weight norms. To this end, we present the per-class weight norms of the Baseline, WD, and CLTR (SGP) models in Fig. \ref{norm}.

The figure reveals a significant imbalance in the weight norms of the Baseline model, which is naively trained on the imbalanced dataset. In contrast, the WD and CLTR (SGP) models exhibit more uniform weight norms across different classes. Interestingly, although CLTR (SGP) starts with the heavily imbalanced weights of the Baseline model, it converges towards a more uniform weight distribution without any explicit penalty on weight growth. Unlike many other CL methods that restrict the plasticity of crucial weights, SGP only constrains the direction of the weight update in the weight space, enabling the model to converge to a more balanced weight distribution. This further demonstrates the effectiveness of CL in addressing LTR problem.

\begin{figure}[t]
\begin{center}
\includegraphics[width=\textwidth]{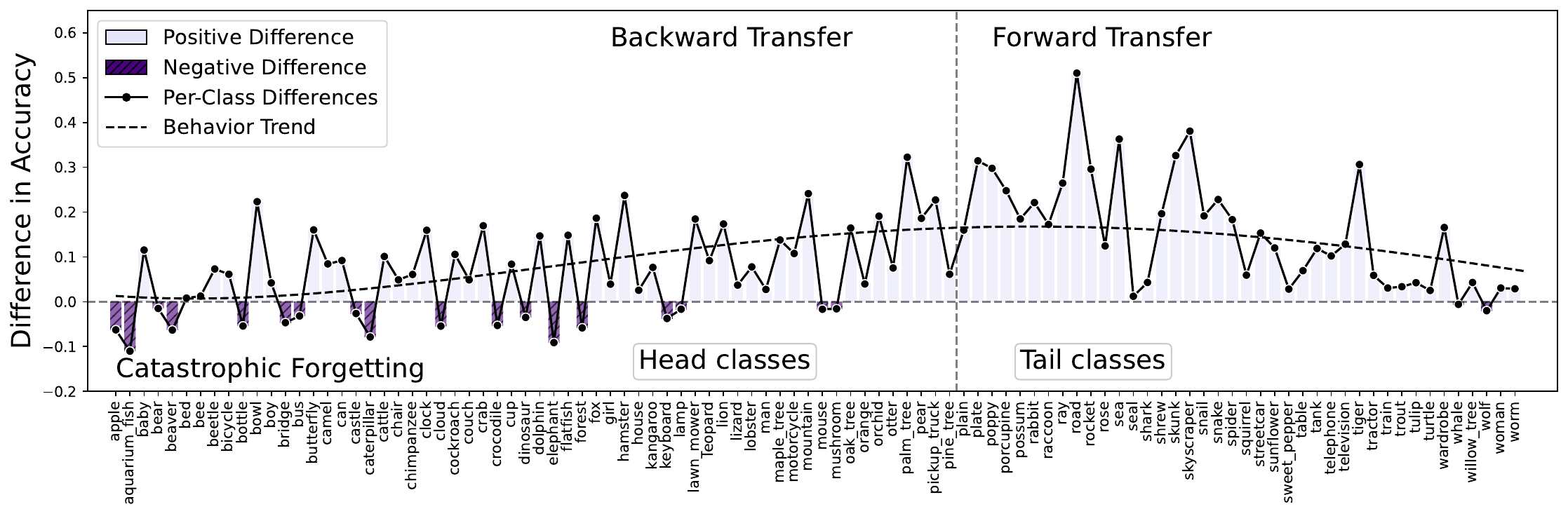}
\end{center}
\caption{The difference in per-class accuracy of CLTR (SGP) and the baseline model. \faSearch}\label{gpmvsbaseline}

\end{figure}
\subsection{Backward/Forward Transfer and Catastrophic Forgetting} \label{app:back}  Prior works discuss three key concepts in the context of CL: catastrophic forgetting, backward transfer, and forward transfer \cite{díazrodríguez2018dont}. As mentioned earlier, catastrophic forgetting occurs when the performance of a class declines after retraining. Despite the use of CL methods, which are designed to mitigate this forgetting, a certain degree of forgetting is still inevitable. Forward transfer is the improvement in performance on a new task after employing CL, which is the central aim of retraining in CL. Finally, backward transfer is a beneficial side-effect where retraining on new samples can actually enhance the model's performance on the previous tasks. This interesting phenomenon in CL has been extensively discussed in previous works in theory and practice \cite{lin2022beyond}.  Now, let us discuss Fig. \ref{gpmvsbaseline}, which presents the difference in per-class accuracy of CLTR (SGP) versus the baseline network. The analysis is based on CIFAR100-LT with an $\IFNew$ of 100.
The figure is divided into three regions corresponding to the scenarios discussed above: catastrophic forgetting (bottom), backward transfer (top-left), and forward transfer (top-right). The bottom region in the figure represents classes that undergo catastrophic forgetting, while the top-right region represents the Tail samples (with a class index larger than 60), which demonstrate improved performance, or forward transfer.
We observe that using SGP as a CL solution for LTR results in very effective improvements in the per-class accuracy of the Tail (forward transfer).  Interestingly, despite the absence of Head data in the retraining process, 42 out of 60 Head classes see some level of improvement after the model is exposed to the Tail samples (backward transfer). This result emphasizes the remarkable potential of CL methods in enhancing performance in both new and previous tasks.

\subsection{CLTR performance on ViT backbone} \label{app:vit}
Our choice of ResNet-32 on CIFAR and ResNeXt-50 on ImageNet followed standard LTR benchmark protocols to ensure fair comparison with prior baselines  \cite{alshammari2022long,fu2022long}. Theorem \ref{theorem_general} is an optimization-level guarantee: it depends only on local loss properties, not on a specific architecture. Therefore, it applies to any differentiable model, provided Assumption \ref{ass:cltr} and the coefficient conditions of Theorem \ref{theorem_general} hold
. To empirically verify this, we have also conducted additional experiments with a ViT-B backbone on CIFAR100-LT:

\begin{table}[ht!]

\caption{Performance of CLTR with a ViT-B backbone on CIFAR100-LT under different imbalance factors.}

\label{tab:vitb_results}

\vspace{0.7em}

\scriptsize

\setlength{\tabcolsep}{6pt}

\centering

\resizebox{0.4\linewidth}{!}{

\begin{tabular}{lccc}

\hline

\addlinespace[0.2em]

\multirow{2}{*}{Method (ViT-B)} & \multicolumn{3}{c}{$\IFNew$} \\

& 10 & 50 & 100 \\

\hline\hline

Baseline \cite{cui2019class} & 66.02 & 54.78 & 50.59 \\

BCE \cite{cui2019class} & 64.63 & 50.65 & 45.50 \\

CB \cite{cui2019class} & \underline{66.30} & \underline{56.18} & \underline{50.63} \\

\rowcolor{lightblue}

CLTR (TPL) & \textbf{69.74} & \textbf{59.71} & \textbf{55.94} \\

\hline

\end{tabular}
}
\end{table}

CLTR (TPL) outperforms the baselines across every imbalance factor, confirming that the framework transfers directly to modern architectures without any modification.

\subsection{Generalization to Long-Tailed Text Classification}\label{app:nlp}

To investigate whether CLTR generalizes beyond vision, we evaluate it on 20 Newsgroups-LT, constructed from the 20 Newsgroups dataset using the standard \texttt{bydate} split, with headers, footers, and quotes removed. The 20 classes are sorted by training frequency and subsampled according to the exponential profile $n_i=n_{\max}IF^{-i/(C-1)}$, with $IF=100$, $C=20$, and $n_{\max}=377$. This produces 1,729 training documents, divided into two class-disjoint partitions of 10 Head classes (1,593 samples) and 10 Tail classes (136 samples), following Proposition~\ref{prop:partitioning}. Evaluation uses the original test split containing 7,532 documents.

We fine-tune the \texttt{sentence-transformers/all-mpnet-base-v2}\footnote{\hyperlink{https://huggingface.co/sentence-transformers/all-mpnet-base-v2}{https://huggingface.co/sentence-transformers/all-mpnet-base-v2}} encoder end-to-end, using mean token pooling followed by dropout ($0.1$) and a linear classifier. Input sequences are truncated or padded to 128 tokens. Training uses AdamW with weight decay $0.01$, batch size 32, gradient clipping at $1.0$, bf16 mixed precision, and a OneCycle learning-rate schedule with 10\% linear warmup and linear decay. The initial learning rate is $3\times10^{-5}$, increased fourfold to $1.2\times10^{-4}$ during Tail training for all sequential methods. Each partition is trained for 15 epochs, while joint training uses 15 epochs on the full dataset. All sequential variants share the same Head-stage initialization. For CLTR (LwF), we use a frozen Head-stage teacher with distillation weight $\alpha=0.03$ and temperature $T=2$. For CLTR (GPM), we use an energy threshold $\epsilon=0.95$ and hard gradient projection across all encoder linear layers, excluding the classifier, with up to 4,000 sampled activation vectors per layer. CLTR variants additionally retain 50 exemplars per Head class for replay.

We compare CLTR against joint cross-entropy training on the long-tailed dataset, naive Head-to-Tail sequential training without a CL mechanism, and balanced training ($IF=1$) as a reference. We evaluate Head, Tail, and balanced accuracy, with balanced accuracy defined as the mean per-class recall across all 20 classes.

\begin{table}[t]
\centering
\caption{Performance on 20 Newsgroups-LT ($IF=100$). Head, Tail, and Balanced denote macro-averaged classification recall (\%).}
\label{tab:nlp}
\small
\begin{tabular}{lccc}
\hline
Method & Head & Tail & Balanced \\
\hline
Balanced training ($IF=1$) & 77.42 & 66.18 & 71.80 \\
Long-tailed joint training & 81.53 & 33.20 & 57.37 \\
Naive sequential & 0.00 & 32.78 & 16.39 \\
CLTR (LwF) & \textbf{79.21} & 41.69 & 60.45 \\
CLTR (GPM) & 77.39 & \textbf{44.72} & \textbf{61.06} \\
\hline
\end{tabular}
\end{table}

As shown in Table~\ref{tab:nlp}, joint long-tailed training exhibits substantial Head bias, while naive sequential training suffers catastrophic forgetting. CLTR mitigates this issue, improving Tail accuracy over joint training by 8.49 and 11.52 percentage points with LwF and GPM, respectively, while achieving balanced accuracy gains of 3.08 and 3.69 points. These results provide preliminary evidence that the benefits of Head-to-Tail continual learning extend to text classification, although broader validation across additional NLP benchmarks and random seeds remains future work.

\section{Broader Impact}\label{app:broad}
LTR is fundamental to applications where data acquisition is expensive or naturally imbalanced like medical imaging (rare disease classification), ecological monitoring (rare species), industrial inspection (rare defects), and satellite imagery (rare land-cover classes). Improving long-tail performance directly benefits these high-impact applications by enabling reliable recognition of underrepresented categories. CLTR is method-agnostic and can be applied with any off-the-shelf CL technique, lowering the barrier to deploying LTR in practice. We do not foresee significant negative societal impact specific to this work; however, like all classification techniques, CLTR could be misapplied in surveillance or biometric contexts where rare-category recognition raises privacy concerns. The pretrained CLIP backbone we use inherits any biases from its pretraining data; downstream practitioners should audit foundation model biases before deployment.

\end{document}